\documentclass{article}
\usepackage{iclr2026_conference,times}
\usepackage{amsmath,amssymb,amsfonts}
\usepackage{booktabs}
\usepackage{float}
\usepackage{graphicx}
\usepackage{xcolor}
\usepackage{hyperref}
\hypersetup{hidelinks}
\usepackage{url}
\iclrfinalcopy


\graphicspath{{./}{figures/scaling/}}

\title{Optimizer Memory Schedules for\\Outscaling the Overtraining Axis}
\author{Katie Everett\thanks{Correspondence to \texttt{everettk@csail.mit.edu}.}\\
MIT CSAIL
\And
Shikai Qiu\\
New York University}

\begin{document}
\raggedbottom
\maketitle
\fancyhead{}

\begin{abstract}
We investigate how optimizers scale across the overtraining axis and show that
relative optimizer performance and optimal hyperparameters change substantially
with training horizon. In particular, we study how matrix-preconditioned methods
(Muon and SOAP) and a momentum-scheduled method (ADANA) scale relative to AdamW.
We compare these four optimizers across models from 51M to 253M parameters and
overtraining (OT) factors from $1\times$ to $256\times$, sweeping the base learning
rate at every setting. The preferred learning rate schedule can reverse across
the overtraining axis, the best weight decay coefficient scales approximately
as $\sqrt{OT}$, and longer horizons generally favor longer fixed memory. ADANA's
scaling advantage over AdamW persists after tuning AdamW's fixed memory
separately at each horizon. Log-time weight decay and momentum cooldown provide
substantial gains for ADANA that compound as training increases. With this
treatment, ADANA outscales AdamW with an exponent advantage close to that
predicted by DANA theory on power-law random features. Muon and SOAP instead
provide roughly constant token-efficiency advantages over AdamW across most of
the measured range, although SOAP may gain further at the highest overtraining
factors. ADANA begins behind both matrix-preconditioned optimizers but closes the
gaps as training increases, surpassing Muon and becoming competitive with SOAP
at our highest OT factors. These results establish training horizon as an
essential axis for optimizer evaluation and design.

\end{abstract}

\section{Introduction}
\label{sec:introduction}

Language models exhibit empirical scaling laws relating loss to model size,
training data, and compute \citep{kaplan2020scaling}. In particular, the
Chinchilla scaling law finds that model size and training tokens should grow
together to minimize loss at fixed pretraining compute, corresponding to
approximately 20 training tokens per parameter
\citep{hoffmann2022empirical}. This result motivated subsequent model and
optimizer scaling studies to evaluate performance near the Chinchilla-optimal
training horizon.
However, for widely deployed language models, recurring inference costs far
exceed the one-time cost of pretraining. This can favor training smaller models
on far more tokens to achieve similar performance with lower inference costs,
a practice known as \emph{overtraining}
\citep{sardana2024beyond,gadre2024overtraining,bian2025inference}. For a
particular model size, the \emph{overtraining factor} expresses the number of
training tokens as a multiple of the compute-optimal value.

It is common for optimizer benchmarks to select hyperparameters and compare
methods at a single training horizon or fixed token-to-parameter ratio.
Optimizer rankings and optimal hyperparameter choices can change with training
duration~\citep{wen2025fantastic,semenov2025benchmarking,wen2026hyperball}, yet
prior work covers a limited range of overtraining factors. We therefore study
how optimizer performance and optimal hyperparameters scale across the
overtraining axis.

The optimizers we study combine different forms of preconditioning with
different treatments of memory. AdamW uses coordinatewise adaptive
preconditioning, whereas Muon and SOAP apply matrix-level
preconditioning~\citep{jordan2024muon,vyas2025soap}; all three use scalar
momentum coefficients that keep their memory horizons fixed throughout
training. DANA instead derives a momentum schedule whose memory grows
throughout training, and ADANA incorporates this schedule into an Adam-style
adaptive optimizer~\citep{ferbach2025dana,ferbach2026adana}. We compare AdamW,
ADANA, Muon, and SOAP to ask how fixed memory, matrix preconditioning, and
scheduled memory affect optimizer scaling across the overtraining axis.

To study this question, we train models spanning 51M to 253M parameters with
each optimizer across overtraining factors ranging from $1\times$ to
$256\times$. For every optimizer, model size, and overtraining factor, we
independently sweep the base learning rate and use the lowest final validation
loss. Following prior work, we measure relative token efficiency at fixed model
size using the ratio of the training tokens required by a baseline optimizer
to those used by the indicated optimizer at the same
loss~\citep{qiu2025hyperparameter,ferbach2026adana,wen2026hyperball}. We call
this ratio the \emph{token multiplier}; values above one mean that the baseline
requires more training tokens and therefore favor the indicated optimizer.

Our experiments first show that optimizer treatments must themselves scale
with the training horizon. The preferred learning rate schedule can reverse
across the overtraining axis, and the best weight decay coefficient scales
approximately as $\sqrt{f}$. For fixed-memory optimizers, longer horizons
generally favor longer memory, while the optimal learning rate decreases as
memory grows. Jointly tuning memory and learning rate at each horizon improves
performance particularly at the lowest and highest OT factors.
We also propose a momentum cooldown rule that bounds the optimizer's memory
timescale during terminal learning rate decay, yielding substantial gains for
ADANA that compound with the benefits of log-time weight decay.

Finally, we find clear optimizer outscaling across the overtraining axis. With
log-time weight decay and momentum cooldown, ADANA outscales AdamW with an
equivalent-OT exponent close to the $2-\kappa$ prediction from DANA theory on
power-law random features~\citep{ferbach2025dana}. By contrast, Muon and SOAP
provide roughly constant token-multiplier advantages over AdamW across most of
the overtraining axis, although SOAP may gain further at our highest OT
factors. ADANA begins substantially behind these matrix-preconditioned
optimizers but gains on both as the training horizon increases, surpassing Muon
and becoming competitive with SOAP. We use joint model--token loss fits and
floor--decay likelihood profiles to examine whether this outscaling arises
from an improved token-decay exponent, a lower finite-model high-token loss
limit, or both.

\begin{figure}[tp]
  \centering
  \includegraphics[width=\textwidth]{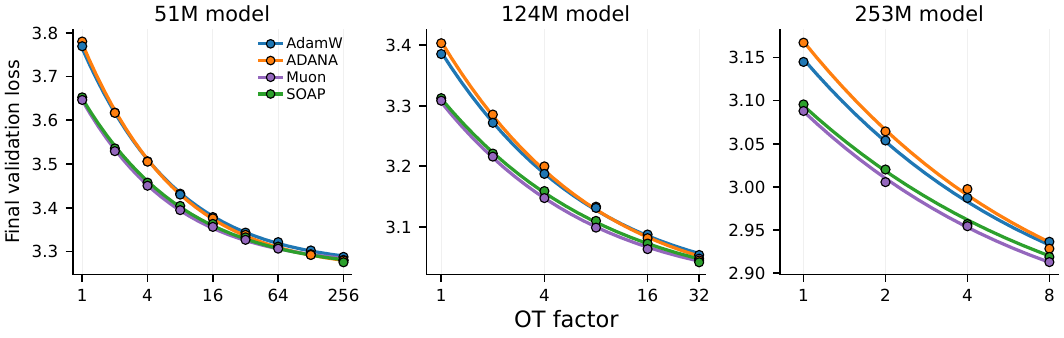}
  \caption{\textbf{Optimizer rankings exhibit crossovers as training extends
  beyond the Chinchilla-optimal horizon.} We report the
  lowest final validation loss after independently sweeping the base learning
  rate for AdamW, ADANA, Muon, and SOAP on 51M, 124M, and 253M models under
  uniform weight decay. Muon and SOAP lead at short horizons, whereas ADANA
  closes its initial gap as the OT factor increases, and SOAP overtakes Muon
  at the longest measured horizons for the 51M and 124M models. Solid curves
  are fitted using our shared-$m$ variant of the Skaling functional
  form~\citep{videau2026skaling}; see Appendix~\ref{app:scaling-fits}.}
  \label{fig:main-gamma-one-loss}
\end{figure}

\begin{figure}[tp]
  \centering
  \includegraphics[width=\textwidth]{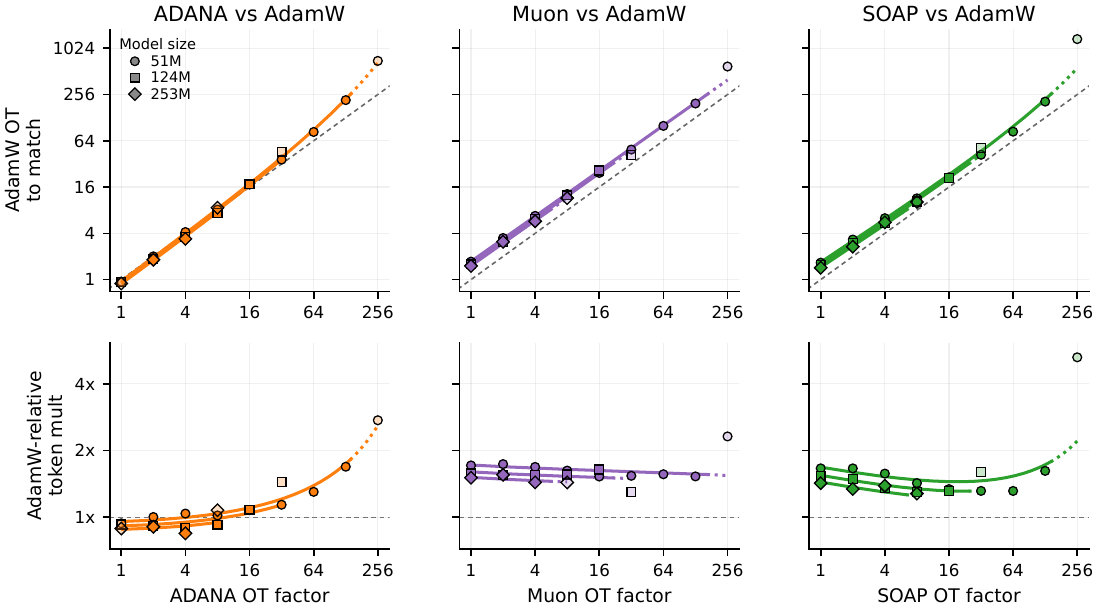}
  \caption{\textbf{Across model sizes under uniform weight decay, ADANA improves
  relative to AdamW as the OT factor increases, whereas Muon and SOAP already
  have an advantage at short horizons. Muon's advantage remains roughly
  constant across OT factors, while SOAP may improve further at the highest OT
  factors.} Top: the vertical axis shows the OT factor AdamW requires to match
  the indicated optimizer's loss at fixed model size; values above the equal-OT
  diagonal favor the indicated optimizer.
  Bottom: the AdamW-relative token multiplier is the equivalent AdamW OT factor
  divided by the optimizer's OT factor, or equivalently the ratio of tokens
  required by AdamW to those used by the optimizer. Curves compare the fitted
  optimizer and AdamW losses, while markers invert measured optimizer losses
  through the fitted AdamW curve. Dotted curve segments and translucent marker
  fills identify target losses below AdamW's best measured loss and therefore
  require extrapolating the AdamW fit.}
  \label{fig:main-token-multiplier-collapse}
\end{figure}

\section{Related Work}
\label{sec:related-work-early}

\textbf{Long and scheduled optimizer memory.} Most optimizers use exponential
averages with coefficients that keep their memory timescales fixed throughout
training, but several recent methods instead combine or schedule multiple
timescales. AdEMAMix combines separate fast and slow momentum buffers, with
distinct coefficients controlling their memory
timescales~\citep{pagliardini2024ademamix}. Schedule-free methods use iterate
averaging, which can be viewed as a form of memory in parameter space that
preserves information from earlier optimizer updates~\citep{defazio2024road}.
ScheduleFree+ reports that the benefit of this parameter-averaging memory
increases with training duration~\citep{defazio2026sfplus}.
\citet{morwani2025connections} connect schedule-free optimizers and AdEMAMix
to accelerated SGD variants that decouple momentum decay from the weight
assigned to the current gradient. Their experiments find larger gains at small
batch sizes, consistent with these approaches being particularly effective in
noise-dominated optimization. DANA derives a momentum timescale that grows
throughout training and improves loss-scaling exponents in the power-law
random features model, while ADANA incorporates this schedule into an
Adam-style adaptive optimizer~\citep{ferbach2025dana,ferbach2026adana}.
Optimal hyperparameters also change with training horizon and batch size:
\citet{bjorck2025tokenhorizons} show empirically that the optimal learning
rate decreases predictably with token horizon, while
\citet{marek2025smallbatch} find that Adam's $\beta_2$ should scale with batch
size to keep the second-moment half-life fixed in tokens rather than optimizer
updates.

Weight decay controls a related form of memory: whereas momentum determines
how long past gradient information influences the optimizer state, decoupled
weight decay determines how long earlier optimizer updates remain influential
in the parameters~\citep{wang2025adamw}. Its optimal timescale depends on the
data-to-model ratio~\citep{bergsma2025powerlines}, and ADANA introduces a
log-time weight decay schedule whose timescale grows throughout
training~\citep{ferbach2026adana}.

\textbf{Matrix-preconditioned optimizers.} Muon, SOAP, and Shampoo can be
viewed as matrix-preconditioned optimizers, in contrast to adaptive optimizers
such as Adam, whose coordinatewise normalization acts as a diagonal
preconditioner. In a high-dimensional least-squares model,
\citet{paquette2026phasesmuon} show that an idealized version of Muon provides
a square-root spectral speedup for eigenmodes large enough to be resolved at
the given batch size, whereas unresolved modes behave like SGD. Empirically,
\citet{qiu2025hyperparameter} find that Muon, SOAP, and Shampoo achieve
approximately constant $1.4\times$ compute-multiplier gains over AdamW across
model widths from 190M to 1.4B when both the learning rate and weight decay are
scaled appropriately.

\textbf{Scaling-law functional forms.} For scaling analyses that vary both
model size and training horizon, \citet{videau2026skaling} propose a functional
form that couples model size and training data through an interaction exponent.
We use this functional form with a model exponent shared across optimizers,
similar to \citet{volkova2026robust}, who show that sharing scaling-law
parameters across optimizers improves identifiability.

\textbf{Optimizer comparisons across training horizons.} While many optimizer
benchmarks compare methods at a single training horizon
or fixed token-to-parameter ratio, recent studies have begun to investigate
how relative optimizer performance changes across the overtraining axis.
\citet{wen2025fantastic} report that, relative to AdamW, the token multipliers
for Muon and SOAP are approximately $1.2\times$--$1.4\times$ on 130M--520M
models from $1\times$ to $8\times$ Chinchilla. Muon performs better through
$4\times$, while SOAP catches or overtakes it at $8\times$ and in their
additional $16\times$ experiments. On the 1.2B model, both token multipliers
decrease to approximately $1.1\times$. \citet{semenov2025benchmarking} find
that AdamW narrows and eventually reverses SOAP's short-horizon advantage as
training extends. They also show that the long-horizon ordering between SOAP
and AdEMAMix depends on retuning AdEMAMix's slow-memory coefficient: SOAP
overtakes AdEMAMix when its shorter-horizon coefficient is reused, whereas
increasing $\beta_3$ restores AdEMAMix's lead. \citet{wen2026hyperball} report
that, relative to AdamW, the token multiplier for standard Muon remains
approximately $1.1\times$ across $1\times$--$8\times$ Chinchilla, whereas the
token multiplier for Muon Hyperball increases across the same range from
approximately $1.2\times$ to $1.3\times$. These results show that optimizer
advantages can change with training horizon and experimental treatment, while
leaving their behavior under substantially greater overtraining unresolved.

\section{Scaling Optimizers Across Training Horizons}
\label{sec:comparison-framework}

\subsection{Momentum and Weight Decay as Memory}
\label{sec:optimizer-memory-view}

Throughout this paper, we use the term \emph{memory} to refer to how long
information from earlier optimizer updates remains influential.

For a momentum state with constant coefficient $\beta$,
\begin{equation}
  m_t
  =
  \beta m_{t-1}+(1-\beta)g_t
  =
  (1-\beta)\sum_{s=1}^{t}\beta^{t-s}g_s,
\end{equation}
where $m_0=0$.  The relative contribution of a gradient from $a$ updates ago
is therefore $\beta^a$.  Its half-life and effective memory window are
\begin{equation}
  H_{1/2}=\frac{\log(1/2)}{\log\beta},
  \qquad
  M=\frac{1}{1-\beta},
\end{equation}
with $H_{1/2}\approx(\log 2)M$ near $\beta=1$.  Fixed-momentum optimizers use
a coefficient that keeps this timescale constant across training steps,
whereas scheduled-memory optimizers vary it throughout training.  A similar
interpretation applies to exponential moving averages of squared gradients.

Whereas momentum controls how long the optimizer state remembers past
gradient information, weight decay controls how quickly the parameters forget
it~\citep{wang2025adamw}.  Under decoupled weight
decay~\citep{loshchilov2019decoupled}, parameter
decay is applied separately from the optimizer direction $d_t$:
\begin{equation}
  \theta_{t+1}=(1-\alpha_t)\theta_t-\eta_t d_t,
\end{equation}
where $\eta_t$ is the learning rate and $\alpha_t$ is the fraction of the
existing parameter removed at update $t$.  When $\alpha_t=\alpha>0$ is
constant, we can equivalently write
\begin{equation}
  \theta_{t+1}
  =
  (1-\alpha)\theta_t
  +
  \alpha\left(-\frac{\eta_t}{\alpha}d_t\right),
\end{equation}
making the parameter iterate an exponential moving average of scaled optimizer
updates.  The influence of an update from $a$ steps ago decays as
$(1-\alpha)^a$, giving an effective memory window of approximately $1/\alpha$
updates.  A time-varying weight decay schedule therefore changes how quickly
the parameters forget updates from different parts of training.

In our study of the overtraining axis, this perspective lets us interpret the
effects of momentum and weight decay through memory timescales measured in
optimizer updates.

\subsection{Experimental Setup}
\label{sec:horizon-dependent-treatments}

We compare AdamW, ADANA, Muon, and SOAP at fixed model size as we scale the
overtraining factor.  We define $f=T/(20P)$, where $T$ is the number of
training tokens and $P$ is the nominal parameter count.  For the 51M, 124M,
and 253M models, we evaluate OT factors up to $256\times$, $32\times$, and
$8\times$, respectively.  All core comparisons use a sequence length of
$2{,}048$ and a global batch size of 256 sequences, so increasing the OT
factor proportionally increases the number of optimizer updates.  At every
setting, we independently sweep the base learning rate and report the lowest
final validation loss.

To determine the experimental setup for our core optimizer comparisons, we
first studied learning rate schedule and weight decay choices across
overtraining factors.  We find that the preferred choices depend on the
overtraining factor and that the ordering between candidates can reverse
across the overtraining axis.  After a shared linear warmup, we compared linear
decay to zero with cosine decay to zero.  For all four optimizers, linear decay
performs better at $1\times$ OT, whereas cosine decay performs better at
$8\times$ and $32\times$ OT.  We also find that the benefit of log-time weight
decay depends on the learning rate endpoint, with larger gains under cosine
decay to $10\%$ than under cosine decay to zero.

On the 51M model, we jointly sweep the base learning rate and weight decay
coefficient for each optimizer, OT factor, and weight decay schedule.
Consistent with the square-root scaling predicted for uniform AdamW by
\citet{bergsma2025powerlines}, the preferred coefficient scales approximately
as $\sqrt{f}$ for both uniform and log-time weight decay.  For most
experiments, we therefore use
$c_{\mathrm{uniform}}(f)=8\sqrt{f}$ or
$c_{\log}(f)=2\sqrt{f}$.  For optimizers that benefit from log-time weight
decay, its advantage over uniform weight decay increases with the overtraining
factor.

At $1\times$ OT on the 253M model, we tune AdamW's $\beta_2$, Muon's matrix
momentum $\beta$, and SOAP's $\beta_2$; each selects $0.98$.  We hold the
remaining memory coefficients at standard values: $\beta_1=0.9$ for AdamW,
$\beta_1=\beta_{\mathrm{Sh}}=0.95$ for SOAP, and
$(\beta_1,\beta_2)=(0.9,0.95)$ for Muon's Adam-routed parameters.  For ADANA,
DANA theory relates $\kappa$ to the power-law spectrum of the
data~\citep{ferbach2025dana}, and the ADANA experiments find that $\kappa=0.85$
performs well across model scales and architectures, consistent with $\kappa$
being a transferable property of the data
distribution~\citep{ferbach2026adana}.  We therefore fix $\kappa=0.85$ and
$\delta=8$ and tune $g_3$ at the same setting, selecting $g_3=8$.  These fixed
ADANA coefficients define a memory schedule that evolves throughout training.
The supporting experiments appear in
Appendices~\ref{app:learning-rate-schedule}, \ref{app:weight-decay}, and
\ref{app:fixed-memory-tuning}; Appendix~\ref{app:full-protocol} gives the
complete treatment definitions.

\begin{figure}[tp]
  \centering
  \includegraphics[width=\textwidth]{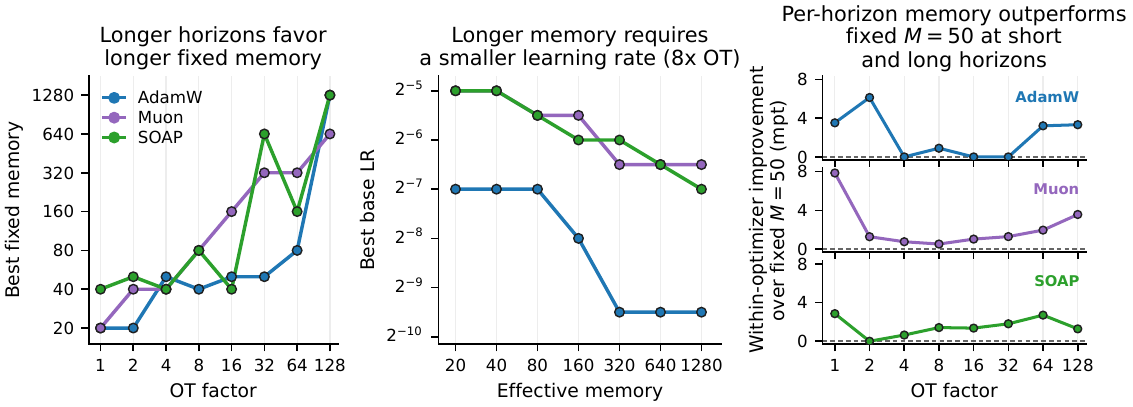}
  \caption{\textbf{Longer horizons favor longer optimizer memory, and longer
  memory requires a smaller learning rate.} All panels use the 51M model.
  Left: the optimal fixed effective memory for AdamW, Muon, and SOAP after
  independently sweeping the base learning rate at every memory and OT
  factor. Middle: at $8\times$ OT, the optimal base learning rate decreases as
  effective memory increases; Figure~\ref{fig:fixed-momentum-learning-rate-scaling}
  shows the corresponding relationship at every measured OT factor. Right:
  the within-optimizer validation-loss improvement from replacing fixed
  $M=50$ with the optimal $(\text{memory},\text{base learning rate})$ pair at
  each horizon; positive values favor per-horizon tuning.}
  \label{fig:main-memory-frontier}
\end{figure}

\subsection{Outscaling and Token Multipliers}

We characterize outscaling through two closely related metrics that compare
how many training tokens two optimizers require to reach the same loss at fixed
model size.  For an indicated optimizer and a specified baseline, the
\emph{equivalent OT factor} is the OT factor the baseline requires to match the
indicated optimizer's loss.  The \emph{token multiplier} is the ratio of the
tokens required by the baseline to those used by the indicated optimizer;
equivalently, it is the equivalent baseline OT factor divided by the indicated
optimizer's OT factor.  Values above the equal-OT diagonal favor the indicated
optimizer.  At fixed model size, this ratio is
the same quantity called the \emph{compute multiplier} by
\citet{qiu2025hyperparameter,ferbach2026adana} and \emph{token-equivalent
speedup} by \citet{wen2026hyperball}.  We refer to it as the \emph{token
multiplier}: values above one mean that the baseline requires more training
tokens and therefore favor the indicated optimizer.

We say that an optimizer \emph{outscales a baseline optimizer across the
overtraining axis} when its token multiplier increases with the OT factor, or
equivalently when its equivalent-OT curve grows faster than the equal-OT
reference.  Because different optimizers or treatments may approach different
high-token validation-loss limits, this increase may arise from an improved
token-decay exponent, a lower high-token validation-loss limit, or both.

We calculate these metrics by inverting baseline loss curves.  Unless
otherwise stated, we fit these curves using the Skaling functional
form~\citep{videau2026skaling}, with the model exponent shared across
optimizers.  Appendix~\ref{app:scaling-fits} gives the complete definition and
compares alternative functional forms.  When the target loss lies below the
baseline optimizer's best measured loss, this inversion requires extrapolation
rather than interpolation.  We denote these regions with dotted curves and
translucent markers.

\subsection{Optimizer Rankings Across Training Horizons}
\label{sec:scaling-results}

Figures~\ref{fig:main-gamma-one-loss}
and~\ref{fig:main-token-multiplier-collapse} show that optimizer rankings and
relative token efficiency change with training horizon.  Muon and SOAP perform
best at most short and intermediate horizons, while ADANA improves relative to
AdamW as the overtraining factor increases; at the longest measured horizons
for the 51M and 124M models, SOAP also overtakes Muon.  The token multipliers
in Figure~\ref{fig:main-token-multiplier-collapse} show that Muon provides an
approximately constant advantage over AdamW, ADANA's advantage increases
consistently across model sizes, and SOAP may gain further at the highest OT
factors.  In the following sections, we investigate how per-horizon
fixed-memory tuning, log-time weight decay, and momentum cooldown improve
optimizer performance across OT factors, and then ask whether the resulting
optimizer treatments outscale one another across the overtraining axis.

\section{Horizon-Tuned Fixed Memory}
\label{sec:optimizer-memory}

Spectral analyses of learning dynamics suggest that optimization resolves
progressively harder modes over the course of training; longer averaging
windows may help accumulate the weak, persistent gradient signal needed to
resolve modes learned later in training
\citep{saxe2014exact,bordelon2020spectrum}. Longer windows also respond more
slowly to changes in the gradient distribution and retain unusually large or
stale gradients for more updates, creating a stability tradeoff. We therefore
ask how the optimal fixed memory changes with overtraining factor and jointly
tune memory and learning rate at every horizon on the 51M model. We note that
the optimal momentum coefficient should depend on batch size
\citep{marek2025smallbatch}, so these results are specific to our global batch
size of 256.

\subsection{Longer horizons favor longer memory}
\label{sec:preferred-fixed-memory}

We sweep AdamW's second-moment coefficient $\beta_2$, Muon's momentum
coefficient $\beta$, and SOAP's second-moment coefficient $\beta_2$, reporting
each through the effective memory window $M$ defined in
Section~\ref{sec:optimizer-memory-view}.
For every $(\text{optimizer},M,\text{OT factor})$ setting, we independently
sweep the base learning rate; at each horizon, we select the
$(M,\text{base learning rate})$ pair with the lowest final validation loss.
For experimental details and complete results, see
Appendix~\ref{app:fixed-memory-tuning}.

We see in Figure~\ref{fig:main-memory-frontier} that longer horizons generally
favor longer fixed memory. From $1\times$ to $128\times$ OT, the optimal memory
increases from $M=20$ to $M=1280$ for AdamW, from $M=20$ to $M=640$ for Muon,
and from $M=40$ to $M=1280$ for SOAP. We do not expect this trend to continue
indefinitely: extremely long windows may become unstable, particularly early
in training when the optimization dynamics change rapidly. Many horizons also
have broad low-regret memory ranges, with several neighboring values performing
similarly. We find that the optimal base learning rate decreases as memory
grows, suggesting that using longer memory requires smaller steps for
stability.

\subsection{Per-horizon tuning improves fixed-memory optimizers}
\label{sec:fixed-memory-counterfactual}

Per-horizon memory tuning produces its largest gains at the low and high ends
of our overtraining range; at moderate horizons, our $M=50$ setting is already
near-optimal. At $128\times$ OT, it improves final validation loss by $3.3$ mpt
for AdamW, $3.6$ mpt for Muon, and $1.3$ mpt for SOAP. Thus, $\beta=0.98$ is a
strong choice at moderate OT factors but does not transfer uniformly across
the full overtraining range.

\begin{figure}[tp]
  \centering
  \includegraphics[width=\textwidth]{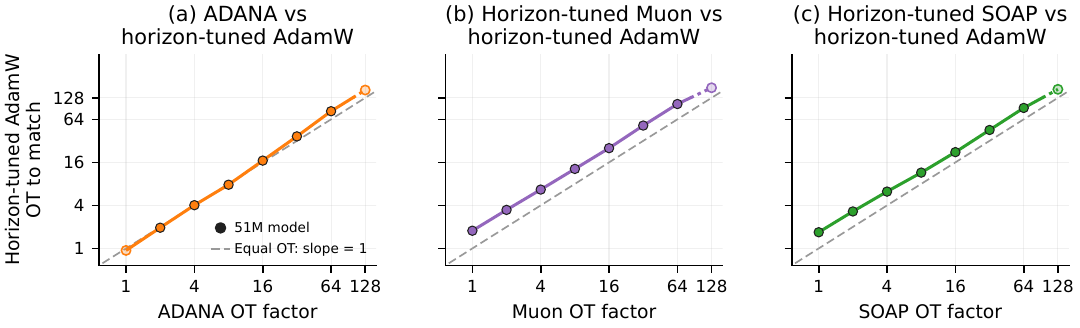}
  \caption{\textbf{After tuning fixed memory separately at every horizon,
  ADANA still gains on AdamW across OT, while Muon and SOAP retain roughly
  constant token-efficiency advantages over AdamW.} On the 51M model under
  uniform weight decay, AdamW, Muon, and SOAP select the optimal
  $(\text{fixed memory},\text{base learning rate})$ pair at every OT factor.
  ADANA uses the fixed scheduled-memory setting
  $(\kappa,\delta,g_3)=(0.85,8,8)$ and sweeps only the base learning rate. The
  vertical axis gives the horizon-tuned AdamW OT
  factor required to attain the same validation loss. Due to the small number
  of points, we compute equivalent OT by piecewise-linear interpolation of
  horizon-tuned AdamW loss in log-OT; translucent markers require extrapolation
  beyond the measured horizon-tuned AdamW loss range.}
  \label{fig:main-horizon-memory-outscaling}
\end{figure}

Figure~\ref{fig:main-horizon-memory-outscaling}a compares ADANA against AdamW
with horizon-tuned memory. ADANA continues to gain on this stronger AdamW
baseline as the overtraining factor increases, showing that its empirical
advantage is not explained by holding AdamW's $\beta_2$ fixed across horizons
and is consistent with a benefit from adapting memory within the run.
Figures~\ref{fig:main-horizon-memory-outscaling}b--c show that horizon-tuned
Muon and SOAP retain roughly constant advantages over horizon-tuned AdamW.

\section{Log-Time Weight Decay Across Training Horizons}
\label{sec:horizon-aware-forgetting}

Under the memory interpretation in Section~\ref{sec:optimizer-memory-view},
the weight decay schedule determines how quickly the parameters forget past
updates during training. In our implementation, the per-update decay fraction
is
\begin{equation}
  \alpha_t=s_t\lambda_t,
\end{equation}
where $s_t\in[0,1]$ is the normalized learning rate schedule. Uniform weight
decay keeps $\lambda_t$ constant across optimizer updates:
\begin{equation}
  \lambda_t^{\mathrm{uniform}}=\frac{c(f)}{S}.
\end{equation}
Log-time weight decay instead decreases the coefficient over
training~\citep{ferbach2026adana}:
\begin{equation}
  \lambda_t^{\log}=\frac{a(f)}{\tau+t}
\end{equation}
Here, $S$ is the total number of optimizer updates and
$\tau=0.1S_{1\times}$.  We use the coefficient rules
$c(f)=8\sqrt{f}$ and $a(f)=2\sqrt{f}$, fitted from our joint weight decay
coefficient and learning rate sweeps.  Their $\sqrt{f}$ dependence agrees
with the scaling predicted by \citet{bergsma2025powerlines}.  These rules
approximate the trend across optimizers rather than selecting the optimal
weight decay coefficient separately at every optimizer and overtraining
factor.  We independently sweep the base learning rate for every optimizer,
weight decay schedule, and overtraining factor.  See
Appendix~\ref{app:weight-decay} for additional details.

\begin{figure}[tp]
  \centering
  \includegraphics[width=\textwidth]{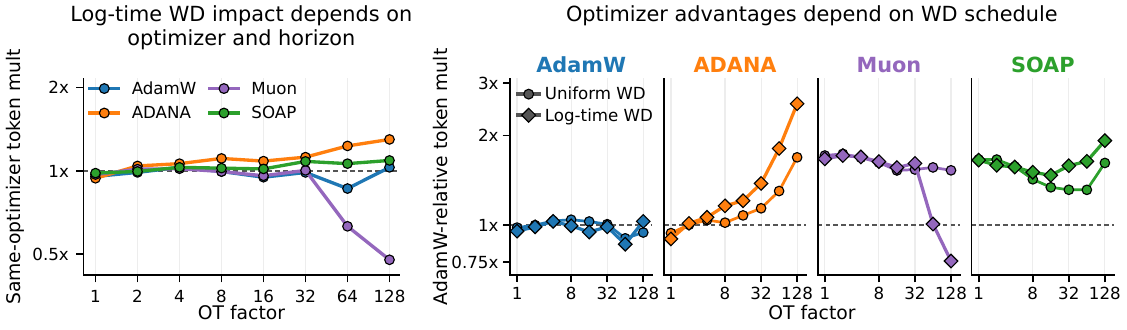}
  \caption{\textbf{Log-time weight decay increasingly benefits ADANA, modestly
  benefits SOAP, has little consistent effect on AdamW, and harms Muon at high
  OT factors.} On the 51M-parameter model, we compare uniform weight decay,
  $c(f)=8\sqrt{f}$, with log-time weight decay, $a(f)=2\sqrt{f}$,
  independently sweeping the base learning rate at every optimizer, weight
  decay schedule, and OT factor. Left: the token multiplier of log-time weight
  decay relative to the same optimizer under uniform weight decay. Right:
  AdamW-relative token multipliers for both weight decay schedules, using
  AdamW with uniform weight decay as the common baseline.}
  \label{fig:main-log-time-wd}
\end{figure}

Figure~\ref{fig:main-log-time-wd} shows that, under these coefficient rules,
the preferred weight decay schedule depends on both the optimizer and training
horizon.  Log-time weight decay increasingly benefits ADANA as the
overtraining factor grows, modestly benefits SOAP, has little consistent
effect on AdamW, and harms Muon at high OT.  The Muon result may reflect a
poorly matched log-time weight decay coefficient rather than the schedule
itself.  Figures~\ref{fig:main-gamma-one-loss}
and~\ref{fig:main-token-multiplier-collapse} use uniform weight decay for all
optimizers.  In Section~\ref{sec:model-token-scaling}, we instead use the
better-performing weight decay schedule for each optimizer when comparing
their strongest treatments.

\begin{figure}[tp]
  \centering
  \includegraphics[width=\textwidth]{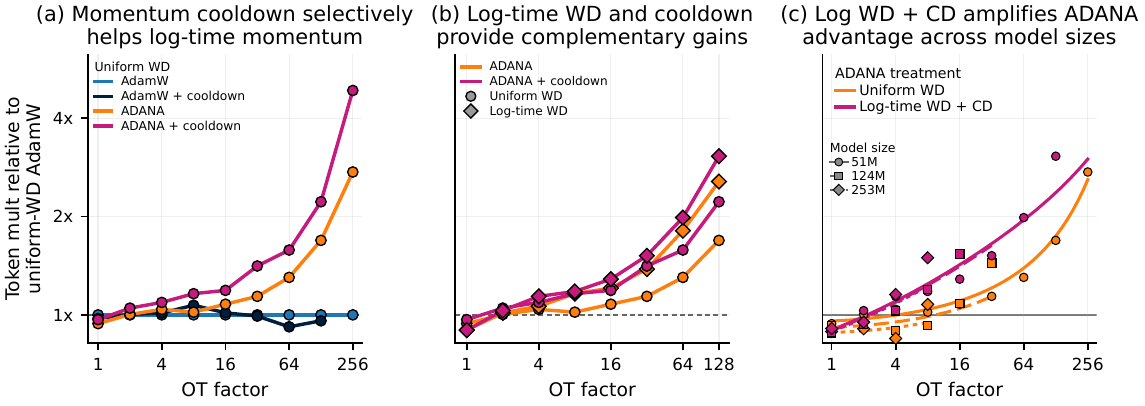}
  \caption{\textbf{Momentum cooldown selectively benefits ADANA's log-time
  memory and combines with log-time weight decay to amplify ADANA's advantage
  as the overtraining factor increases, consistently across model sizes.} All
  panels use AdamW with uniform weight decay at the same model size as the
  token-multiplier baseline. Left: on the 51M model under uniform weight decay,
  momentum cooldown has little consistent effect on fixed-memory AdamW but
  substantially improves ADANA at long horizons. Middle: log-time weight decay
  and momentum cooldown each improve ADANA; combined, they yield the largest
  gains at long horizons. Right: across all three model sizes, the combined
  treatment outperforms ADANA with uniform weight decay. The resulting ADANA
  token multiplier follows an approximately straight line on log--log axes. We
  therefore use ADANA with log-time weight decay and momentum cooldown in the
  subsequent comparisons of the strongest treatments.}
  \label{fig:main-momentum-cooldown}
\end{figure}

\section{Momentum Cooldown for Growing Memory}
\label{sec:momentum-cooldown}

In this section, we propose a formulation of momentum cooldown, motivated by
the idea that ADANA's log-time momentum schedule makes its memory longest near
the end of training, when the learning rate approaches zero and responsiveness
may become more important.  Our momentum cooldown bounds the optimizer's memory
timescale during terminal learning rate decay.  Applied to ADANA, it preserves
the growing memory window through most of training but shortens it near the end
of the run.  A similar idea to reduce momentum at the end of training has been
used for Muon in nanochat~\citep{karpathy2025nanochat}.

We define the current learning rate decay timescale as
\begin{equation}
  \tau_{\eta,t}
  =
  \frac{\eta_t}{|\eta_{t+1}-\eta_t|},
\end{equation}
taking $\tau_{\eta,t}=\infty$ when the learning rate is constant or
increasing.  For any momentum state with coefficient $\beta_t$, we define its
original memory timescale as
\begin{equation}
  \tau_{\mathrm{mom},t}
  =
  \frac{1}{1-\beta_t}.
\end{equation}
In momentum cooldown, we replace this timescale with
\begin{equation}
  \tau_{\mathrm{CD},t}
  =
  \max\!\left(
    1,
    \min\!\left\{
      \tau_{\mathrm{mom},t},
      \tau_{\eta,t}
    \right\}
  \right),
  \qquad
  \beta_{\mathrm{CD},t}
  =
  1-\frac{1}{\tau_{\mathrm{CD},t}}.
\end{equation}
See Appendix~\ref{app:momentum-cooldown} for the complete definitions and
ablations.

Figure~\ref{fig:main-momentum-cooldown} shows that momentum cooldown has little
consistent effect on fixed-memory AdamW but substantially improves ADANA at
long horizons.  Log-time weight decay and momentum cooldown each improve ADANA
alone, and the gains compound when the two interventions are combined.  Across
all three model sizes, the combined treatment outperforms ADANA with uniform
weight decay.  In particular, its AdamW-relative token multiplier appears
approximately straight when plotted on log--log axes, motivating the outscaling
analysis in Section~\ref{sec:model-token-scaling}.  We therefore use ADANA with log-time
weight decay and momentum cooldown in the subsequent strongest-treatment
comparisons.  Our experiments measure improvements in final validation loss;
we leave additional analysis of the mechanism behind this benefit and
comparison of alternative momentum cooldown rules to future work.

\section{Outscaling Across the Overtraining Axis}
\label{sec:model-token-scaling}
\label{sec:outscaling-definition}

Using the treatment choices identified in the preceding sections, we now
compare how the resulting optimizer treatments scale relative to one another
across the overtraining axis.

\subsection{ADANA outscales AdamW}
\label{sec:outscaling-observations}

The DANA momentum schedule was originally derived to improve loss-scaling
exponents relative to SGD in the power-law random features model
\citep{ferbach2025dana}.  In this analysis, DANA accelerates an effective
optimization clock, which reparameterizes training time according to the rate
at which spectral modes are learned.  When the momentum damping schedule
scales as $t^{-\kappa}$, the momentum contribution to this clock grows as
\[
  \left(\int_0^t s^{-\kappa/2}\,ds\right)^2 \propto t^{2-\kappa},
\]
whereas the SGD clock grows linearly with $t$.  Consequently, matching DANA
after $t$ updates requires SGD to train for a horizon proportional to
$t^{2-\kappa}$.  In terms of our equivalent-OT metric, this would give DANA an
equivalent-OT exponent of $2-\kappa$ relative to SGD.

The PLRF model provides a tractable setting for high-dimensional optimization
theory, but whether this outscaling behavior transfers to the adaptive
versions, ADANA and AdamW, and to transformers is an open empirical question.
We therefore ask how ADANA's token efficiency relative to AdamW evolves across
the overtraining axis and how its equivalent-OT scaling compares with the
$2-\kappa$ prediction.

\begin{figure}[tp]
  \centering
  \includegraphics[width=\textwidth]{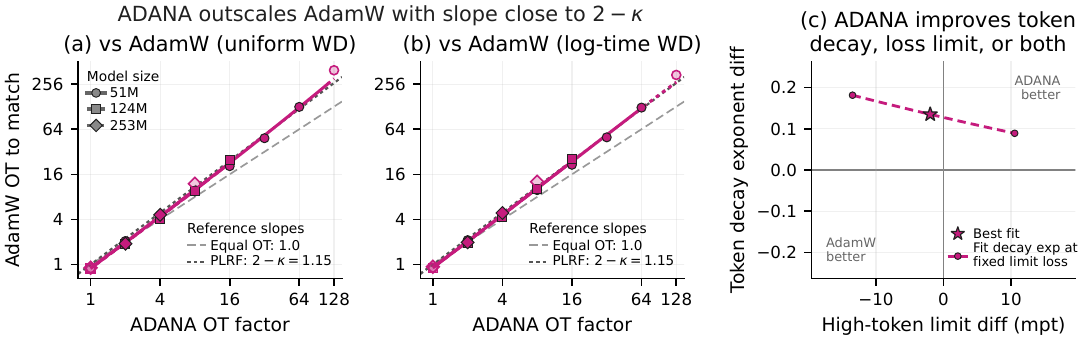}
  \caption{\textbf{With log-time weight decay and momentum cooldown, ADANA
  outscales AdamW with an equivalent OT slope close to the DANA PLRF prediction
  $2-\kappa$.} Left and middle: the AdamW OT factor required to match ADANA's
  loss at fixed model size, using AdamW with uniform weight decay and log-time
  weight decay as baselines, respectively.  Across model sizes and both
  baselines, the fitted log--log slopes range from $1.15$ to $1.20$, close to
  the PLRF prediction $2-\kappa=1.15$ for $\kappa=0.85$.  Dotted segments and
  translucent markers require extrapolating the AdamW fit.  Right: relative to
  AdamW with log-time weight decay, we plot the ADANA token-decay exponent
  difference implied by each fixed high-token-limit difference under the
  shared-$m$ Skaling fit.  The exponent difference remains above zero even
  where the high-token limit favors AdamW, providing evidence for an improved
  ADANA token-decay exponent while leaving the high-token-limit difference less
  well identified.}
  \label{fig:main-adana-outscaling-synthesis}
\end{figure}

Figure~\ref{fig:main-adana-outscaling-synthesis} compares ADANA with log-time
weight decay and momentum cooldown against AdamW with either uniform or
log-time weight decay.  Across all three model sizes and both AdamW baselines,
the fitted equivalent-OT exponents range from $1.15$ to $1.20$, close to
$2-\kappa=1.15$ for the $\kappa=0.85$ treatment used throughout our
experiments.  The floor--decay tradeoff analysis also shows that ADANA's fitted
token-decay exponent advantage remains positive throughout the range of
high-token-limit differences consistent with the data.
Appendix~\ref{app:outscaling-analysis} gives the complete equivalent-OT and
floor--decay analyses.

\subsection{Matrix-preconditioned optimizers provide approximately constant
gains over AdamW}

\begin{figure}[tp]
  \centering
  \includegraphics[width=\textwidth]{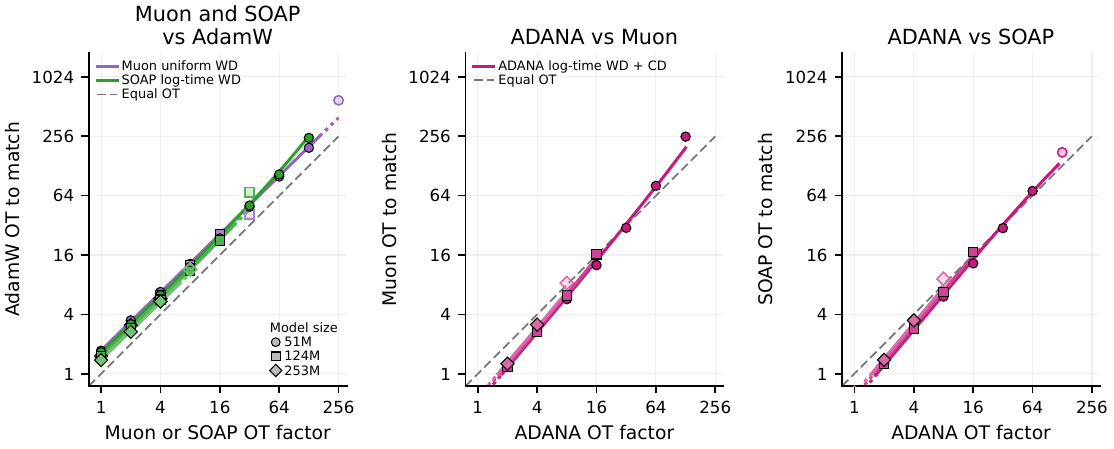}
  \caption{\textbf{ADANA versus matrix-preconditioned optimizers under their
  strongest treatments: ADANA closes the gap to Muon and SOAP as the
  overtraining factor increases.} AdamW and Muon use uniform weight decay,
  SOAP uses log-time weight decay, and ADANA uses log-time weight decay with
  momentum cooldown. Left: Muon and SOAP provide roughly constant
  token-multiplier advantages over AdamW across most of the overtraining axis,
  although SOAP may gain further at the highest OT factors. This suggests that
  matrix preconditioning may provide an approximately constant efficiency
  multiplier over AdamW. ADANA begins behind both matrix-preconditioned
  optimizers, but has a larger fitted token-decay exponent and closes the gap
  as the training horizon increases, catching Muon and becoming competitive
  with SOAP. Dotted segments and
  translucent markers indicate baseline extrapolation.}
  \label{fig:main-matrix-preconditioner-outscaling-synthesis}
\end{figure}

Figure~\ref{fig:main-matrix-preconditioner-outscaling-synthesis} compares Muon
and SOAP with AdamW. Across most overtraining factors, Muon provides a
$1.4\times$--$1.7\times$ token multiplier over AdamW, while SOAP provides a
$1.3\times$--$1.7\times$ multiplier. These results are consistent with the
matrix-preconditioned optimizers providing roughly constant token multipliers
over AdamW across OT factors. Whether this behavior persists at still higher
OT factors is unclear: SOAP's token multiplier rises to approximately
$1.9\times$ at $128\times$ OT on the 51M model, suggesting that it may gain
further with additional overtraining.

\subsection{ADANA outscales the matrix-preconditioned optimizers}
\label{sec:matrix-preconditioned-outscaling}

Finally, we compare ADANA's horizon-dependent gains with the
matrix-preconditioned optimizers. ADANA starts substantially behind: at
$2\times$ OT, its token multiplier is $0.59\times$--$0.63\times$ relative to
Muon and $0.63\times$--$0.69\times$ relative to SOAP across the three model
sizes. As the training horizon increases, ADANA closes these gaps, surpassing
Muon and becoming competitive with SOAP at the highest OT factors we test. It
appears that ADANA outscales Muon and may outscale SOAP, but these trends rely
heavily on the highest OT factors we test; it remains uncertain whether they
continue at even higher OT factors.

\section{Discussion and Limitations}
\label{sec:limitations}

Optimizer evaluation should treat training horizon as part of the experimental
design. An optimizer selected near the compute-optimal horizon may not remain
strongest when the same model is trained substantially longer, and its
preferred learning rate schedule, weight decay treatment, and memory can also
change across the overtraining axis. Long-horizon comparisons therefore
require scaling rules for hyperparameters that are often selected using
experiments at only one horizon.

Our results suggest complementary roles for preconditioning and scheduled
memory. In our experiments, Muon and SOAP provide large advantages over AdamW
across most horizons, while ADANA's relative efficiency increases with training
horizon. Tuning fixed memory separately at every horizon improves the
fixed-memory optimizers but does not explain ADANA's scaling advantage,
suggesting that adapting memory within a training run provides an additional
benefit.

Future work should establish how these conclusions transfer across larger
models, architectures, datasets, and training horizons. Preliminary experiments
in Appendix~\ref{sec:preconditioner-order} show that reversing the order of
adaptive preconditioning and scheduled momentum retains most of ADANA's
performance, indicating that scheduled memory can remain effective under
another preconditioner ordering. Future work could extend these experiments to
larger models and longer training horizons.

Our main optimizer comparisons use a global sequence batch of $256$.
Appendix~\ref{app:batch-steps} reports a fixed-token experiment at $8\times$
OT in which ADANA begins losing efficiency at smaller batch sizes than AdamW,
despite jointly optimizing the learning rate with $\beta_2$ for AdamW and
$g_3$ for ADANA. Because increasing batch size simultaneously reduces the
number of optimizer updates and gradient noise, and because other optimizer
hyperparameters remain fixed, this experiment does not identify the mechanism
or establish a general batch scaling law for ADANA. We therefore treat batch
dependence as an important limitation and leave a fully controlled study
across batch size, update count, and effective batch size to future work.

\section{Conclusion}
\label{sec:conclusion}
Optimizer rankings and optimal treatments change as models are trained beyond
their compute-optimal token budgets. ADANA's scaling advantage over AdamW
persists after tuning AdamW's fixed memory separately at each training horizon.
Log-time weight decay and momentum cooldown further amplify this advantage,
producing equivalent-OT scaling close to the $2-\kappa$ prediction from DANA
theory on power-law random features. Matrix-preconditioned optimizers provide
large but approximately constant gains over AdamW across most of the measured
range, while ADANA begins behind them and closes the gap as training increases.
Treating training horizon as an essential experimental axis reveals optimizer
behavior hidden by single-horizon comparisons and suggests that scheduling
optimizer memory is a promising route to methods whose relative efficiency
improves with additional training.

\subsubsection*{Acknowledgments}
The authors are grateful to the Google TPU Research Cloud program, MIT CSAIL,
and Kaiming He for providing the compute that made this work possible. We thank
Elliot Paquette and Ziyin Liu for technical discussions. We thank an anonymous
reviewer of a prior paper for suggesting that we investigate the interaction
between the learning rate schedule endpoint and the advantage of log-time
weight decay. SQ thanks the support of the Two Sigma Fellowship.

\bibliography{refs}

@article{wen2025fantastic,
  title={Fantastic Pretraining Optimizers and Where to Find Them},
  author={Wen, Kaiyue and Hall, David and Ma, Tengyu and Liang, Percy},
  journal={arXiv preprint arXiv:2509.02046},
  year={2025},
  url={https://arxiv.org/abs/2509.02046}
}

@article{wen2026hyperball,
  title={Fantastic Pretraining Optimizers and Where to Find Them {II}: {Hyperball} Optimization},
  author={Wen, Kaiyue and Dang, Xingyu and Lyu, Kaifeng and Ma, Tengyu and Liang, Percy},
  journal={arXiv preprint arXiv:2606.16899},
  year={2026},
  url={https://arxiv.org/abs/2606.16899}
}

@article{semenov2025benchmarking,
  title={Benchmarking Optimizers for Large Language Model Pretraining},
  author={Semenov, Andrei and Pagliardini, Matteo and Jaggi, Martin},
  journal={arXiv preprint arXiv:2509.01440},
  year={2025},
  url={https://arxiv.org/abs/2509.01440}
}

@article{pagliardini2024ademamix,
  title={The AdEMAMix Optimizer: Better, Faster, Older},
  author={Pagliardini, Matteo and Ablin, Pierre and Grangier, David},
  journal={arXiv preprint arXiv:2409.03137},
  year={2024},
  url={https://arxiv.org/abs/2409.03137}
}

@article{morwani2025connections,
  title={Connections between Schedule-Free Optimizers, {AdEMAMix}, and Accelerated {SGD} Variants},
  author={Morwani, Depen and Vyas, Nikhil and Zhang, Hanlin and Kakade, Sham},
  journal={arXiv preprint arXiv:2502.02431},
  year={2025},
  url={https://arxiv.org/abs/2502.02431}
}

@article{paquette2026phasesmuon,
  title={Phases of {Muon}: When {Muon} Eclipses {SignSGD}},
  author={Paquette, Elliot and Marshall, Noah and Benigni, Lucas and Wang, Guangyuan and Agarwala, Atish and Paquette, Courtney},
  journal={arXiv preprint arXiv:2605.09552},
  year={2026},
  url={https://arxiv.org/abs/2605.09552}
}

@article{ferbach2025dana,
  title={Dimension-adapted Momentum Outscales SGD},
  author={Ferbach, Damien and Everett, Katie and Gidel, Gauthier and Paquette, Elliot and Paquette, Courtney},
  journal={arXiv preprint arXiv:2505.16098},
  year={2025},
  url={https://arxiv.org/abs/2505.16098}
}

@article{ferbach2026adana,
  title={Logarithmic-time Schedules for Scaling Language Models with Momentum},
  author={Ferbach, Damien and Paquette, Courtney and Gidel, Gauthier and Everett, Katie and Paquette, Elliot},
  journal={arXiv preprint arXiv:2602.05298},
  year={2026},
  url={https://arxiv.org/abs/2602.05298}
}

@article{charles2025communication,
  title={Communication-Efficient Language Model Training Scales Reliably and Robustly: Scaling Laws for {DiLoCo}},
  author={Charles, Zachary and Teston, Gabriel and Dery, Lucio and Rush, Keith and Fallen, Nova and Garrett, Zachary and Szlam, Arthur and Douillard, Arthur},
  journal={arXiv preprint arXiv:2503.09799},
  year={2025},
  url={https://arxiv.org/abs/2503.09799}
}

@techreport{radford2019language,
  title={Language Models are Unsupervised Multitask Learners},
  author={Radford, Alec and Wu, Jeffrey and Child, Rewon and Luan, David and Amodei, Dario and Sutskever, Ilya},
  institution={OpenAI},
  year={2019},
  url={https://cdn.openai.com/better-language-models/language-models.pdf}
}

@misc{karpathy2025nanochat,
  title={{nanochat}: The Best {ChatGPT} That \$100 Can Buy},
  author={Karpathy, Andrej},
  year={2025},
  howpublished={\url{https://github.com/karpathy/nanochat}},
  note={Software repository}
}

@inproceedings{penedo2024fineweb,
  title={The {FineWeb} Datasets: Decanting the Web for the Finest Text Data at Scale},
  author={Penedo, Guilherme and Kydl{\'i}{\v c}ek, Hynek and Ben Allal, Loubna and Lozhkov, Anton and Mitchell, Margaret and Raffel, Colin and von Werra, Leandro and Wolf, Thomas},
  booktitle={Advances in Neural Information Processing Systems},
  volume={37},
  year={2024},
  url={https://arxiv.org/abs/2406.17557}
}

@article{touvron2023llama2,
  title={{Llama 2}: Open Foundation and Fine-Tuned Chat Models},
  author={Touvron, Hugo and Martin, Louis and Stone, Kevin and others},
  journal={arXiv preprint arXiv:2307.09288},
  year={2023},
  url={https://arxiv.org/abs/2307.09288}
}

@article{defazio2026sfplus,
  title={ScheduleFree+: Scaling Learning-Rate-Free and Schedule-Free Learning to Large Language Models},
  author={Defazio, Aaron},
  journal={arXiv preprint arXiv:2605.19095},
  year={2026},
  url={https://arxiv.org/abs/2605.19095}
}

@inproceedings{everett2024scaling,
  title={Scaling Exponents Across Parameterizations and Optimizers},
  author={Everett, Katie E. and Xiao, Lechao and Wortsman, Mitchell and Alemi, Alexander A. and Novak, Roman and Liu, Peter J. and Gur, Izzeddin and Sohl-Dickstein, Jascha and Kaelbling, Leslie Pack and Lee, Jaehoon and Pennington, Jeffrey},
  booktitle={Proceedings of the 41st International Conference on Machine Learning},
  series={Proceedings of Machine Learning Research},
  volume={235},
  pages={12666--12700},
  year={2024},
  publisher={PMLR},
  url={https://proceedings.mlr.press/v235/everett24a.html}
}

@inproceedings{kingma2015adam,
  title={Adam: A Method for Stochastic Optimization},
  author={Kingma, Diederik P. and Ba, Jimmy},
  booktitle={International Conference on Learning Representations},
  year={2015},
  url={https://arxiv.org/abs/1412.6980}
}

@inproceedings{loshchilov2019decoupled,
  title={Decoupled Weight Decay Regularization},
  author={Loshchilov, Ilya and Hutter, Frank},
  booktitle={International Conference on Learning Representations},
  year={2019},
  url={https://openreview.net/forum?id=Bkg6RiCqY7}
}

@inproceedings{wang2025adamw,
  title={How to Set {AdamW}'s Weight Decay as You Scale Model and Dataset Size},
  author={Wang, Xi and Aitchison, Laurence},
  booktitle={International Conference on Machine Learning},
  year={2025},
  url={https://arxiv.org/abs/2405.13698}
}

@inproceedings{vyas2025soap,
  title={{SOAP}: Improving and Stabilizing Shampoo Using Adam},
  author={Vyas, Nikhil and Morwani, Depen and Zhao, Rosie and Kwun, Mujin and Shapira, Itai and Brandfonbrener, David and Janson, Lucas and Kakade, Sham},
  booktitle={International Conference on Learning Representations},
  year={2025},
  url={https://openreview.net/forum?id=IDxZhXrpNf}
}

@misc{jordan2024muon,
  title={Muon: An Optimizer for Hidden Layers in Neural Networks},
  author={Jordan, Keller and Jin, Yuchen and Boza, Vlado and You, Jiacheng and Cecista, Franz and Newhouse, Laker and Bernstein, Jeremy},
  year={2024},
  url={https://kellerjordan.github.io/posts/muon/}
}

@article{qiu2025hyperparameter,
  title={Hyperparameter Transfer Enables Consistent Gains of Matrix-Preconditioned Optimizers Across Scales},
  author={Qiu, Shikai and Chen, Zixi and Phan, Hoang and Lei, Qi and Wilson, Andrew Gordon},
  journal={arXiv preprint arXiv:2512.05620},
  year={2025},
  url={https://arxiv.org/abs/2512.05620}
}

@article{kaplan2020scaling,
  title={Scaling Laws for Neural Language Models},
  author={Kaplan, Jared and McCandlish, Sam and Henighan, Tom and Brown, Tom B. and Chess, Benjamin and Child, Rewon and Gray, Scott and Radford, Alec and Wu, Jeffrey and Amodei, Dario},
  journal={arXiv preprint arXiv:2001.08361},
  year={2020},
  url={https://arxiv.org/abs/2001.08361}
}

@inproceedings{hoffmann2022empirical,
  title={An Empirical Analysis of Compute-Optimal Large Language Model Training},
  author={Hoffmann, Jordan and Borgeaud, Sebastian and Mensch, Arthur and Buchatskaya, Elena and Cai, Trevor and Rutherford, Eliza and de Las Casas, Diego and Hendricks, Lisa Anne and Welbl, Johannes and Clark, Aidan and Hennigan, Thomas and Noland, Eric and Millican, Katherine and van den Driessche, George and Damoc, Bogdan and Guy, Aurelia and Osindero, Simon and Simonyan, Kar{\'e}n and Elsen, Erich and Vinyals, Oriol and Rae, Jack and Sifre, Laurent},
  booktitle={Advances in Neural Information Processing Systems},
  volume={35},
  pages={30016--30030},
  year={2022},
  url={https://papers.neurips.cc/paper_files/paper/2022/hash/c1e2faff6f588870935f114ebe04a3e5-Abstract-Conference.html}
}

@misc{austin2025scalingbook,
  title={How to Scale Your Model: A Systems View of {LLM}s on {TPU}s},
  author={Austin, Jacob and Douglas, Sholto and Frostig, Roy and Levskaya, Anselm and Chen, Charlie and Vikram, Sharad and Lebron, Federico and Choy, Peter and Ramasesh, Vinay and Webson, Albert and Pope, Reiner},
  year={2025},
  howpublished={\url{https://jax-ml.github.io/scaling-book/}},
  note={Online book}
}

@inproceedings{paquette2024phases,
  title={{4+3} Phases of Compute-Optimal Neural Scaling Laws},
  author={Paquette, Elliot and Paquette, Courtney and Xiao, Lechao and Pennington, Jeffrey},
  booktitle={Advances in Neural Information Processing Systems},
  volume={37},
  pages={16459--16537},
  year={2024},
  url={https://proceedings.neurips.cc/paper_files/paper/2024/hash/1dccfc3ee01871d05e33457c61037d59-Abstract-Conference.html}
}

@article{hu2024minicpm,
  title={{MiniCPM}: Unveiling the Potential of Small Language Models with Scalable Training Strategies},
  author={Hu, Shengding and Tu, Yuge and Han, Xu and He, Chaoqun and Cui, Ganqu and Long, Xiang and Zheng, Zhi and Fang, Yewei and Huang, Yuxiang and Zhao, Weilin and Zhang, Xinrong and Thai, Zheng Leng and Zhang, Kaihuo and Wang, Chongyi and Yao, Yuan and Zhao, Chenyang and Zhou, Jie and Cai, Jie and Zhai, Zhongwu and Ding, Ning and Jia, Chao and Zeng, Guoyang and Li, Dahai and Liu, Zhiyuan and Sun, Maosong},
  journal={arXiv preprint arXiv:2404.06395},
  year={2024},
  url={https://arxiv.org/abs/2404.06395}
}

@article{hagele2024scaling,
  title={Scaling Laws and Compute-Optimal Training Beyond Fixed Training Durations},
  author={H{\"a}gele, Alexander and Bakouch, Elie and Kosson, Atli and Ben Allal, Loubna and von Werra, Leandro and Jaggi, Martin},
  journal={arXiv preprint arXiv:2405.18392},
  year={2024},
  url={https://arxiv.org/abs/2405.18392}
}

@inproceedings{sardana2024beyond,
  title={Beyond Chinchilla-Optimal: Accounting for Inference in Language Model Scaling Laws},
  author={Sardana, Nikhil and Portes, Jacob and Doubov, Sasha and Frankle, Jonathan},
  booktitle={Proceedings of the 41st International Conference on Machine Learning},
  year={2024},
  url={https://arxiv.org/abs/2401.00448}
}

@article{muennighoff2023data,
  title={Scaling Data-Constrained Language Models},
  author={Muennighoff, Niklas and Rush, Alexander M. and Barak, Boaz and Le Scao, Teven and Piktus, Aleksandra and Tazi, Nouamane and Pyysalo, Sampo and Wolf, Thomas and Raffel, Colin},
  journal={arXiv preprint arXiv:2305.16264},
  year={2023},
  url={https://arxiv.org/abs/2305.16264}
}

@article{volkova2026robust,
  title={Towards Robust Scaling Laws for Optimizers},
  author={Volkova, Alexandra and Safaryan, Mher and Lampert, Christoph H. and Alistarh, Dan},
  journal={arXiv preprint arXiv:2602.07712},
  year={2026},
  url={https://arxiv.org/abs/2602.07712}
}

@inproceedings{yang2022tensor,
  title={Tensor Programs {V}: Tuning Large Neural Networks via Zero-Shot Hyperparameter Transfer},
  author={Yang, Greg and Hu, Edward J. and Babuschkin, Igor and Sidor, Szymon and Liu, Xiaodong and Farhi, David and Ryder, Nick and Pachocki, Jakub and Chen, Weizhu and Gao, Jianfeng},
  booktitle={Advances in Neural Information Processing Systems},
  volume={34},
  year={2021},
  url={https://arxiv.org/abs/2203.03466}
}

@inproceedings{gupta2018shampoo,
  title={Shampoo: Preconditioned Stochastic Tensor Optimization},
  author={Gupta, Vineet and Koren, Tomer and Singer, Yoram},
  booktitle={Proceedings of the 35th International Conference on Machine Learning},
  year={2018},
  url={https://proceedings.mlr.press/v80/gupta18a.html}
}

@inproceedings{loshchilov2017sgdr,
  title={{SGDR}: Stochastic Gradient Descent with Warm Restarts},
  author={Loshchilov, Ilya and Hutter, Frank},
  booktitle={International Conference on Learning Representations},
  year={2017},
  url={https://openreview.net/forum?id=Skq89Scxx}
}

@inproceedings{bergsma2025straight,
  title={Straight to Zero: Why Linearly Decaying the Learning Rate to Zero Works Best for {LLMs}},
  author={Bergsma, Shane and Dey, Nolan and Gosal, Gurpreet and Gray, Gavia and Soboleva, Daria and Hestness, Joel},
  booktitle={International Conference on Learning Representations},
  year={2025},
  url={https://arxiv.org/abs/2502.15938}
}

@inproceedings{bergsma2025powerlines,
  title={Power Lines: Scaling Laws for Weight Decay and Batch Size in {LLM} Pre-training},
  author={Bergsma, Shane and Dey, Nolan and Gosal, Gurpreet and Gray, Gavia and Soboleva, Daria and Hestness, Joel},
  booktitle={Advances in Neural Information Processing Systems},
  volume={38},
  year={2025},
  doi={10.52202/085713-4171},
  url={https://proceedings.neurips.cc/paper_files/paper/2025/hash/b5f78a17a94da3e34c935515d1b6adae-Abstract-Conference.html}
}

@misc{marek2025smallbatch,
  title={Small Batch Size Training for Language Models: When Vanilla {SGD} Works, and Why Gradient Accumulation Is Wasteful},
  author={Marek, Martin and Lotfi, Sanae and Somasundaram, Aditya and Wilson, Andrew Gordon and Goldblum, Micah},
  year={2025},
  eprint={2507.07101},
  archivePrefix={arXiv},
  primaryClass={cs.LG},
  url={https://arxiv.org/abs/2507.07101}
}

@inproceedings{saxe2014exact,
  title={Exact Solutions to the Nonlinear Dynamics of Learning in Deep Linear Neural Networks},
  author={Saxe, Andrew M. and McClelland, James L. and Ganguli, Surya},
  booktitle={International Conference on Learning Representations},
  year={2014},
  url={https://arxiv.org/abs/1312.6120}
}

@inproceedings{bordelon2020spectrum,
  title={Spectrum Dependent Learning Curves in Kernel Regression and Wide Neural Networks},
  author={Bordelon, Blake and Canatar, Abdulkadir and Pehlevan, Cengiz},
  booktitle={Proceedings of the 37th International Conference on Machine Learning},
  series={Proceedings of Machine Learning Research},
  volume={119},
  pages={1024--1034},
  year={2020},
  publisher={PMLR},
  url={https://proceedings.mlr.press/v119/bordelon20a.html}
}

@misc{apertus2025,
  title={Apertus: Democratizing Open and Compliant {LLM}s for Global Language Environments},
  author={{Project Apertus} and others},
  year={2025},
  eprint={2509.14233},
  archivePrefix={arXiv},
  primaryClass={cs.CL},
  url={https://arxiv.org/abs/2509.14233}
}

@inproceedings{defazio2024road,
  title={The Road Less Scheduled},
  author={Defazio, Aaron and Yang, Xingyu and Mehta, Harsh and Mishchenko, Konstantin and Khaled, Ahmed and Cutkosky, Ashok},
  booktitle={Advances in Neural Information Processing Systems},
  year={2024},
  url={https://arxiv.org/abs/2405.15682}
}

@misc{jax2018github,
  title={{JAX}: Composable Transformations of {P}ython+{N}um{P}y Programs},
  author={Bradbury, James and Frostig, Roy and Hawkins, Peter and Johnson, Matthew James and Katariya, Yash and Leary, Chris and Maclaurin, Dougal and Necula, George and Paszke, Adam and VanderPlas, Jake and Wanderman-Milne, Skye and Zhang, Qiao},
  year={2018},
  howpublished={\url{https://github.com/jax-ml/jax}},
  note={Version 0.6.2}
}

@misc{flax2020github,
  title={{Flax}: A Neural Network Library and Ecosystem for {JAX}},
  author={Heek, Jonathan and Levskaya, Anselm and Oliver, Avital and Ritter, Marvin and Rondepierre, Bertrand and Steiner, Andreas and van Zee, Marc},
  year={2024},
  howpublished={\url{https://github.com/google/flax}},
  note={Version 0.10.6}
}

@misc{jaxsplash2023,
  title={Splash Attention: Sparse Flash Attention for {TPU}},
  author={{The JAX Authors}},
  year={2023},
  howpublished={\url{https://github.com/jax-ml/jax}},
  note={JAX Pallas Splash Attention implementation}
}

@inproceedings{jouppi2017tpu,
  title={In-Datacenter Performance Analysis of a Tensor Processing Unit},
  author={Jouppi, Norman P. and others},
  booktitle={Proceedings of the 44th Annual International Symposium on Computer Architecture},
  pages={1--12},
  year={2017},
  doi={10.1145/3079856.3080246}
}

@article{chen2016sublinear,
  title={Training Deep Nets with Sublinear Memory Cost},
  author={Chen, Tianqi and Xu, Bing and Zhang, Chiyuan and Guestrin, Carlos},
  journal={arXiv preprint arXiv:1604.06174},
  year={2016},
  url={https://arxiv.org/abs/1604.06174}
}

@inproceedings{rajbhandari2020zero,
  title={{ZeRO}: Memory Optimizations Toward Training Trillion Parameter Models},
  author={Rajbhandari, Samyam and Rasley, Jeff and Ruwase, Olatunji and He, Yuxiong},
  booktitle={Proceedings of the International Conference for High Performance Computing, Networking, Storage and Analysis},
  year={2020},
  doi={10.1109/SC41405.2020.00024}
}

@article{videau2026skaling,
  title={Skaling: Chinchilla's Exponents Meet Kaplan's Coupling},
  author={Videau, Mathurin and Youbi-Idrissi, Badr and Lopez-Paz, David and Ahuja, Kartik},
  journal={arXiv preprint arXiv:2608.07222},
  year={2026},
  doi={10.48550/arXiv.2608.07222},
  url={https://arxiv.org/abs/2608.07222}
}

@article{ziyin2020laprop,
  title={{LaProp}: Separating Momentum and Adaptivity in {Adam}},
  author={Liu, Ziyin and Wang, Zhikang T. and Ueda, Masahito},
  journal={arXiv preprint arXiv:2002.04839},
  year={2020},
  doi={10.48550/arXiv.2002.04839},
  url={https://arxiv.org/abs/2002.04839}
}

@article{gadre2024overtraining,
  title={Language Models Scale Reliably with Over-Training and on Downstream Tasks},
  author={Gadre, Samir Yitzhak and Smyrnis, Georgios and Shankar, Vaishaal and Gururangan, Suchin and Wortsman, Mitchell and Shao, Rulin and Mercat, Jean and Fang, Alex and Li, Jeffrey and Keh, Sedrick and Xin, Rui and Nezhurina, Marianna and Vasiljevic, Igor and Jitsev, Jenia and Soldaini, Luca and Dimakis, Alexandros G. and Ilharco, Gabriel and Koh, Pang Wei and Song, Shuran and Kollar, Thomas and Carmon, Yair and Dave, Achal and Heckel, Reinhard and Muennighoff, Niklas and Schmidt, Ludwig},
  journal={arXiv preprint arXiv:2403.08540},
  year={2024},
  url={https://arxiv.org/abs/2403.08540}
}

@inproceedings{bian2025inference,
  title={Scaling Inference-Efficient Language Models},
  author={Bian, Song and Yan, Minghao and Venkataraman, Shivaram},
  booktitle={Proceedings of the 42nd International Conference on Machine Learning},
  year={2025},
  url={https://arxiv.org/abs/2501.18107}
}

@inproceedings{bjorck2025tokenhorizons,
  title={Scaling Optimal {LR} Across Token Horizons},
  author={Bj{"o}rck, Johan and Benhaim, Alon and Chaudhary, Vishrav and Wei, Furu and Song, Xia},
  booktitle={International Conference on Learning Representations},
  year={2025},
  url={https://openreview.net/forum?id=WYL4eFLcxG}
}

@article{dey2025completep,
  title={Don't Be Lazy: {CompleteP} Enables Compute-Efficient Deep Transformers},
  author={Dey, Nolan and Zhang, Bin Claire and Noci, Lorenzo and Li, Mufan and Bordelon, Blake and Bergsma, Shane and Pehlevan, Cengiz and Hanin, Boris and Hestness, Joel},
  journal={arXiv preprint arXiv:2505.01618},
  year={2025},
  url={https://arxiv.org/abs/2505.01618}
}

@article{mlodozeniec2025completedp,
  title={Completed Hyperparameter Transfer across Modules, Width, Depth, Batch and Duration},
  author={Mlodozeniec, Bruno and Ablin, Pierre and B{\'e}thune, Louis and Busbridge, Dan and Klein, Michal and Ramapuram, Jason and Cuturi, Marco},
  journal={arXiv preprint arXiv:2512.22382},
  year={2025},
  url={https://arxiv.org/abs/2512.22382}
}

@article{shulgin2026hyperparameter,
  title={Deriving Hyperparameter Scaling Laws via Modern Optimization Theory},
  author={Shulgin, Egor and von R{"u}tte, Dimitri and Zhang, Tianyue H. and Ajroldi, Niccol{\`o} and Sch{"o}lkopf, Bernhard and Orvieto, Antonio},
  journal={arXiv preprint arXiv:2603.15958},
  year={2026},
  url={https://arxiv.org/abs/2603.15958}
}

@article{khona2026soapmuon,
  title={{SOAP}, {Muon}, and Beyond: Pushing {LLM} Pretraining Scales},
  author={Khona, Mikail and Vavre, Aditya and Wang, Boxiang and Fu, Deyu and Wu, Hao and Chrzanowski, Mike and Catanzaro, Bryan and Mudigere, Dheevatsa and Pool, Jeff and Lightstone, Michael and Shoeybi, Mohammad and Patwary, Mostofa and Tajbakhsh, Nima and Blankevoort, Tijmen},
  journal={arXiv preprint arXiv:2607.20548},
  year={2026},
  url={https://arxiv.org/abs/2607.20548}
}

@article{morwani2024shampoo,
  title={A New Perspective on Shampoo's Preconditioner},
  author={Morwani, Depen and Shapira, Itai and Vyas, Nikhil and Malach, Eran and Kakade, Sham and Janson, Lucas},
  journal={arXiv preprint arXiv:2406.17748},
  year={2024},
  url={https://arxiv.org/abs/2406.17748}
}

@article{anil2020scalable,
  title={Scalable Second Order Optimization for Deep Learning},
  author={Anil, Rohan and Gupta, Vineet and Koren, Tomer and Regan, Kevin and Singer, Yoram},
  journal={arXiv preprint arXiv:2002.09018},
  year={2020},
  url={https://arxiv.org/abs/2002.09018}
}

@article{shi2023distributed,
  title={A Distributed Data-Parallel {PyTorch} Implementation of the Distributed Shampoo Optimizer for Training Neural Networks At-Scale},
  author={Shi, Hao-Jun Michael and Lee, Tsung-Hsien and Iwasaki, Shintaro and Gallego-Posada, Jose and Li, Zhijing and Rangadurai, Kaushik and Mudigere, Dheevatsa and Rabbat, Michael},
  journal={arXiv preprint arXiv:2309.06497},
  year={2023},
  url={https://arxiv.org/abs/2309.06497}
}

@article{liu2025muon,
  title={{Muon} Is Scalable for {LLM} Training},
  author={Liu, Jingyuan and Su, Jianlin and Yao, Xingcheng and Jiang, Zhejun and Lai, Guokun and Du, Yulun and Qin, Yidao and Xu, Weixin and Lu, Enzhe and Yan, Junjie and Chen, Yanru and Zheng, Huabin and Liu, Yibo and Liu, Shaowei and Yin, Bohong and He, Weiran and Zhu, Han and Wang, Yuzhi and Wang, Jianzhou and Dong, Mengnan and Zhang, Zheng and Kang, Yongsheng and Zhang, Hao and Xu, Xinran and Zhang, Yutao and Wu, Yuxin and Zhou, Xinyu and Yang, Zhilin},
  journal={arXiv preprint arXiv:2502.16982},
  year={2025},
  url={https://arxiv.org/abs/2502.16982}
}

@article{shah2025muon,
  title={Practical Efficiency of {Muon} for Pretraining},
  author={Shah, Ishaan and Polloreno, Anthony M. and Stratos, Karl and Monk, Philip and Chaluvaraju, Adarsh and Hojel, Andrew and Ma, Andrew and Thomas, Anil and Tanwer, Ashish and Shah, Darsh J. and Nguyen, Khoi and Smith, Kurt and Callahan, Michael and Pust, Michael and Parmar, Mohit and Rushton, Peter and Mazarakis, Platon and Kapila, Ritvik and Srivastava, Saurabh and Singla, Somanshu and Romanski, Tim and Vanjani, Yash and Vaswani, Ashish},
  journal={arXiv preprint arXiv:2505.02222},
  year={2025},
  url={https://arxiv.org/abs/2505.02222}
}

@inproceedings{bergsma2025collapse,
  title={Scaling with Collapse: Efficient and Predictable Training of {LLM} Families},
  author={Bergsma, Shane and Zhang, Bin Claire and Dey, Nolan and Muhammad, Shaheer and Gosal, Gurpreet and Hestness, Joel},
  booktitle={International Conference on Learning Representations},
  year={2026},
  url={https://arxiv.org/abs/2509.25087}
}

@inproceedings{bergsma2025trec,
  title={Predicting Training Re-evaluation Curves Enables Effective Data Curriculums for {LLM}s},
  author={Bergsma, Shane and Dey, Nolan and Hestness, Joel},
  booktitle={International Conference on Learning Representations},
  year={2026},
  url={https://arxiv.org/abs/2509.25380}
}
\bibliographystyle{iclr2026_conference}

\clearpage
\appendix

\section{Extended Related Work}
\label{app:extended-related-work}

\paragraph{Training beyond compute-optimal horizons.}
Data-constrained scaling studies characterize how repeated data changes the
benefit of continued training when the available unique data are limited
\citep{muennighoff2023data}. Warmup-Stable-Decay schedules separate an
extensible stable-training phase from terminal learning rate decay
\citep{hu2024minicpm}, while reusable training trajectories allow scaling
behavior to be evaluated across multiple training durations without restarting
each run from initialization \citep{hagele2024scaling}. Together, these works
treat training horizon as an experimental axis distinct from model size.

\paragraph{Hyperparameter transfer across scale.}
The maximal-update parameterization and $\mu$Transfer enable zero-shot
hyperparameter transfer across model width by prescribing how initialization
and learning rates should scale \citep{yang2022tensor}. Subsequent work derives
optimizer-dependent width exponents, develops depth-aware parameterizations,
and extends hyperparameter transfer across modules, width, depth, batch size,
and training duration
\citep{everett2024scaling,dey2025completep,mlodozeniec2025completedp}.
For matrix-preconditioned optimizers, preserving gains across model scale
requires optimizer-appropriate learning rate and weight decay transfer
\citep{qiu2025hyperparameter}. Complementary theoretical work derives coupled
scaling rules for learning rate, momentum, and batch size from the iteration or
token budget \citep{shulgin2026hyperparameter}.

\paragraph{Optimizer memory and weight decay timescales.}
Recent work uses normalized optimizer timescales to align training curves
across model scales and characterize how information from different portions
of training remains influential
\citep{bergsma2025collapse,bergsma2025trec}. In particular,
\citet{bergsma2025powerlines} fit a normalized AdamW weight decay timescale
proportional to $(D/N)^{-0.527}$, where $D$ is the token horizon and $N$ is
model size. At fixed model size, this implies that the corresponding memory
window measured in tokens grows approximately as $D^{0.473}$. This provides
additional context for studying momentum and weight decay as horizon-dependent
memory timescales.

\paragraph{Preconditioner order and scalable matrix optimizers.}
Adam accumulates momentum before applying coordinatewise adaptive
normalization, whereas LaProp applies normalization before momentum
accumulation \citep{kingma2015adam,ziyin2020laprop}. This distinction motivates
our experiments on preconditioner order. Shampoo introduced structured
matrix-valued preconditioning, with subsequent work developing scalable and
distributed implementations
\citep{gupta2018shampoo,anil2020scalable,shi2023distributed,
morwani2024shampoo}. SOAP applies Adam-like updates in Shampoo's evolving
eigenbasis, while Muon approximately orthogonalizes matrix-valued updates
\citep{vyas2025soap,jordan2024muon}. Recent studies further examine the
practical implementation and large-scale behavior of these optimizers
\citep{liu2025muon,shah2025muon,khona2026soapmuon}.

\clearpage
\section{Experimental details}
\label{app:full-protocol}

Unless explicitly stated otherwise, all experiments use the model family,
initialization, data, training, and evaluation protocols defined in this
section.

\subsection{Notation, architecture, and model sizes}

{Table~\ref{tab:notation} defines the notation used for the model family and
for nominal compute accounting.}

\begin{table}[h]
  \centering
  \small
  \setlength{\tabcolsep}{6pt}
  \renewcommand{\arraystretch}{1.08}
  {
  \begin{tabular}{@{}lp{0.72\linewidth}@{}}
    \toprule
    Symbol & Definition \\
    \midrule
    $H$ & Number of attention heads \\
    $D$ & Model dimension \\
    $D_{\mathrm{head}}$ & Dimension of each attention head \\
    $N$ & Number of Transformer blocks \\
    $F$ & Feed-forward intermediate dimension \\
    $L$ & Sequence length in tokens \\
    $V$ & Padded vocabulary size \\
    $B$ & Global batch size in sequences \\
    $S$ & Number of optimizer updates \\
    $T=SBL$ & Number of training tokens \\
    $P$ & Nominal parameter count used for scaling and compute accounting \\
    $P_{\mathrm{nonemb}}$ & Number of non-embedding parameters \\
    $P_{\mathrm{train}}$ & Exact trainable parameter count, including both vocabulary matrices and learned normalization scales \\
    $C=6PT$ & Nominal training compute in FLOPs \\
    $\mathrm{TPP}=T/P$ & Training tokens per nominal parameter \\
    $f=T/(20P)=\mathrm{TPP}/20$ & OT factor: training-token horizon in multiples of 20 tokens per nominal parameter \\
    \bottomrule
  \end{tabular}}
  \caption{{Notation used for the architecture, training budget, and nominal compute.}}
  \label{tab:notation}
\end{table}

{We use decoder-only causal Transformers.  The architecture, initialization,
and width, depth, and attention co-scaling setup broadly follow the Enoki
ladder in \citet{ferbach2026adana}.  Relative to that setup, we replace GeLU
with SwiGLU, LayerNorm with RMSNorm, and QK-LayerNorm with QK-RMSNorm, and use
a constant token embedding initialization.  The resulting Transformer blocks
are architecturally similar to the Llama 2 style architecture
\citep{touvron2023llama2} used in \citet{qiu2025hyperparameter}.

The co-scaling rule, introduced in \citet{charles2025communication} and used
in \citet{ferbach2026adana}, is
\[
D_{\mathrm{head}}=64,\qquad D=64H,\qquad
N=\lfloor3H/4\rfloor,\qquad F=4D,
\]}

\begin{table}[tp]
  \centering
  \small
  \setlength{\tabcolsep}{6pt}
  \renewcommand{\arraystretch}{1.12}
  {
  \begin{tabular}{@{}p{0.28\linewidth}p{0.64\linewidth}@{}}
    \toprule
    Component & Choice \\
    \midrule
    Block normalization & Pre-RMSNorm; learned scale, no bias; $\epsilon=10^{-6}$ \\
    Attention & Multi-head attention with $D_{\mathrm{head}}=64$ \\
    Q/K normalization & QK-RMSNorm before RoPE; learned scale, no bias; $\epsilon=10^{-6}$ \\
    Positional encoding & RoPE with base $10{,}000$ \\
    Activation & SwiGLU \\
    Tokenizer & GPT-2 with vocabulary padded to $V=50{,}304$ \\
    Embedding / readout tying & Untied \\
    Biases & None \\
    Dropout & None \\
    Nominal parameter count & $P=16ND^2+DV$ \\
    \bottomrule
  \end{tabular}}
  \caption{{Model architecture.}}
  \label{tab:architecture}
\end{table}

{The nominal count includes the readout matrix but omits the input embedding matrix, although the two are untied.  \citet{kaplan2020scaling} found that excluding embedding parameters produced cleaner scaling trends across model depths.  We nevertheless include the readout because it is applied as a dense vocabulary projection, whereas the input embedding is accessed by lookup.}

\begin{table}[tp]
  \centering
  \small
  \setlength{\tabcolsep}{2pt}
  {
  \begin{tabular}{@{}lrrrrrrr@{}}
    \toprule
    Model & $D$ & $N$ & $F$ & $P_{\rm nonemb}$ & $P$ & $P_{\rm train}$ & $S_{1\times}$ \\
    \midrule
    51M  & 512  & 6  & 2,048 & 25,173,248 & 50,921,472 & 76,684,544 & 1,942 \\
    124M & 768  & 9  & 3,072 & 84,950,400 & 123,568,128 & 162,217,344 & 4,713 \\
    253M & 1,024 & 12 & 4,096 & 201,353,728 & 252,837,888 & 304,376,320 & 9,645 \\
    458M & 1,280 & 15 & 5,120 & 393,257,600 & 457,605,120 & 522,035,840 & 17,456 \\
    757M & 1,536 & 18 & 6,144 & 679,536,384 & 756,744,192 & 834,070,272 & 28,867 \\
    1.17B & 1,792 & 21 & 7,168 & 1,079,064,448 & 1,169,129,472 & 1,259,353,984 & 44,598 \\
    \bottomrule
  \end{tabular}}
  \caption{{Model sizes. Model names use rounded nominal parameter
  counts $P=16ND^2+DV$. Non-embedding and exact trainable parameter counts
  include learned normalization scales.}}
  \label{tab:model-ladder}
\end{table}

{Throughout the paper, model size labels such as 51M and 253M refer to
rounded nominal parameter counts $P$, not exact trainable parameter counts.}

{We use overtraining (OT) factor as a concise convention for reporting training
horizons: $1\times$ OT denotes 20 training tokens per nominal parameter,
$2\times$ denotes 40, and so forth.  The 20-tokens-per-parameter heuristic is
motivated by \citet{hoffmann2022empirical}, but we do not assume that this
horizon is compute-optimal for every optimizer, model family or dataset.  For
each model, the $1\times$ horizon uses
$S_{1\times}=\lfloor20P/(BL)\rfloor$ optimizer updates.  We report the
realized token count after this integer rounding.}

{Following \citet{hoffmann2022empirical} and
\citet{austin2025scalingbook}, we approximate training compute as
\[
C=6PT,
\]
where $P$ is the nominal parameter count and $T$ is the number of training
tokens.  This first-order estimate counts approximately $2PT$ FLOPs for the
forward pass and $4PT$ FLOPs for the backward pass through the parameterized
matrix multiplications.}

\subsection{Initialization}

{Following the ScaledGPT initialization used in \citet{ferbach2026adana},
ordinary projections use fan-in scaling, while attention and feed-forward
residual-output projections receive an additional $1/\sqrt{2N}$ depth factor.
This depth correction follows the initialization used in GPT-2
\citep{radford2019language}.  Because residual-output projections are
initialized independently with zero-mean weights, distinct residual increments
have zero cross-covariance in expectation over initialization.  Scaling each of
the $2N$ residual-output projections by $1/\sqrt{2N}$ therefore keeps the
expected aggregate variance contributed by the residual branches of order one
as depth changes.  We initialize token embeddings separately with the
width-independent constant standard deviation $0.02$.  Residual branches are
added to the residual stream with coefficient one; no additional residual
multiplier is applied.}

{All matrix entries are drawn independently from untruncated, zero-mean Gaussian
distributions.  Table~\ref{tab:initialization} reports their standard
deviations; learned normalization scales are initialized to one.}

\begin{table}[tp]
  \centering
  \small
  \setlength{\tabcolsep}{8pt}
  \renewcommand{\arraystretch}{1.10}
  {
  \begin{tabular}{@{}p{0.58\linewidth}p{0.26\linewidth}@{}}
    \toprule
    Parameter & Standard deviation \\
    \midrule
    Token embedding & $0.02$ \\
    Query, key, and value projections & $D^{-1/2}$ \\
    Attention output projection & $(2ND)^{-1/2}$ \\
    Feed-forward input projection & $D^{-1/2}$ \\
    SwiGLU gate projection & $D^{-1/2}$ \\
    Feed-forward output projection & $(2NF)^{-1/2}$ \\
    Readout matrix & $D^{-1/2}$ \\
    Learned normalization scales & $1$ \\
    \bottomrule
  \end{tabular}}
  \caption{{Initialization standard deviations.}}
  \label{tab:initialization}
\end{table}

\subsection{Common training controls}

{Table~\ref{tab:training-controls} lists the training controls shared across
the primary experiments unless explicitly stated otherwise.  For an
experiment with $S$ optimizer updates and tuned peak learning rate $\eta$,
every candidate learning rate schedule uses the same linear warmup.}

\begin{table}[tp]
  \centering
  \small
  \setlength{\tabcolsep}{8pt}
  \renewcommand{\arraystretch}{1.10}
  {
  \begin{tabular}{@{}p{0.30\linewidth}p{0.62\linewidth}@{}}
    \toprule
    Control & Default \\
    \midrule
    Sequence length & $L=2{,}048$ tokens \\
    Global batch size & $B=256$ sequences \\
    Gradient clipping & Global $\ell_2$ norm of $1.0$, applied after gradient accumulation \\
    Learning rate warmup & Linear from $0.01\eta$ to $\eta$ over $W=\max\{100,\min\{\lfloor S/50\rfloor,5{,}000\}\}$ updates \\
    \bottomrule
  \end{tabular}}
  \caption{{Common training controls.}}
  \label{tab:training-controls}
\end{table}

{The learning rate schedule after warmup and the weight decay treatment are
selected experimentally in Appendices~\ref{app:learning-rate-schedule}
and~\ref{app:weight-decay}, respectively.}

\subsection{Dataset, training order, seeds, and evaluation}

{We use the \texttt{sample/350BT} subset of FineWeb
\citep{penedo2024fineweb} at revision
\texttt{9bb295ddab0e05d785b879661af7260fed5140fc}.  We encode each
document with the GPT-2 tokenizer~\citep{radford2019language} and append one
GPT-2 end-of-text token.  We use source files 0--507 for training and reserve
source files 508--509 for validation.  The resulting training set contains
360.5 billion tokens.  Training and validation are drawn from disjoint
FineWeb source files; we do not perform additional deduplication across the
two partitions.}

{Training follows a fixed sequential order through the tokenized corpus.
End-of-text tokens separate documents, and training sequences may span
document boundaries.  We do not shuffle the training data.  All experiments
begin at the start of the corpus and use this same ordering.  Consequently,
experiments with the same global batch size and training horizon process the
same tokens in the same optimizer updates.}

{Unless explicitly stated otherwise, every hyperparameter configuration starts
from a fresh initialization with random seed 42.  The training order is fixed
independently of the seed, so configurations at the same model and training
coordinate share both their initialization and their ordered training data.
We use one seed per configuration for the hyperparameter sweeps in this paper.
These matched comparisons reduce nuisance variation between nearby
configurations, but they do not estimate variation across random
initializations.}

{From the two held-out source files, we construct one fixed validation set
containing 1,920 sequences and 3,932,160 scored next-token targets.  Every
model is evaluated on the same targets in the same order.  We report mean
per-token next-token cross-entropy over the complete validation set.  This
full-set loss is used for both hyperparameter selection and reported loss
comparisons.}

% INTERNAL: Historical evaluation eligibility and attention-backend eras are
% tracked in audits/optimizer_spec/OT_PAPER_INTEGRATION_BOARD.md; they are not
% part of the reader-facing protocol description.

\subsection{Hyperparameter sweep interiority}
\label{app:sweep-interiority}

Throughout the paper, we assess hyperparameter sweeps using the same
boundary-interiority criterion.  We regard a one-dimensional sweep as
interior only when its minimum and the complete set of points within one
millipoint of that minimum lie away from both tested boundaries.  For nested
sweeps, we assess the inner learning rate sweep independently at every value
of the outer hyperparameter before interpreting regret or selecting a minimum
along the outer axis.  We then apply the same one-millipoint interiority
criterion to the outer axis.

\subsection{Hardware, numerical precision, and parallel execution}

{We implement the models in JAX~\citep{jax2018github} and
Flax~\citep{flax2020github} and train on TPU v4-8 and v6e-4
accelerators~\citep{jouppi2017tpu} using causal Splash attention with
512-token kernel tiles and a fused backward pass.}

{We use mixed-precision training.  Model parameters and floating-point
optimizer states are stored in \texttt{float32}; parameter gradients and
RMSNorm statistics are computed in \texttt{float32}; and model activations
and the operands of dense and attention matrix multiplications generally use
\texttt{bfloat16}.  The JAX Splash implementation uses \texttt{float32} for
selected attention accumulations~\citep{jaxsplash2023}.  The training loss
materializes the vocabulary logits and computes full-vocabulary cross-entropy.
For validation, logits are promoted from \texttt{bfloat16} to
\texttt{float32} before computing cross-entropy and device-local loss sums;
the ordered batch sums are then combined on the host using \texttt{float64}
summation.}

{For gradient accumulation, we divide the global batch contiguously into
equal microbatches.  We average the resulting gradients, apply global-norm
clipping to the accumulated batch-mean gradient, and then perform one
optimizer update.  Gradient accumulation therefore preserves the global batch
size, data ordering, and number of optimizer updates.  It produces the same
batch-mean gradient in exact arithmetic, although the different reduction
order can introduce small finite-precision differences.}

{When activation rematerialization is enabled, we apply it independently to
each Transformer block~\citep{chen2016sublinear}.  Intermediate activations
within the block are discarded after the forward pass and recomputed during
the backward pass, reducing activation memory at the cost of additional
computation.}

{For configurations requiring parameter sharding, we use ZeRO Stage 3 fully
sharded data parallelism over a one-dimensional TPU
mesh~\citep{rajbhandari2020zero}.  Model parameters, gradients, and optimizer
states are partitioned across devices rather than replicated.}

\section{Optimizer definitions and implementation details}
\label{app:optimizer-contracts}

\subsection{Common update wrapper}

Let $g_t$ denote the gradient averaged over the global batch after combining
any accumulated microbatch gradients.  The optimizer receives the globally
clipped gradient
\begin{equation}
  \widetilde g_t
  =
  \frac{g_t}{\max\!\left(1,\lVert g_t\rVert_2/G\right)},
  \qquad G=1.
\end{equation}
We write the scheduled base learning rate as $\eta_t=\eta s_t$, where $\eta$
is the tuned peak learning rate and $s_t\in[0,1]$ is a dimensionless schedule
multiplier that includes warmup and decay.  To express layerwise learning rate
tuning without carrying separate multipliers through every optimizer
definition, we define the groupwise peak learning rate
\begin{equation}
  \eta_i = \eta r_i\gamma_i,
  \qquad
  \eta_{t,i}=s_t\eta_i,
\end{equation}
where $r_i$ is an optimizer-native route multiplier and $\gamma_i$ is the
tunable layerwise multiplier, each constant within its parameter group.  For
optimizers without distinct native routes, $r_i=1$.  The $\gamma=1$ setting
fixes $\gamma_i=1$ for every group but does not remove optimizer-native route
multipliers.

Following decoupled weight decay~\citep{loshchilov2019decoupled}, we implement
weight decay as a separate parameter update term rather than as an $L_2$
penalty in the training objective.  Given an optimizer-specific direction
$d_{t,i}$, the common parameter update is
\begin{equation}
  \theta_{t+1,i}
  =
  \theta_{t,i}
  - s_t\eta_i d_{t,i}
  - s_t\lambda_{t,i}\theta_{t,i}.
  \label{eq:common-optimizer-update}
\end{equation}
The data and decay terms are computed from $\theta_{t,i}$ and applied in the
same update.  Decay is therefore not added to the gradient, accumulated in the
optimizer moments, or transformed by an optimizer preconditioner.  It follows
the normalized schedule $s_t$, but is independent of the peak learning rate,
the native route multiplier, and the layerwise multiplier.

For the uniform WD schedule,
\begin{equation}
  \lambda_{t,i}=m_i^{\mathrm{WD}}\frac{c}{S},
\end{equation}
where $S$ is the number of optimizer updates, $c$ is the weight decay
coefficient, and $m_i^{\mathrm{WD}}\in\{0,1\}$ is the decay mask.  In the
absence of a data update, a decayed parameter is therefore multiplied at
update $t$ by $1-s_tc/S$.

\subsection{AdamW}

We use AdamW~\citep{loshchilov2019decoupled,kingma2015adam}.  Adam maintains
exponential moving averages of the first and second elementwise moments of the
clipped gradient,
\begin{align}
  m_t &= \beta_1m_{t-1}+(1-\beta_1)\widetilde g_t, \\
  v_t &= \beta_2v_{t-1}+(1-\beta_2)\widetilde g_t^{\odot 2},
\end{align}
with $m_0=v_0=0$.  We apply the standard bias corrections
\begin{equation}
  \widehat m_t=\frac{m_t}{1-\beta_1^t},
  \qquad
  \widehat v_t=\frac{v_t}{1-\beta_2^t},
\end{equation}
and define the Adam direction elementwise as
\begin{equation}
  d_t=\frac{\widehat m_t}{\sqrt{\widehat v_t}+\epsilon}.
\end{equation}
We use $\epsilon=10^{-15}$.
Substituting this direction into Equation~\ref{eq:common-optimizer-update}
gives
\begin{equation}
  \theta_{t+1,i}
  =
  \theta_{t,i}
  - s_t\eta_i
    \frac{\widehat m_{t,i}}{\sqrt{\widehat v_{t,i}}+\epsilon}
  - s_t\lambda_{t,i}\theta_{t,i}.
\end{equation}
The groupwise learning rate therefore affects the Adam data update, while
independent weight decay remains independent of the peak learning rate and the
layerwise multiplier.

\subsection{Muon}

We apply Muon~\citep{jordan2024muon} to the hidden-layer attention and
feed-forward matrices.  Muon applies Nesterov momentum followed by
Newton--Schulz matrix orthogonalization.  Following
\citet{qiu2025hyperparameter}, we use Adam for token embeddings, the readout
matrix, learned normalization scales, and biases.  The Adam route has native
learning rate multiplier $r_i=1.6$, while the Muon route has $r_i=1$.

For an eligible hidden weight matrix, let $G_{t,i}$ denote the effective
matrix obtained from its clipped gradient.  Rank-three attention kernels are
reshaped to matrices before the Muon transformation and restored to their
original shapes afterward.  Muon updates its momentum buffer and forms the
Nesterov lookahead as
\begin{align}
  B_{t,i} &= \beta B_{t-1,i}+(1-\beta)G_{t,i}, \\
  H_{t,i} &= \beta B_{t,i}+(1-\beta)G_{t,i},
\end{align}
with $B_{0,i}=0$.  Starting from
\begin{equation}
  X_0=\frac{H_{t,i}}{\lVert H_{t,i}\rVert_F+\epsilon_{\mathrm{NS}}},
  \qquad \epsilon_{\mathrm{NS}}=10^{-8},
\end{equation}
we transpose $X_0$ when necessary so that its shorter dimension comes first,
then apply five Newton--Schulz iterations,
\begin{align}
  A_j &= X_jX_j^\top, \\
  X_{j+1} &= c_0X_j+(c_1A_j+c_2A_j^2)X_j,
\end{align}
where
\begin{equation}
  (c_0,c_1,c_2)=(3.4445,-4.7750,2.0315).
\end{equation}

After restoring the original orientation and tensor shape, we apply no
additional shape-dependent multiplier, and the Muon direction is
\begin{equation}
  d_{t,i}=X_5.
\end{equation}
Parameters assigned to the Adam route use the Adam direction from the
preceding subsection.

\subsection{SOAP}

We use SOAP~\citep{vyas2025soap}, which applies Adam in the eigenbasis of a
Shampoo preconditioner.  Following \citet{wen2025fantastic}, we apply SOAP to
every learned matrix, including the token embedding, attention and
feed-forward kernels, and readout matrix.  Rank-three attention kernels are
reshaped to their effective matrix representations and restored afterward.
Learned normalization scales and biases use the Adam direction defined above.
Both routes use $r_i=1$.

For each matrix, we partition every eligible axis into blocks of at most 512
coordinates, zero-padding and subsequently cropping ragged boundary blocks.
An axis longer than 10,000 coordinates is left unpreconditioned.
Consequently, the embedding and readout receive one-sided SOAP, with an
identity basis on the vocabulary axis.  For gradient block $G_{t,b}$, SOAP
updates the left and right Gram matrices as
\begin{align}
  L_{t,b}
  &=
  \beta_{\mathrm{Sh}}L_{t-1,b}
  +(1-\beta_{\mathrm{Sh}})G_{t,b}G_{t,b}^{\top}, \\
  R_{t,b}
  &=
  \beta_{\mathrm{Sh}}R_{t-1,b}
  +(1-\beta_{\mathrm{Sh}})G_{t,b}^{\top}G_{t,b}.
\end{align}

Let $Q^L_{t,b}$ and $Q^R_{t,b}$ denote the bases used for the current update.
After an initialization-only optimizer call that constructs these bases by
eigendecomposition, index $t=1,2,\ldots$ counts applied SOAP updates.  SOAP
maintains its first moment in the original coordinates and its second moment
in the rotated coordinates:
\begin{align}
  M_{t,b}
  &=
  \beta_1M_{t-1,b}+(1-\beta_1)G_{t,b}, \\
  \overline G_{t,b}
  &=
  (Q^L_{t,b})^\top G_{t,b}Q^R_{t,b}, \\
  \overline M_{t,b}
  &=
  (Q^L_{t,b})^\top M_{t,b}Q^R_{t,b}, \\
  V_{t,b}
  &=
  \beta_2V_{t-1,b}+(1-\beta_2)\overline G_{t,b}^{\odot 2}.
\end{align}
The blockwise SOAP direction is
\begin{align}
  \overline D_{t,b}
  &=
  \frac{\sqrt{1-\beta_2^t}}{1-\beta_1^t}
  \frac{\overline M_{t,b}}{\sqrt{V_{t,b}}+\epsilon},
  \label{eq:soap-blockwise-direction} \\
  D_{t,b}
  &=
  Q^L_{t,b}\overline D_{t,b}(Q^R_{t,b})^\top.
\end{align}

We use $\epsilon=10^{-15}$ in the denominator of the blockwise SOAP direction
in Equation~\ref{eq:soap-blockwise-direction} and add no diagonal
regularization to the Gram matrices ($\epsilon_{\mathrm{matrix}}=0$).

We refresh the left and right bases after every $K=10$ applied updates.
Before each refresh, we compute the Rayleigh quotient of every current basis
vector.  For example, for a left-basis column $q_j$,
\begin{equation}
  \widehat\lambda^L_{t,b,j}
  =
  q_j^\top L_{t,b}q_j,
\end{equation}
which estimates the gradient variance along that direction.  We sort the
basis vectors in decreasing order of these estimates and apply the same
permutation to the corresponding rows or columns of $V_{t,b}$.  We then
obtain the refreshed basis from a QR decomposition of the Gram matrix applied
to the sorted basis,
\begin{equation}
  Q^L_{t+1,b}
  =
  \operatorname{qr}\!\left(
    L_{t,b}Q^{L,\mathrm{sorted}}_{t,b}
  \right),
\end{equation}
with the analogous operation on the right.  Thus the second-moment slots are
reordered by approximate spectral rank but are not otherwise transported
through the change of basis.

Finally, we merge and crop the blocks, restore the original tensor shape, and
set $d_{t,i}=D_{t,i}$.  We apply no additional spectral normalization or
shape-dependent multiplier.

\subsection{ADANA}

We use ADANA~\citep{ferbach2025dana,ferbach2026adana}, following the
released PyTorch implementation used for the language-model experiments.
Let $t=1,2,\ldots$ count applied optimizer updates; gradient-accumulation
microbatches do not advance this clock.  ADANA uses the same log-time weight
for its first and second moments,
\begin{equation}
  \Delta_t=\frac{\delta}{\delta+t},
\end{equation}
and updates
\begin{align}
  m_t
  &=
  (1-\Delta_t)m_{t-1}
  +\Delta_t\widetilde g_t, \\
  v_t
  &=
  (1-\Delta_t)v_{t-1}
  +\Delta_t\widetilde g_t^{\odot 2},
\end{align}
with $m_0=v_0=0$.  We apply no additional bias correction.

Define the time-dependent momentum amplification
\begin{equation}
  \chi_t=(t+1)^{1-\kappa}+1.
\end{equation}
The ADANA direction is
\begin{equation}
  d_t
  =
  \frac{
    g_2\widetilde g_t
    +g_3\chi_t m_t
  }{
    \sqrt{v_t}+\epsilon
  },
\end{equation}
where all operations involving $v_t$ are elementwise.  We use
$\epsilon=10^{-15}$.  We fix $g_2=1$ and $\delta=8$, while $g_3$ and $\kappa$
control the strength and growth rate of the momentum contribution and are
treated as momentum hyperparameters.  The additional $+1$ in $\chi_t$ and the
first-update clock follow the released implementation rather than the
displayed ADANA pseudocode.

Our implementation exposes the optional ADANA--MK4 coordinatewise cap through
\texttt{clipsnr}, but every ADANA experiment reported in this paper disables
this option and uses no MK4 clipping.

\subsection{Parameter routing}

Table~\ref{tab:parameter-routing} summarizes the optimizer direction,
layerwise learning rate group, and weight decay mask for each parameter class.
The three layerwise factors are $\gamma_e$ for the token embedding and other
vector-like parameters, $\gamma_h$ for hidden matrices, and $\gamma_r$ for the
readout matrix.

\begin{table}[tp]
  \centering
  \small
  \setlength{\tabcolsep}{4pt}
  \resizebox{\textwidth}{!}{%
  \begin{tabular}{lcccccc}
    \toprule
    Parameter class & AdamW & Muon & SOAP & ADANA & Learning rate group & Weight decay \\
    \midrule
    Token embedding & Adam & Adam & SOAP & ADANA & $\gamma_e$ & Yes \\
    Attention and feed-forward matrices & Adam & Muon & SOAP & ADANA & $\gamma_h$ & Yes \\
    Readout matrix & Adam & Adam & SOAP & ADANA & $\gamma_r$ & Yes \\
    Learned normalization scales & Adam & Adam & Adam & ADANA & $\gamma_e$ & No \\
    \bottomrule
  \end{tabular}}
  \caption{Parameter routing for the four primary optimizers.  The weight decay column
  gives the common independent weight decay mask.}
  \label{tab:parameter-routing}
\end{table}

All native route multipliers are $r_i=1$, except for Muon's Adam route,
which uses $r_i=1.6$.  Models in the primary architecture contain no bias
parameters.  When adapting
our methods to architectures with biases, we recommend treating them like
learned normalization scales: use the same optimizer route and the $\gamma_e$
embedding-group learning rate, and exclude them from weight decay.

% INTERNAL: Versioned implementation identities and resolved treatment
% payloads belong in experiment provenance and the optimizer audit documents,
% not in the reader-facing routing table.

\section{Learning rate schedule selection}
\label{app:learning-rate-schedule}

We next investigate how the learning rate decay schedule interacts with
training horizon and find a crossover between linear and cosine decay across
OT factors.  For each schedule, we jointly tune the base learning rate and
weight decay coefficient at each OT factor.  Linear decay performs better at
$1\times$ OT for all four optimizers, whereas cosine decay performs better at
$8\times$ and $32\times$ OT.  A learning rate schedule selected at a
conventional training horizon therefore need not transfer to a substantially
overtrained regime.  Since our experiments focus primarily on high
overtraining factors, we use cosine-decay-to-zero for the remainder of the
paper.

Both schedules begin with a linear warmup from $0.01\eta$ to the peak
learning rate $\eta$ over
$W=\max\{100,\min\{\lfloor S/50\rfloor,5{,}000\}\}$ optimizer updates, where
$S$ is the total number of optimizer updates.  For the remaining $S-W$
updates, let $u_t\in[0,1]$ denote the fraction of post-warmup training
completed, and write the learning rate as $\eta_t=\eta s_t$.  We compare linear
decay to zero with cosine decay to zero~\citep{loshchilov2017sgdr},
\begin{equation}
  s_t^{\mathrm{linear}}=1-u_t,
  \qquad
  s_t^{\mathrm{cosine}}=\frac{1+\cos(\pi u_t)}{2}.
  \label{eq:zero-ending-schedules}
\end{equation}

\begin{figure}[tp]
  \centering
  \includegraphics[width=0.52\textwidth]{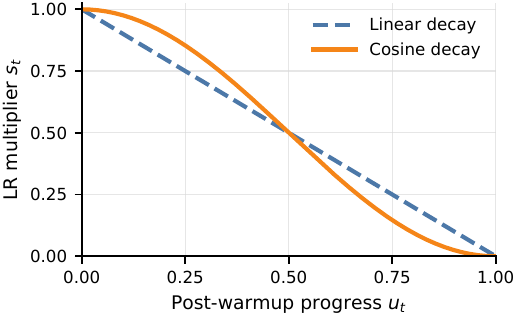}
  \caption{\textbf{Linear and cosine learning rate decay schedules after
  warmup.} Both schedules use the same linear warmup before decaying from the
  peak learning rate to zero.}
  \label{fig:lr-schedule-shapes}
\end{figure}

We compare linear and cosine decay on the 51M model at $1\times$, $8\times$,
and $32\times$ OT for AdamW, ADANA, Muon, and SOAP.  We hold the
optimizer-specific momentum hyperparameters fixed throughout this comparison:
$\beta_1=0.9$ and $\beta_2=0.95$ for AdamW; $g_3=8$ and $\kappa=0.85$ for
ADANA; $\beta=0.95$ for Muon; and $\beta_1=0.95$, $\beta_2=0.99$, and
$\beta_{\mathrm{Sh}}=0.95$ for SOAP.

To account for possible interactions between the learning rate schedule and
weight decay coefficient, we sweep the base learning rate $\eta$ and weight
decay coefficient $c$ separately for each optimizer, OT factor, and schedule.
We use the uniform WD schedule, defined here by a per-update coefficient
$c/S$ before multiplication by the learning rate schedule $s_t$.  In
particular, $s_t$ multiplies both the optimizer's data-dependent update and the
independent weight decay update, making an interaction between the learning
rate schedule shape and the weight decay coefficient plausible.  For each
schedule, we extend the learning rate sweep at every weight decay coefficient,
OT factor, and optimizer setting until the minimum is attained in the interior
of the sweep.  We likewise extend the weight decay range whenever the best
pair lies on its boundary.  We then report the lowest final validation loss
over the resulting $(\text{learning rate},\text{weight decay})$ grid for each
optimizer and OT factor.

Let $L_s^*$ denote the lowest final validation loss over the resulting
$(\text{learning rate},\text{weight decay})$ grid for schedule
$s\in\{\mathrm{linear},\mathrm{cosine}\}$.  We summarize the schedule
advantage as
\[
  10^3\left(L_{\mathrm{linear}}^*-L_{\mathrm{cosine}}^*\right)
\]
in millipoints, so positive values indicate that cosine decay achieves the
lower final validation loss.  Appendix~\ref{app:weight-decay} separately
selects the weight decay schedule and its WD coefficient scaling rule.  The
exact fixed momentum treatments, tested WD coefficients, and complete learning
rate sweeps appear in Appendix~\ref{app:complete-sweeps}.

\begin{figure}[tp]
  \centering
  \includegraphics[width=\textwidth]{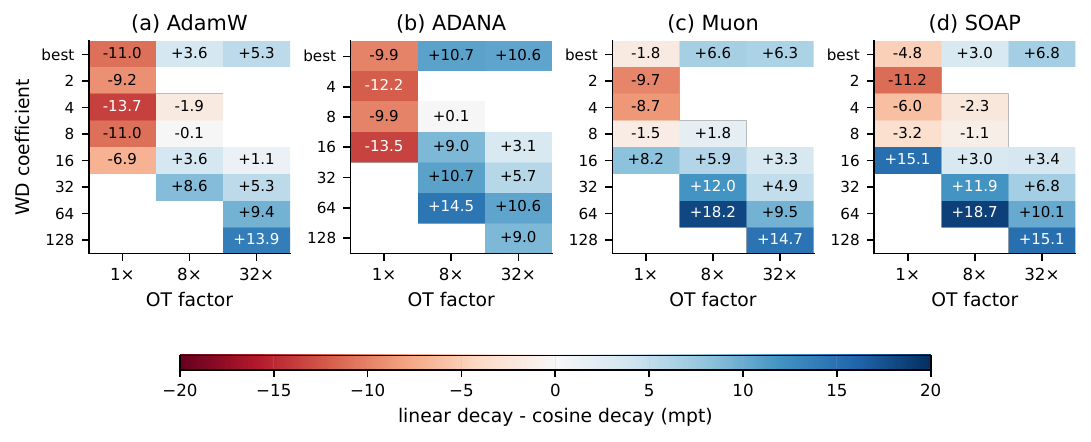}
  \caption{\textbf{Cosine decay vs linear decay learning rate schedule
  advantage across OT factor and coefficient of the uniform WD schedule on
  the 51M parameter model.} Each cell reports
  $10^3(L_{\mathrm{linear}}-L_{\mathrm{cosine}})$ after selecting the best
  base learning rate separately for each schedule at the indicated weight
  decay coefficient; positive values favor cosine decay to zero. The row
  labeled ``best'' selects the best weight decay coefficient for each OT
  factor separately for each schedule. Momentum hyperparameters are fixed to
  AdamW $(\beta_1=0.9,\beta_2=0.95)$, ADANA
  $(g_3=8,\kappa=0.85)$, Muon $(\beta=0.95)$, and SOAP
  $(\beta_1=0.95,\beta_2=0.99,\beta_{\mathrm{Sh}}=0.95)$.}
  \label{fig:lr-schedule-coefficient-resolved}
\end{figure}

\begin{figure}[tp]
  \centering
  \includegraphics[width=\textwidth]{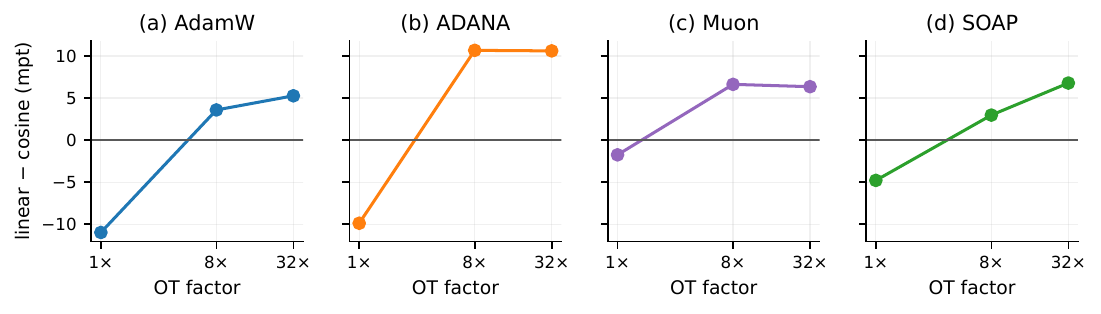}
  \caption{\textbf{High overtraining factors favor cosine decay over linear
  decay on the 51M parameter model for all optimizers.} Each point reports
  $10^3(L_{\mathrm{linear}}-L_{\mathrm{cosine}})$ after selecting the best
  base learning rate and weight decay coefficient pair for each schedule;
  positive values favor cosine decay to zero. Momentum hyperparameters are
  fixed as in Figure~\ref{fig:lr-schedule-coefficient-resolved}.}
  \label{fig:lr-schedule-profiled-advantage}
\end{figure}

Figure~\ref{fig:lr-schedule-profiled-advantage} summarizes the
horizon-dependent crossover.  After selecting the best learning rate and
weight decay coefficient pair for each schedule, linear decay achieves the
lower loss at $1\times$ OT for all four optimizers, whereas cosine decay
achieves the lower loss at $8\times$ and $32\times$ OT for all four.  Thus, a
schedule selected at $1\times$ OT does not transfer unchanged to the more
overtrained regimes studied here.

Figure~\ref{fig:lr-schedule-coefficient-resolved} shows that the crossover is
not produced by a single favorable weight decay coefficient: the high-OT
advantage of cosine decay appears across multiple coefficients.  The same
crossover remains in the row labeled ``best,'' where the weight decay
coefficient is selected separately for each schedule and OT factor.

Because the remaining experiments in this paper focus primarily on high OT
factors, we use cosine decay to zero after warmup as the primary learning rate
schedule.  This is a protocol choice for the regimes studied here, rather than
a claim that cosine decay is universally optimal.

This comparison is restricted to the 51M model, the two schedule families
considered here, fixed optimizer-specific momentum settings, and the uniform
WD schedule.  Appendix~\ref{app:weight-decay} next studies weight decay
schedules and coefficient scaling, while Appendix~\ref{app:complete-sweeps}
reports the complete learning rate sweeps underlying this comparison.

\section{Weight decay schedules and coefficient scaling}
\label{app:weight-decay}

After selecting cosine decay to zero as the learning rate schedule, we next
study two decisions governing weight decay within and across training
horizons.  The WD schedule specifies how the per-update decay coefficient
changes across optimizer updates, whereas the WD coefficient scaling rule
specifies how its overall coefficient changes with OT factor.  Such a scaling
rule is needed when we do not retune weight decay at every training horizon,
because the best WD coefficient at one OT factor does not generally remain
best at another.  We first define the two WD schedules and fix the log-time
offset convention, then compare the schedules under cosine decay to zero and
derive coefficient scaling rules for experiments where full per-horizon
tuning is infeasible.  Appendix~\ref{app:complete-sweeps} reports the complete
learning rate-by-coefficient grids underlying these analyses.

Following decoupled weight decay~\citep{loshchilov2019decoupled}, we implement
weight decay as a separate parameter update term rather than as an $L_2$
penalty in the training objective.  We apply weight decay to all learned
parameters except learned normalization scales, including both the token
embedding and readout matrices.  Our models do not contain biases, but for
architectures with learned biases, we recommend excluding them from weight
decay.  For parameters included in the weight decay mask, the independent
decay contribution at update $t$ is $-s_t\lambda_t\theta_t$, where
$s_t\in[0,1]$ denotes the normalized learning rate schedule.

We use uniform WD to denote the standard schedule in which the per-update
coefficient, before multiplication by $s_t$, is constant across optimizer
updates.  Following \citet{wang2025adamw}, we divide the WD coefficient by the
total number of optimizer updates $S$:
\begin{equation}
  \lambda_t^{\mathrm{uniform}}=\frac{c_{\mathrm{uniform}}}{S}.
  \label{eq:wd-uniform-schedule}
\end{equation}
For a fixed learning rate schedule shape, this approximately prevents
cumulative decay from increasing with the number of updates.

We use log-time WD to denote the schedule proposed in
\citet{ferbach2026adana}, in which the per-update coefficient decreases
inversely with the shifted update index:
\begin{equation}
  \lambda_t^{\log}=\frac{c_{\log}}{\tau+t}.
  \label{eq:wd-logtime-schedule}
\end{equation}
We use $c$ for the WD coefficient throughout and add a schedule subscript only
when multiple WD schedules appear in the same expression.  The coefficient
values are not directly comparable across schedules because they parameterize
different decay shapes.

\subsection{Protocol choice for the log-time WD offset}

Applying log-time WD across training horizons requires specifying how its
offset changes when the number of optimizer updates changes.  Let $S$ denote
the total number of optimizer updates in the current experiment and
$S_{1\times}$ the number at $1\times$ OT.  The schedule in
\citet{ferbach2026adana} uses an offset proportional to the current training
horizon.  We instead anchor the offset to the $1\times$ OT horizon:
\begin{equation}
  \underbrace{
    \lambda_{t,\mathrm{horizon}}^{\log}
    = \frac{c}{0.1S+t}
  }_{\text{horizon-normalized}}
  \qquad
  \underbrace{
    \lambda_{t,1\times}^{\log}
    = \frac{c}{0.1S_{1\times}+t}
  }_{\text{$1\times$-anchored, used here}}.
  \label{eq:log-wd-offsets}
\end{equation}
The two definitions coincide at $1\times$ OT.  At longer horizons, the
horizon-normalized definition stretches the initial phase of the WD schedule
with the full training horizon, whereas the $1\times$-anchored definition
preserves the initial timescale used at $1\times$ OT.  We adopt
$\tau=0.1S_{1\times}$ as our protocol convention for extending log-time WD
across the OT axis.  This choice is not intended as a claim that the anchored
offset is universally optimal.  Unless otherwise stated, all log-time WD
experiments in this paper use this convention.

\subsection{Interaction with the learning rate endpoint}

The language model experiments in \citet{ferbach2026adana} use cosine decay
to 10\% of the peak learning rate.  Comparing log-time WD with uniform WD for
AdamW and AdEMAMix, they report approximately 10\% compute-efficiency gains
that appear to increase with model scale.  The literature generally finds that
decaying the learning rate to zero outperforms schedules with higher endpoints.
For example, \citet{bergsma2025straight} find that linear decay to zero
consistently outperforms cosine decay to 10\% across a range of language model
pretraining settings.  We therefore test whether the magnitude of the log-time
WD advantage depends on the learning rate endpoint.

To isolate this interaction, we hold the base learning rate fixed at
$\log_2\eta=-8$ for AdamW and $-12.5$ for ADANA and match the total parameter
shrinkage produced by the two WD schedules in the absence of gradient
updates.  Within each optimizer and learning rate endpoint pair, all other
training coordinates are identical.  Under weight decay alone, a parameter
evolves according to
\begin{equation}
  \theta_{t+1}=(1-s_t\lambda_t)\theta_t.
\end{equation}
The fraction of its initial value retained after $S$ updates is therefore
\begin{equation}
  R(\lambda)=\prod_{t=0}^{S-1}(1-s_t\lambda_t),
  \qquad
  \theta_S=R(\lambda)\theta_0.
  \label{eq:pure-decay-retention}
\end{equation}
For a log-time WD coefficient $c_{\log}=2$, we choose the uniform WD
coefficient $c_{\mathrm{uniform}}$ by solving
\begin{equation}
  \prod_{t=0}^{S-1}
  \left(1-s_t\frac{c_{\mathrm{uniform}}}{S}\right)
  =
  \prod_{t=0}^{S-1}
  \left(1-s_t\frac{c_{\log}}{\tau+t}\right).
  \label{eq:matched-pure-decay-retention}
\end{equation}
We solve Equation~\ref{eq:matched-pure-decay-retention} numerically in log
space using the exact discrete learning rate schedule, including warmup.  This
gives $c_{\mathrm{uniform}}=11.500447$ for cosine decay to 10\% and
$c_{\mathrm{uniform}}=12.319252$ for cosine decay to zero.  The matching
depends only on the learning rate schedule and training horizon, so we use the
same coefficients for AdamW and ADANA.  This comparison therefore isolates the
effect of when weight decay is applied while holding its cumulative parameter
shrinkage fixed.  Because the coefficients are determined by retention
matching rather than tuned separately, it does not compare the best achievable
loss under the two WD schedules.

\begin{figure}[tp]
  \centering
  \includegraphics[width=\textwidth]{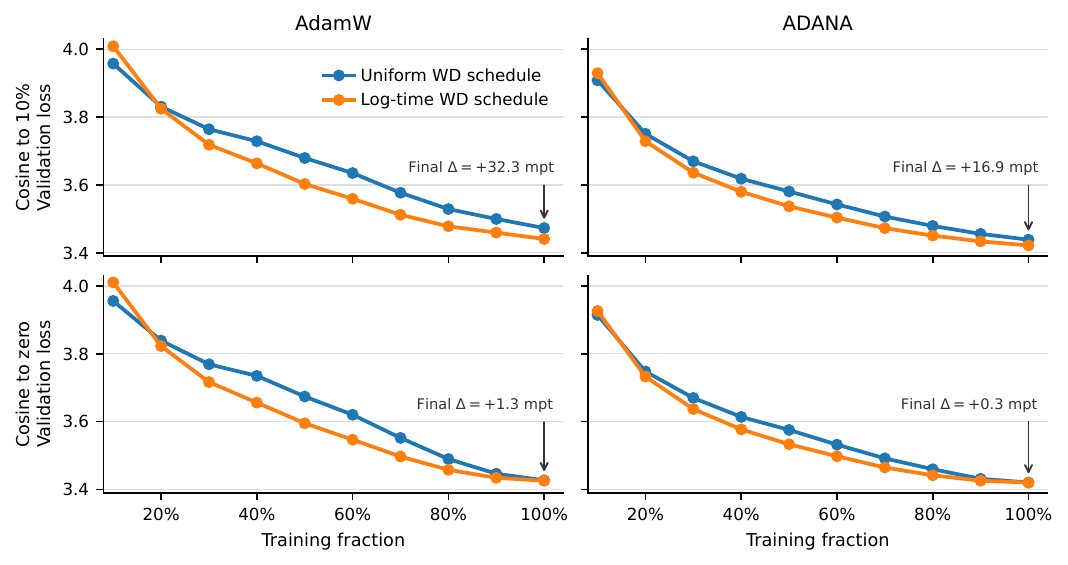}
  \caption{\textbf{The apparent advantage of the log-time WD schedule is
  substantially diminished when the learning rate decays to zero rather than
  to 10\% of its peak.}  We compare the uniform and log-time WD schedules with
  the same base learning rate and matched total parameter shrinkage from
  weight decay alone on the 51M parameter model at $8\times$ OT.  Markers show
  validation evaluations at selected training steps, and lines linearly
  interpolate between them.  With
  a 10\% endpoint, the log-time WD schedule reduces final validation loss by
  32.3 millipoints for AdamW and 16.9 millipoints for ADANA; with a zero
  endpoint, the corresponding differences are 1.3 and 0.3 millipoints.}
  \label{fig:wd-lr-endpoint-interaction-loss-curves}
\end{figure}

Under both learning rate endpoints, log-time WD develops an advantage during
training; with decay to zero, however, most of this advantage disappears
during the final learning rate cooldown.  The results in
\citet{ferbach2026adana} therefore do not by themselves determine which WD
schedule is preferable under our cosine-decay-to-zero protocol.

\subsection{Comparing uniform and log-time WD schedules across OT factors}

\begin{figure}[h]
  \centering
  \includegraphics[width=\textwidth]{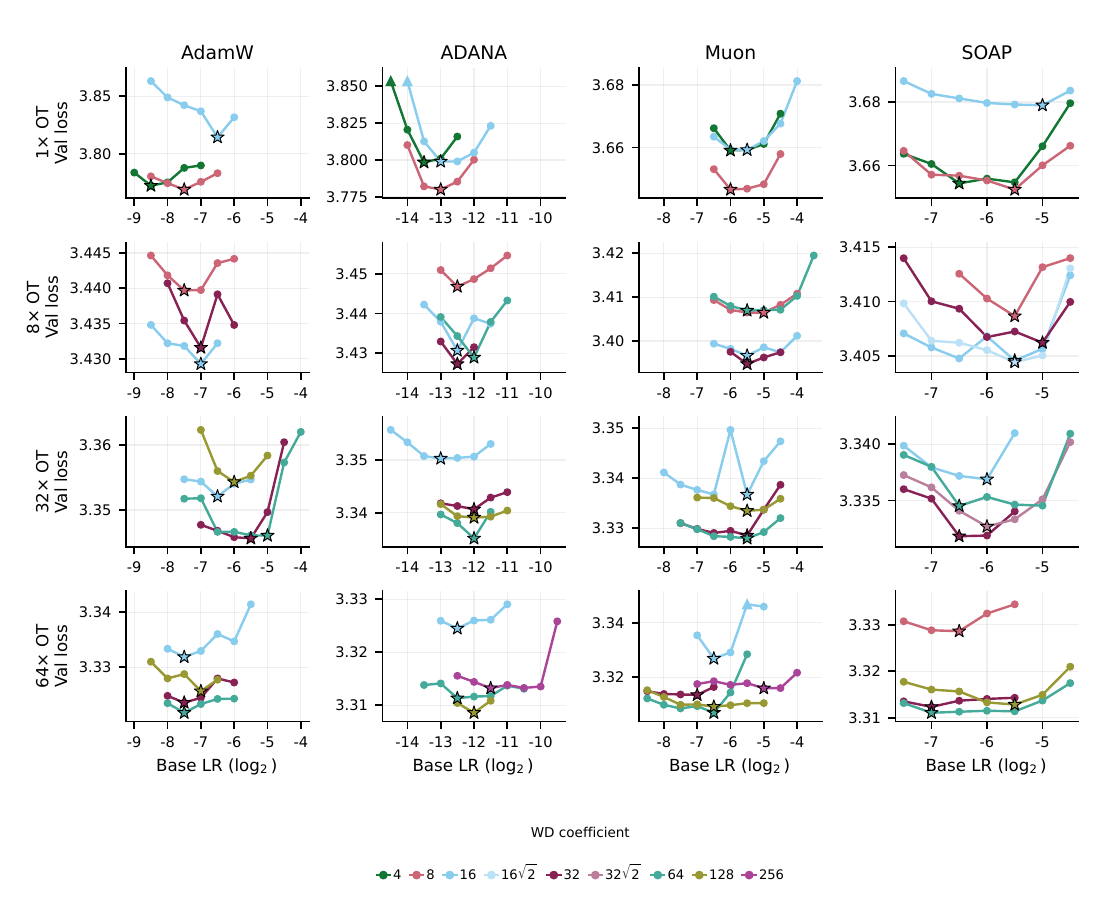}
  \caption{\textbf{Base learning rate sweeps at each uniform WD coefficient
  on the 51M parameter model with cosine decay to zero.} Columns are
  optimizers and rows are OT factors. Each colored curve holds $c$ fixed;
  circles are measured learning rates and black-edged stars mark the minimum
  within each sweep. All 60 sweeps satisfy the interiority criterion in
  Section~\ref{app:sweep-interiority}. Facets use independent vertical ranges
  to expose the local learning rate basin; upward triangles retain observed
  losses above a facet's displayed range.}
  \label{fig:wd-uniform-clock-learning-rate-sweeps}
\end{figure}

Because our primary learning rate schedule decays to zero, we next compare
uniform and log-time WD under this endpoint.  For each optimizer, WD schedule,
and OT factor, we evaluate a two-dimensional grid over base learning rate and
WD coefficient.  At each WD coefficient, we sweep the base learning rate
independently and then compare the resulting minimum losses across
coefficients, following the sweep-interiority criteria in
Section~\ref{app:sweep-interiority}.  The summary figures report quantities
minimized over base learning rate, while Section~\ref{app:complete-sweeps}
reports the coefficient grids in Table~\ref{tab:wd-coefficient-grids} and
shows every evaluated point in
Figures~\ref{fig:wd-uniform-clock-learning-rate-sweeps}
and~\ref{fig:wd-log-clock-learning-rate-sweeps}.  This comparison therefore
makes no assumption about how the WD coefficient should scale across training
horizons.

\begin{figure}[tp]
  \centering
  \includegraphics[width=\textwidth]{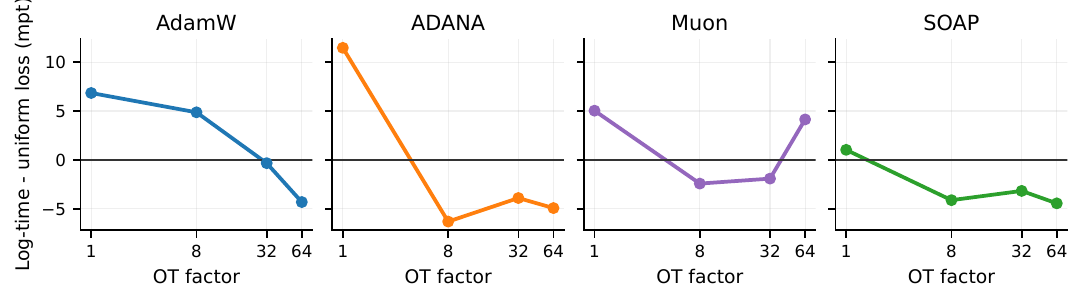}
  \caption{\textbf{The relative performance of the uniform and log-time WD
  schedules depends on both the optimizer and OT factor.}  On the 51M
  parameter model
  with cosine decay to zero and fixed optimizer-specific momentum, each point
  shows $10^3(L_{\mathrm{log}}-L_{\mathrm{uniform}})$ after selecting the best
  $(\text{base learning rate},\text{WD coefficient})$ pair for each WD
  schedule.  Positive values favor the uniform WD schedule, while negative
  values favor the log-time WD schedule.}
  \label{fig:wd-clock-comparison-across-ot}
\end{figure}

Uniform WD performs better at $1\times$ OT for AdamW, ADANA, and Muon.  At
higher OT factors, log-time WD becomes competitive or better for some
optimizers, including ADANA and SOAP at the longest measured horizons, but it
is not uniformly superior across optimizers or horizons.

This comparison allows the WD coefficient to vary independently at every OT
factor.  It therefore compares the two WD schedules under per-horizon tuning
rather than providing a coefficient scaling rule that transfers across
training horizons.

\subsection{WD coefficient scaling rules across OT factors}
\label{app:wd-coefficient-scaling}

When computationally feasible, the best way to set weight decay is to jointly
tune the base learning rate and WD coefficient at every training horizon.
However, this joint sweep is often computationally impractical.  We therefore
investigate how the WD coefficient should scale with training horizon to
determine a scaling rule that can be used in place of per-horizon WD tuning.

An equivalent view of AdamW treats the parameter iterate itself as an
exponential moving average of preconditioned updates: in conventional notation
with constant learning rate $\eta$ and weight decay $\lambda$, the averaging
coefficient is $\eta\lambda$ and the resulting memory timescale is
$1/(\eta\lambda)$ updates \citep{wang2025adamw}. This view has recently been
used to explain other horizon-dependent phenomena. \citet{bergsma2025collapse}
show that the normalized AdamW timescale controls training-loss-curve shape and
must be matched, together with tokens per parameter, for curves to collapse
across model scales. \citet{bergsma2025trec} predict when training data are
retained from AdamW's implicit EMA coefficients and use the resulting
re-evaluation curves to design data curricula. We therefore interpret WD
coefficient scaling as controlling a normalized optimizer memory, rather than
only the strength of regularization.

Prior work in \citet{bergsma2025powerlines} asks the same question for AdamW
under a uniform WD schedule: how should the WD coefficient scale with model
size, training horizon, and batch size?  They define the normalized AdamW
timescale
\begin{equation}
  \tau
  =
  \frac{B}{\eta\lambda D}
  =
  \frac{1}{\eta\lambda S},
\end{equation}
and find that the optimal timescale scales approximately as
\begin{equation}
  \tau_{\mathrm{opt}}
  \propto
  \left(\frac{D}{N}\right)^{-0.527}.
\end{equation}
They also find that, at fixed model size and training horizon, the optimal raw
AdamW coefficient $\lambda$ scales approximately linearly with batch size
below the critical batch size.  In our uniform WD parameterization, $c/S$
corresponds to the peak effective decay coefficient $\eta\lambda$, and hence
$c=1/\tau$.  Thus, at fixed token horizon, their batch-scaling prescription
corresponds to holding $c$ fixed as batch size varies: the factor of $B$ is
already absorbed through $S=D/B$.  Their horizon-scaling result then implies
\begin{equation}
  \frac{c(f)}{c(1)}
  =
  f^{0.527}
  \approx
  \sqrt{f}.
\end{equation}

Power Lines studies uniform WD with AdamW; we extend this analysis to log-time
WD and the broader set of optimizers considered here.

\begin{figure}[tp]
  \centering
  \includegraphics[width=\textwidth]{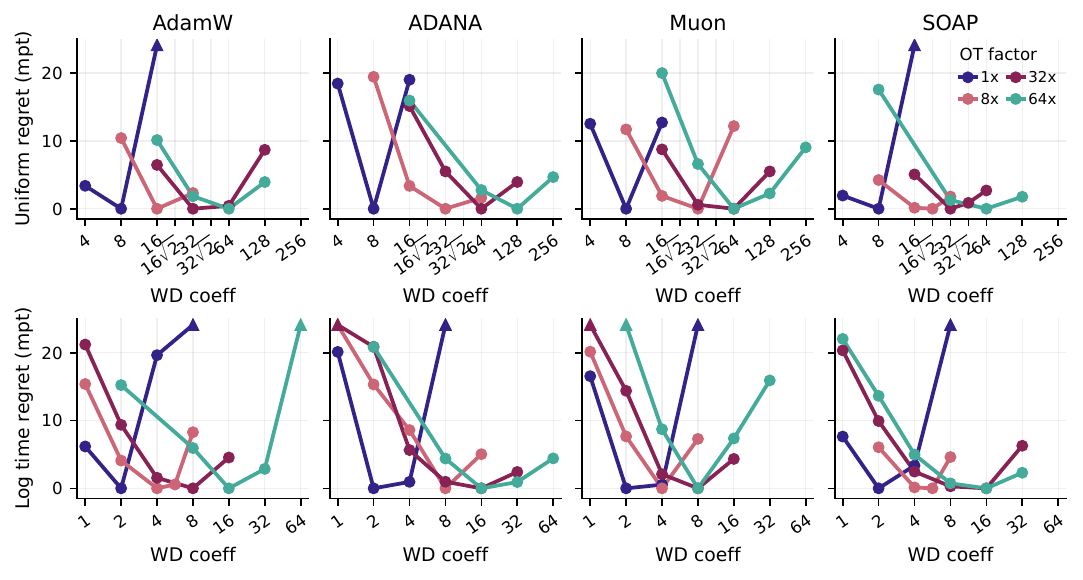}
  \caption{\textbf{The best WD coefficient changes with OT factor.}  Each
  curve shows coefficient regret at one OT factor for a
  fixed optimizer and WD schedule on the 51M parameter model.  The regret of
  coefficient $c$ is
  $10^3\!\left[L(c)-\min_{c'}L(c')\right]$, where $L(c)$ is the best final
  validation loss over the base learning rate sweep at coefficient $c$.  The
  movement of the minima across curves shows that a coefficient selected at
  one training horizon need not transfer to another.  The learning rate
  schedule uses cosine decay to zero and optimizer-specific momentum is held
  fixed.  Upward triangles at the upper boundary represent regrets greater
  than $25$ millipoints.}
  \label{fig:wd-coefficient-regret-across-ot}
\end{figure}

Using the per-horizon joint sweeps described above,
Figure~\ref{fig:wd-coefficient-regret-across-ot} shows the resulting regret for
each WD coefficient after minimizing over the base learning rate,
\begin{equation}
  \mathcal{R}(c;f)
  =
  10^3\left[
    L_f(c)-\min_{c'}L_f(c')
  \right],
  \label{eq:wd-coefficient-regret}
\end{equation}
where $L_f(c)$ is the best final validation loss over the learning rate sweep
at coefficient $c$.

The best WD coefficient generally increases with OT factor for both WD
schedules, although its precise value and the width of the low-regret region
differ across optimizers.  Consequently, fixing the coefficient selected at
$1\times$ OT can incur increasing regret at longer training horizons.

To quantify this trend, let $c^*_{o,s}(f)$ denote the best coefficient for
optimizer $o$, WD schedule $s$, and OT factor $f$ after minimizing over the
base learning rate.  We anchor the uniform and log-time WD schedules at their
$1\times$ OT coefficients, $c_{\mathrm{uniform}}(1)=8$ and $c_{\log}(1)=2$,
and fit a separate power-law exponent for each WD schedule across optimizers
by least squares in log space:
\begin{equation}
  \log_2 c^*_{o,s}(f)
  =
  \log_2 c_s(1)
  +
  \alpha_s\log_2 f
  +
  \epsilon_{o,s,f}.
  \label{eq:wd-coefficient-power-law-fit}
\end{equation}
The independently fitted exponents are $0.527$ for uniform WD and $0.473$ for
log-time WD.

\begin{figure}[tp]
  \centering
  \includegraphics[width=\textwidth]{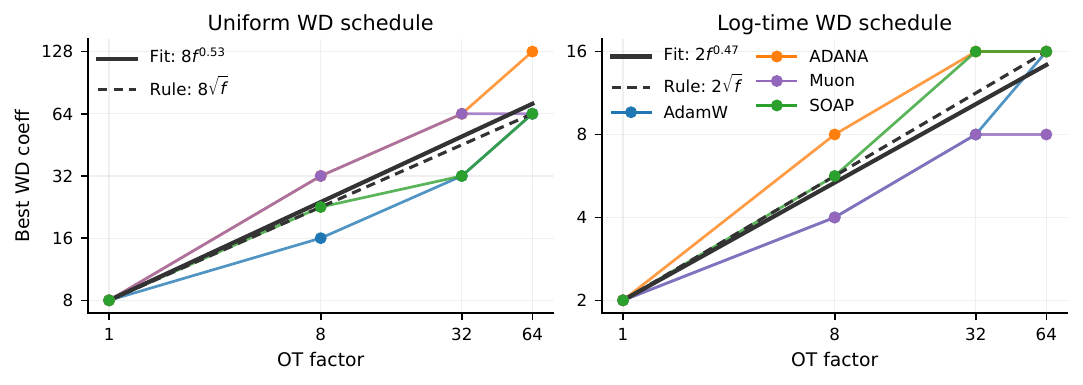}
  \caption{\textbf{The best WD coefficients are consistent with square-root
  scaling across OT factor.}  For each optimizer, WD schedule, and OT factor
  on the 51M parameter model, points show the WD coefficient from the best
  $(\text{base learning rate},\text{WD coefficient})$ pair in the
  two-dimensional grid.  The solid black curves show the separate power-law
  fits in Equation~\ref{eq:wd-coefficient-power-law-fit}, while the dashed
  curves show the adopted square-root rules.  The learning rate schedule uses
  cosine decay to zero and optimizer-specific momentum is held fixed.}
  \label{fig:wd-coefficient-power-law-fit}
\end{figure}

Both fitted exponents are close to $1/2$.  We therefore use the square-root
exponent for both schedules, with schedule-specific $1\times$ OT coefficients
determined by our sweeps:
\begin{equation}
  c_{\mathrm{uniform}}(f)=8\sqrt{f},
  \qquad
  c_{\log}(f)=2\sqrt{f}.
  \label{eq:wd-coefficient-scaling-rules}
\end{equation}
We use these prescriptions when repeating the complete learning rate and WD
coefficient grid is computationally impractical, but do not claim that either
rule is universally optimal.  When direct joint tuning is feasible, we
instead use the best measured $(\text{base learning rate},\text{WD
coefficient})$ pair.

\section{Fixed-Memory Tuning Across Training Horizons}
\label{app:fixed-memory-tuning}

Spectral analyses of deep linear and kernel learning show that higher-eigenvalue
modes are learned earlier, whereas lower-eigenvalue modes require longer
training \citep{saxe2014exact,bordelon2020spectrum}.  This progression suggests
that averaging gradients over longer windows may become increasingly useful at
extended training horizons: longer windows can accumulate weak but persistent
directional signal needed to resolve harder modes.  This motivation is
consistent with the long-memory benefits observed
for AdEMAMix, the growing-memory construction of DANA, and the use of AdEMAMix
in the long-duration Apertus pretraining runs
\citep{pagliardini2024ademamix,ferbach2025dana,apertus2025}.  However, longer
memory also responds more slowly to changes in the gradient distribution and
can retain stale or unusually large gradients for longer, creating a competing
stability consideration.  For fixed-memory optimizers---which use scalar
coefficients such as $\beta$ to maintain a constant effective memory throughout
a run---we therefore ask how the memory coefficient should vary across training
horizons.

\subsection{Per-horizon memory tuning}

We conduct this analysis on the 51M-parameter model, where it is
computationally feasible to jointly sweep one optimizer-specific memory
coefficient and the base learning rate at every overtraining factor.  For
AdamW, we vary the second-moment coefficient $\beta_2$ and fix
$\beta_1=0.9$.  For Muon, we vary its native momentum coefficient $\beta$ and
fix its non-matrix AdamW path at $(\beta_1,\beta_2)=(0.9,0.95)$.  For SOAP, we
vary the rotated-basis second-moment coefficient $\beta_2$ and fix
$\beta_1=\beta_{\mathrm{Sh}}=0.95$ with preconditioner refresh interval
$K=10$.  Our choice to tune AdamW's $\beta_2$ is supported by
\citet{marek2025smallbatch}, who find that preserving the second-moment
half-life transfers better across batch sizes while $\beta_1=0.9$ remains
effective.

We report each momentum coefficient using the effective memory window $M$
defined in Section~\ref{sec:optimizer-memory-view}. The $M=50$ setting used in
our uniform-weight-decay comparison corresponds to $\beta=0.98$ and a
half-life of approximately $34$ optimizer updates. At fixed batch size,
holding $\beta$ constant as the overtraining factor increases makes this
window a progressively smaller fraction of training.

For every optimizer, overtraining factor, and value of $M$, we independently
sweep the base learning rate.  For all experiments, we sweep over the memory
grid $M\in\{20,40,80,160,320,640,1280\}$.  We additionally test $M=10$ at six
short-horizon settings whose minima initially reached the lower boundary, and
extend AdamW and SOAP to $M=2560$ at $128\times$ OT.  We evaluate these sweeps
using the interiority criteria in Appendix~\ref{app:sweep-interiority}.  Within
each $(\text{optimizer},M,\text{OT factor})$ setting, we first require the
optimal learning rate to be interior in the learning rate sweep.  After
minimizing over learning rate, we require the optimal memory to be interior in
the memory sweep at every $(\text{optimizer},\text{OT factor})$ setting.  At
each horizon, the per-horizon treatment selects the
$(M,\text{base learning rate})$ pair with the lowest final validation loss.

\subsection{Horizon-Tuned Memory Results}

\paragraph{Longer horizons favor longer memory.}
Figure~\ref{fig:main-memory-frontier} shows that longer training horizons
generally favor longer fixed memory.  From $1\times$ to $128\times$ OT, the
optimal memory increases from $M=20$ to $M=640$ for Muon, from $M=20$ to
$M=1280$ for AdamW, and from $M=40$ to $M=1280$ for SOAP, although the
trajectories are not always monotonic.

\begin{figure}[tp]
  \centering
  \includegraphics[width=\textwidth]{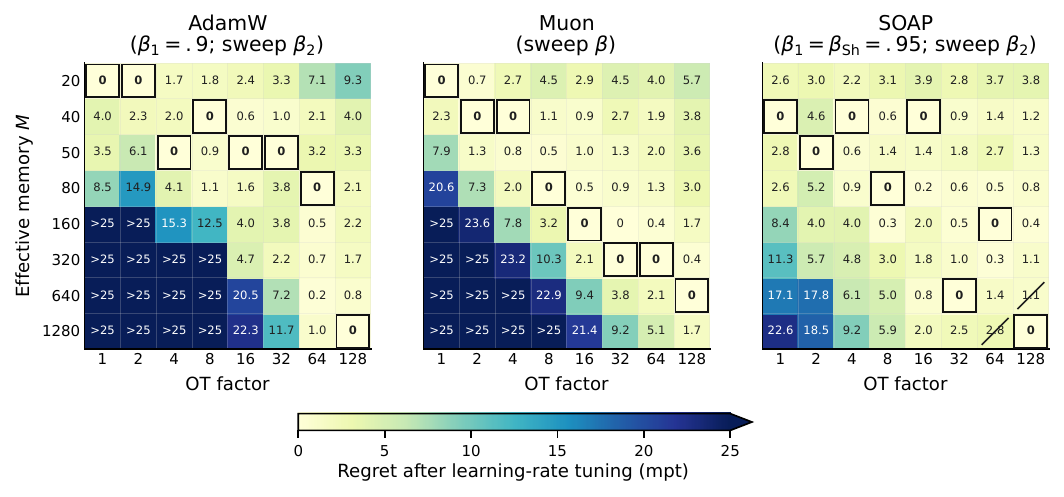}
  \caption{\textbf{The low-regret region shifts toward longer fixed memory as
  the training horizon increases.} For each optimizer, effective memory, and
  OT factor, we independently sweep the base learning rate and report regret
  relative to the best memory for that optimizer and horizon.  Black outlines
  mark the minimizing memory. Diagonal slashes indicate points whose learning
  rate sweeps were incomplete.}
  \label{fig:fixed-memory-regret-heatmap}
\end{figure}

We do not infer a memory scaling rule from these optima.
Figure~\ref{fig:fixed-memory-regret-heatmap} shows that many horizons have broad
low-regret regions, so small loss differences can substantially change the
minimizing $M$.  We instead use the optimal memory at each horizon as a
per-horizon tuned comparison and leave a transferable memory prescription to
future work.

\paragraph{Memory and learning rate must be tuned jointly.}

\begin{figure}[tp]
  \centering
  \includegraphics[width=\textwidth]{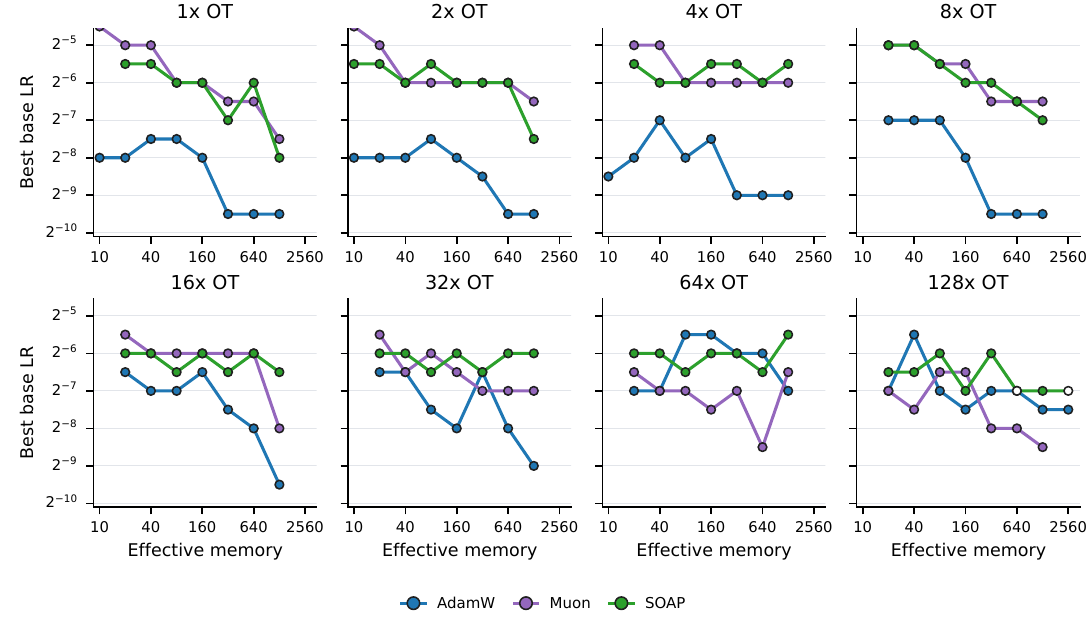}
  \caption{\textbf{The best base learning rate generally decreases as fixed
  memory grows, with optimizer- and horizon-dependent deviations.} Each panel
  fixes the OT factor and compares AdamW, Muon, and SOAP across effective
  memory.  At each memory, the plotted value is the base learning rate
  attaining the lowest final validation loss in an independent learning rate
  sweep. Open circles indicate points whose learning rate sweeps were
  incomplete.}
  \label{fig:fixed-momentum-learning-rate-scaling}
\end{figure}

Figure~\ref{fig:fixed-momentum-learning-rate-scaling} shows that the
loss-minimizing base learning rate generally decreases as effective memory
grows, although the relationship depends on optimizer and horizon.  This
coupling creates a tradeoff: longer memory becomes useful at long horizons but
is typically paired with a smaller learning rate, which can slow learning
throughout the run.  A fixed-memory optimizer must accept one compromise
between memory and learning rate throughout training, whereas scheduled memory
can use shorter memory early and lengthen it as training proceeds.  This
provides one plausible explanation for why per-horizon tuned constant memory
does not fully match scheduled memory.

\paragraph{Per-horizon tuning improves fixed-memory optimizers.}

\begin{figure}[tp]
  \centering
  \includegraphics[width=\textwidth]{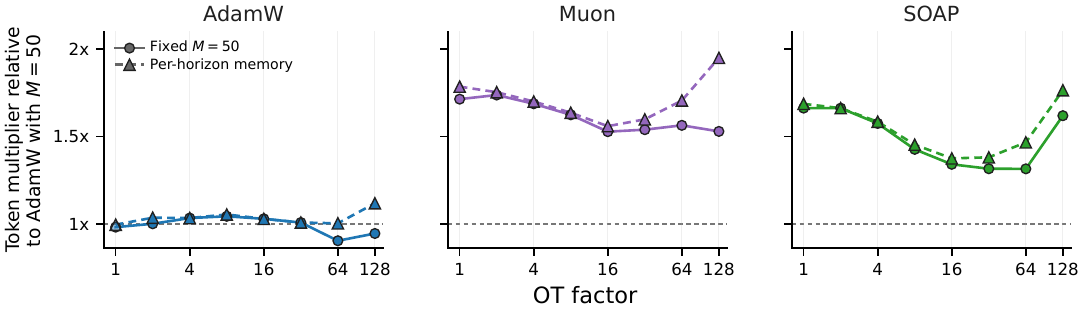}
  \caption{\textbf{Per-horizon memory tuning improves fixed-memory
  optimizers.} For the 51M model, circles show each optimizer with fixed
  effective memory $M=50$, after independently sweeping the base learning
  rate at every OT factor.  Triangles show the lowest measured loss after
  jointly sweeping fixed memory and base learning rate at each horizon.  We
  convert these loss values to token multipliers relative to AdamW with
  $M=50$, corresponding to $\beta_2=0.98$, by inverting the fixed-$M=50$ AdamW
  curve from the shared-$m$ Skaling fit.  Because scaling fits are not well
  identified from this small set of horizon-tuned points, we connect adjacent
  points with piecewise-linear segments rather than fitted curves.  Per-horizon
  memory tuning provides
  modest gains for AdamW but substantially strengthens Muon and SOAP at the
  longest measured horizons.}
  \label{fig:fixed-memory-adamw-relative-per-horizon}
\end{figure}

The fixed-memory comparison uses $\beta_2=0.98$ for AdamW and SOAP and
$\beta=0.98$ for Muon, corresponding to $M=50$.  The per-horizon sweeps show
that $M=50$ is near-optimal at moderate horizons, but tends to provide too much
memory at the shortest horizons and too little at the longest.  The gains
therefore appear at both ends of our overtraining range and are largest at
$128\times$ OT.  At $128\times$ OT, per-horizon tuning improves validation loss
by $3.3$ mpt for AdamW, $3.6$ mpt for Muon, and $1.3$ mpt for SOAP.
Figure~\ref{fig:fixed-memory-adamw-relative-per-horizon} expresses the
resulting performance as token multipliers relative to fixed-$M=50$ AdamW.

\paragraph{Interpretation and limitations.}

These experiments act as an oracle for constant-memory optimizers: at each
training horizon, they select the best combination of memory and learning rate
under the setting where the memory remains fixed throughout training.  The
results are limited to the 51M model and one memory coefficient per optimizer.

Complete learning rate sweeps for this section are included in
Figures~\ref{fig:fixed-momentum-lr-sweeps-low-ot}
and~\ref{fig:fixed-momentum-lr-sweeps-high-ot} of
Appendix~\ref{app:complete-sweeps}.

\section{Scaling Fits and Token Multipliers}
\label{app:scaling-fits}

\subsection{Primary comparison and measured support}

We fit the 76 uniform weight decay loss values used in
Figures~\ref{fig:main-gamma-one-loss}
and~\ref{fig:main-token-multiplier-collapse}.  These experiments contain nine
OT factors from $1\times$ through $256\times$ for the 51M model, six from
$1\times$ through $32\times$ for the 124M model, and four from $1\times$
through $8\times$ for the 253M model, for each of AdamW, ADANA, Muon, and
SOAP.  At every (optimizer, model size, OT factor) setting, we independently
sweep the base learning rate and use the lowest final validation loss.

\begin{table}[tp]
  \centering
  \small
  \setlength{\tabcolsep}{8pt}
  \renewcommand{\arraystretch}{1.12}
  \begin{tabular}{@{}ll@{}}
    \toprule
    Optimizer & Prescribed optimizer-specific settings \\
    \midrule
    AdamW & $\beta_1=0.9,\ \beta_2=0.98$ \\
    Muon & $\beta=0.98$; Adam route $\beta_1=0.9,\ \beta_2=0.95$ \\
    SOAP & $\beta_1=0.95,\ \beta_2=0.98,\ \beta_{\mathrm{Sh}}=0.95,\ K=10$ \\
    ADANA & $g_2=1,\ g_3=8,\ \kappa=0.85,\ \delta=8$ \\
    \bottomrule
  \end{tabular}
  \caption{Optimizer-specific settings used in the primary comparison. The
  settings are held fixed across model sizes and OT factors; ADANA's memory
  changes within training according to its prescribed log-time rule. The base
  learning rate is swept independently at every coordinate.}
  \label{tab:primary-optimizer-treatments}
\end{table}

In this section, we fit validation loss jointly as a function of
model size and training tokens.  We use these fits to estimate scaling
exponents and high-token loss limits and to define the smooth AdamW reference
curve used to calculate token multipliers.  These quantities therefore depend
on the chosen functional form.

\subsection{Joint model--token functional form}

Let $P$ denote nominal model size and $T$ denote the number of
training tokens, and define the dimensionless coordinates
$p=P/P_0$ and $t=T/T_0$, with $P_0=10^8$ parameters and
$T_0=2\times10^9$ tokens.  Throughout the main text, we model the loss values
with the joint model--token functional form proposed by
\citet{videau2026skaling}, with the residual loss floor $E$ and high-token
model exponent $m$ shared across optimizers:
\[
  \mathcal L_o(P,T)
  = E + \left[
      A_o p^{-m/k_o} + B_o t^{-\beta_o}
    \right]^{k_o},
\]
where $o$ indexes the optimizer.  We refer to this specification as the
shared-$m$ Skaling fit.  The parameters $A_o$, $B_o$, $\beta_o$, and $k_o$ are
optimizer specific, so the four-optimizer fit contains 18 free parameters.

At fixed model size, the high-token limit is
\[
  \mathcal L_{o,\infty}(P)
  = E + A_o^{k_o}p^{-m},
\]
and the excess above this limit decays asymptotically as $T^{-\beta_o}$.
Accordingly, we refer to $\beta_o$ as the fixed-model token-decay exponent
under this functional form.  In the infinite-model limit, the token exponent
is $k_o\beta_o$.  At fixed model size, however, high-token excess loss decays
with exponent $\beta_o$.

Because we have only three model sizes, we cannot reliably estimate a separate
model exponent for each optimizer.  We therefore share $m$ across optimizers
and report optimizer-specific model exponents as a sensitivity analysis; this
modeling choice does not imply that their true exponents are equal.

\subsection{Alternative functional forms and model selection}

Before selecting the shared-$m$ Skaling fit, we considered six alternative
functional forms.

Our first alternative is the Chinchilla functional form, which uses separate
additive terms for model size and training tokens \citep{hoffmann2022empirical},
\[
  \mathcal L_o(P,T)
  = E + A_o p^{-\alpha_o} + B_o t^{-\beta_o}.
\]
We use a shared residual floor $E$ and optimizer-specific values of
$A_o$, $\alpha_o$, $B_o$, and $\beta_o$.

Our second alternative is a parameter--token interaction model motivated by
the loss decompositions on power-law random features models in
\citet{paquette2024phases},
\[
  \mathcal L_o(P,T)
  = E + A_o p^{-\alpha_o} + C_o p^{-\gamma_o}t^{-\delta_o},
\]
where the interaction amplitude $C_o$ and exponents $\gamma_o$ and $\delta_o$
are positive.  This form allows the marginal effect of training tokens to
depend on model size while preserving monotonic improvement in either
coordinate.

We next compare two additive forms using the overtraining factor $f$,
\[
  \mathcal L_o(P,f)
  = E + A_o p^{-\alpha_o} + C_o p^{-u_o}f^{-\delta_o},
\]
with one version sharing $\alpha$ across optimizers and one using
optimizer-specific $\alpha_o$.

Finally, we compare two additional variants within the Skaling family: a
version with optimizer-specific high-token model exponents
$m_o=k_o\alpha_o$, and a version that shares the inner exponent $\alpha$ while
retaining optimizer-specific $k_o$, so that $m_o=k_o\alpha$ remains optimizer
specific.

We compare the models using root mean squared error (RMSE), small-sample
corrected Akaike information criterion (AICc), Bayesian information criterion
(BIC), residual structure, and structured holdout tests.  Lower AICc and BIC
indicate a better tradeoff between fit error and model complexity.

In each structured holdout test, we remove a scientifically meaningful group
of loss values, refit the model, and evaluate its predictions on that omitted
group.  We consider five tests: holding out an entire OT factor; holding out
one complete model size; holding out each model size's longest horizon;
holding out each model size's two longest horizons; and holding out the
longest horizon from all three model sizes simultaneously.  These tests
evaluate prediction across horizon or model-size boundaries.  Ordinary
leave-one-loss-value-out prediction is less informative because neighboring
optimizers and horizons remain in the fit.

\begin{table}[tp]
  \centering
  \small
  \setlength{\tabcolsep}{5pt}
  \begin{tabular}{@{}lrrrr@{}}
    \toprule
    Functional form & Parameters & RMSE (mpt) & AICc & BIC \\
    \midrule
    Chinchilla additive & 17 & 14.349 & -600.54 & -571.47 \\
    Parameter--token interaction & 21 & 14.850 & -580.77 & -548.94 \\
    Additive OT, shared model exponent & 18 & 3.272 & -821.82 & -791.86 \\
    Additive OT, optimizer-specific model exponents & 21 & 3.113 & -818.24 & -786.41 \\
    Skaling, optimizer-specific model exponents & 21 & 3.079 & -819.90 & -788.07 \\
    \textbf{Shared-$m$ Skaling} & 18 & 3.251 & -822.78 & -792.83 \\
    Skaling, shared inner exponent & 18 & 3.290 & -820.96 & -791.01 \\
    \bottomrule
  \end{tabular}
  \caption{In-sample fit and information criteria for the candidate scaling
  forms on the 76 uniform weight decay loss values.}
  \label{tab:scaling-functional-form-selection}
\end{table}

\begin{figure}[tp]
  \centering
  \includegraphics[width=\textwidth]{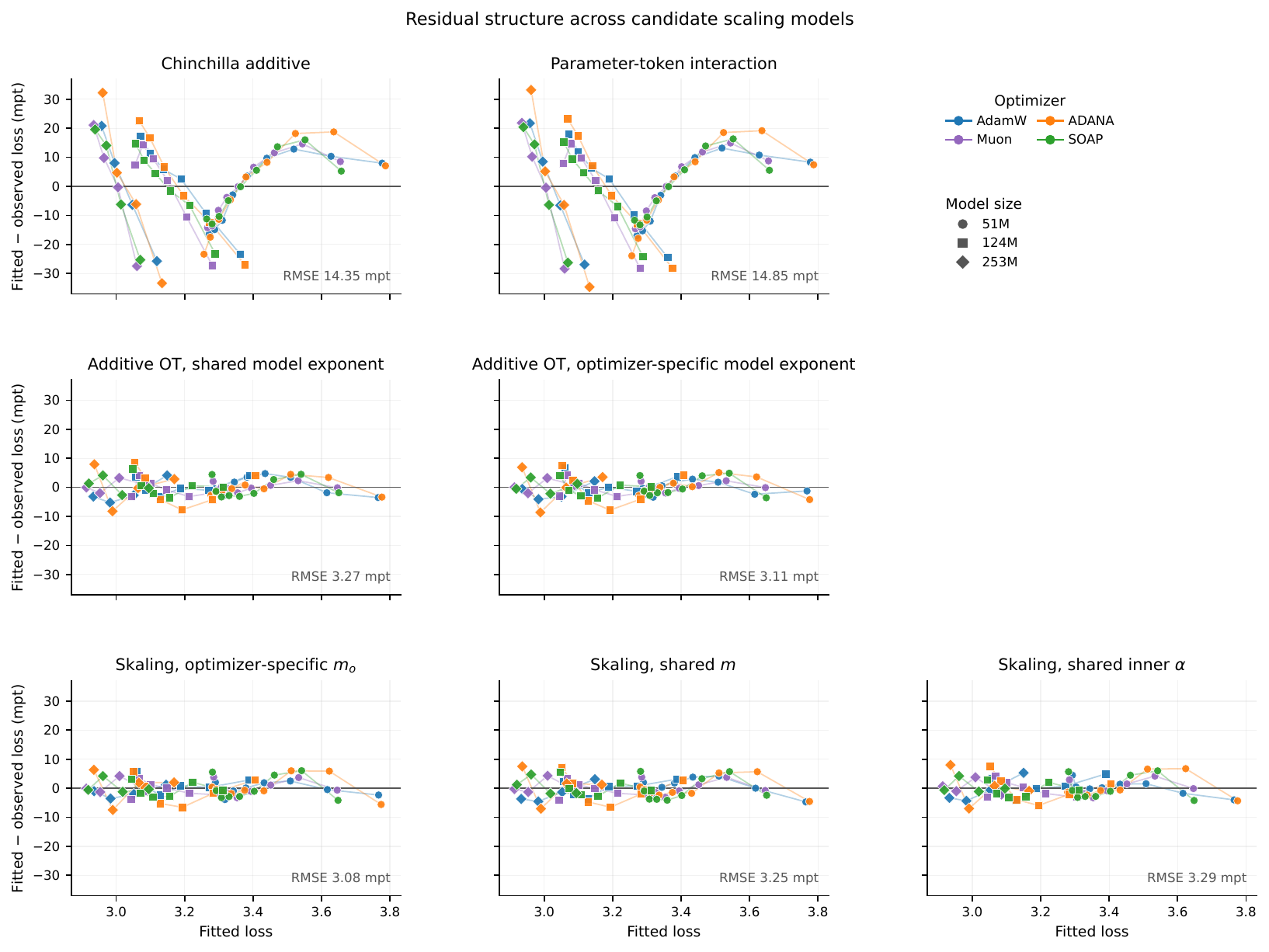}
  \caption{\textbf{The separable Chinchilla model and the parameter--token
  interaction model leave systematic residuals.} Residuals are fitted minus observed validation
  loss in millipoints.  The Chinchilla-additive and parameter--token
  interaction models retain a pronounced U-shaped structure, whereas the five
  additive-OT and Skaling candidates remain substantially closer to zero
  across the observed loss range.}
  \label{fig:scaling-candidate-residuals}
\end{figure}

The Chinchilla and parameter--token interaction forms are not competitive on
these experiments.  Their RMSEs exceed $14$ millipoints, compared with
$3.08$--$3.29$ millipoints for the remaining five candidates.  As
Figure~\ref{fig:scaling-candidate-residuals} shows, both also leave a
pronounced U-shaped residual pattern: they overpredict the lowest and highest
losses while underpredicting intermediate losses.  We therefore restrict the
structured-holdout comparison below to the remaining five candidates, which
fit the observed loss values substantially better.

For the five better-fitting models,
Table~\ref{tab:scaling-structured-holdout-summary} reports pooled RMSE in
millipoints for the structured holdout tests.  We additionally report every
fold's training count, held-out count, and RMSE in
Table~\ref{tab:scaling-structured-holdout-folds}.

\begin{table}[tp]
  \centering
  \scriptsize
  \setlength{\tabcolsep}{3pt}
  \resizebox{\textwidth}{!}{%
  \begin{tabular}{@{}lrrrrr@{}}
    \toprule
    Functional form & Hold out OT & Hold out longest & Hold out two longest & Joint frontier & Hold out model \\
    \midrule
    Additive OT, shared model exponent & 5.36 & 5.89 & 6.00 & 6.92 & 36.30 \\
    Additive OT, optimizer-specific model exponents & 5.64 & 6.06 & 5.99 & 7.65 & 13.98 \\
    Skaling, optimizer-specific model exponents & 5.94 & 5.99 & 5.49 & 7.70 & 184.49 \\
    \textbf{Shared-$m$ Skaling} & 5.73 & 5.86 & 5.53 & 6.95 & 59.45 \\
    Skaling, shared inner exponent & 5.97 & 6.11 & 5.61 & 7.32 & 59.68 \\
    \bottomrule
  \end{tabular}}
  \caption{Pooled RMSE in millipoints for the structured holdout tests for the
  five better-fitting models.}
  \label{tab:scaling-structured-holdout-summary}
\end{table}

\begin{table}[tp]
  \centering
  \scriptsize
  \setlength{\tabcolsep}{2.5pt}
  \resizebox{\textwidth}{!}{%
  \begin{tabular}{@{}llrrrrrrr@{}}
    \toprule
    & & & & \multicolumn{2}{c}{Additive OT} & \multicolumn{3}{c}{Skaling} \\
    \cmidrule(lr){5-6}\cmidrule(lr){7-9}
    Test & Held-out group & Train $n$ & Held-out $n$ & Shared $\alpha$ & Optimizer-specific $\alpha_o$ & Optimizer-specific $m_o$ & \textbf{Shared $m$} & Shared $\alpha$ \\
    \midrule
    Hold out OT & $1\times$ & 64 & 12 & 9.49 & 10.35 & 11.37 & 10.75 & 11.32 \\
    & $2\times$ & 64 & 12 & 3.51 & 3.69 & 4.12 & 3.97 & 4.21 \\
    & $4\times$ & 64 & 12 & 5.37 & 5.46 & 5.09 & 5.02 & 4.99 \\
    & $8\times$ & 64 & 12 & 3.80 & 3.51 & 3.45 & 3.82 & 4.03 \\
    & $16\times$ & 68 & 8 & 2.99 & 2.76 & 2.60 & 2.80 & 2.96 \\
    & $32\times$ & 68 & 8 & 5.42 & 5.80 & 5.34 & 4.86 & 4.93 \\
    & $64\times$ & 72 & 4 & 2.68 & 2.76 & 3.23 & 3.17 & 2.91 \\
    & $128\times$ & 72 & 4 & 2.18 & 2.26 & 1.24 & 1.13 & 1.48 \\
    & $256\times$ & 72 & 4 & 4.58 & 4.43 & 5.98 & 6.10 & 6.26 \\
    \midrule
    Hold out model & 51M & 40 & 36 & 46.19 & 19.74 & 256.43 & 35.72 & 15.75 \\
    & 124M & 52 & 24 & 18.73 & 4.58 & 64.19 & 61.61 & 76.52 \\
    & 253M & 60 & 16 & 30.58 & 4.50 & 86.77 & 90.69 & 87.04 \\
    \midrule
    Hold out longest horizon & 51M, $256\times$ & 72 & 4 & 4.58 & 4.43 & 5.98 & 6.10 & 6.26 \\
    & 124M, $32\times$ & 72 & 4 & 7.42 & 8.11 & 7.15 & 6.31 & 6.50 \\
    & 253M, $8\times$ & 72 & 4 & 5.29 & 4.98 & 4.55 & 5.08 & 5.53 \\
    \midrule
    Hold out two longest & 51M, $128$--$256\times$ & 68 & 8 & 4.81 & 4.76 & 5.53 & 5.63 & 5.60 \\
    & 124M, $16$--$32\times$ & 68 & 8 & 7.31 & 7.73 & 6.23 & 5.73 & 6.04 \\
    & 253M, $4$--$8\times$ & 68 & 8 & 5.60 & 5.04 & 4.58 & 5.22 & 5.15 \\
    \midrule
    Joint frontier & All longest horizons & 64 & 12 & 6.92 & 7.65 & 7.70 & 6.95 & 7.32 \\
    \bottomrule
  \end{tabular}}
  \caption{Per-fold support and RMSE in millipoints for each structured
  holdout test for the five better-fitting models.}
  \label{tab:scaling-structured-holdout-folds}
\end{table}

We select the shared-$m$ Skaling fit as our primary scaling fit and use it
throughout the main text.  Among the candidate functional forms that also have 18
parameters, it has the lowest observed AICc and BIC, and its errors when
predicting held-out long horizons are comparable to those of the
best-performing alternative.  Allowing optimizer-specific model exponents
reduces the in-sample RMSE by $0.171$ mpt, but adds three parameters.  The
paired-coordinate bootstrap does not consistently rank the candidate models,
so this selection is not evidence that the shared-$m$ form is uniquely
correct.

Predictions for a completely held-out model size are substantially less
accurate than predictions across the OT axis for every candidate.  Because
only three model sizes are available, we share $m$ across optimizers to reduce
overfitting along the model-size axis; we do not interpret the current
experiments as identifying optimizer-specific model-scaling laws.  In
Appendix~\ref{app:floor-decay-tradeoff}, we check whether the qualitative
conclusions persist across both the Skaling and additive OT families rather
than relying on a single parameterization.

\begin{figure}[tp]
  \centering
  \includegraphics[width=\textwidth]{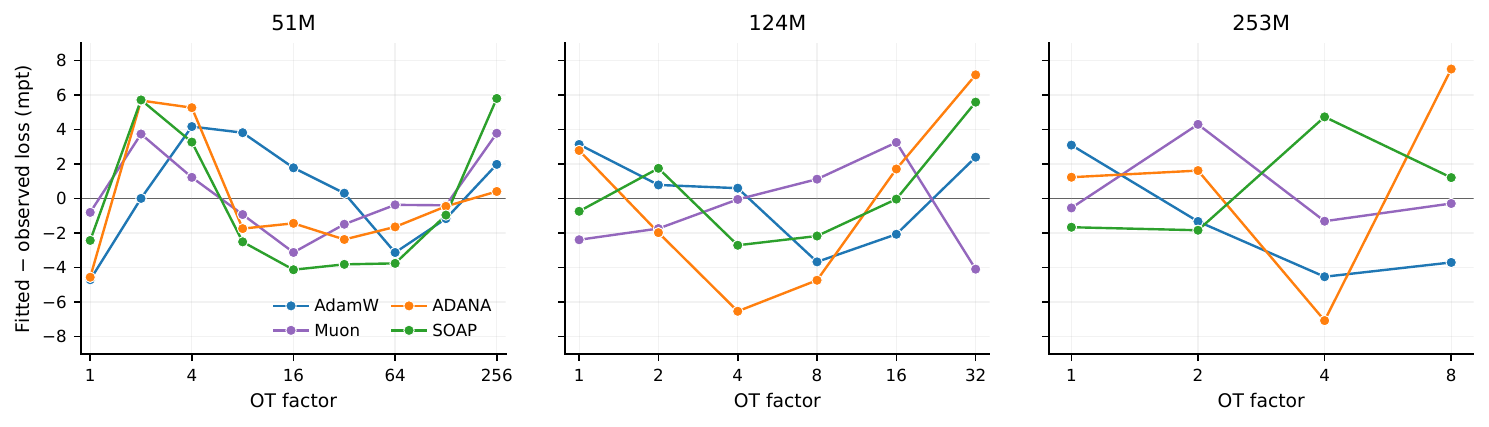}
  \caption{\textbf{Residuals of the selected shared-$m$ Skaling fit remain
  small across model sizes and OT values.} Residuals are fitted minus observed
  validation loss in millipoints, shown separately for the 51M, 124M, and
  253M models.  Lines connect adjacent OT factors for the same optimizer.  The
  overall RMSE is $3.251$ millipoints.}
  \label{fig:shared-m-skaling-residuals-by-ot}
\end{figure}

\subsection{Token multipliers}

Following prior work, we compare optimizers using the ratio of the training
tokens required by a baseline optimizer to those used by an indicated
optimizer at the same loss.  At fixed model size, this is the same quantity
called the \emph{compute multiplier} by
\citet{qiu2025hyperparameter,ferbach2026adana} and \emph{token-equivalent
speedup} by \citet{wen2026hyperball}.  We refer to it as the \emph{token
multiplier}.  A value above one means that the baseline requires more tokens
to attain the same loss and therefore favors the indicated optimizer.

At fixed model size $p$, let $t_b(\ell;p)$ denote the normalized number of
training tokens required by a baseline optimizer $b$ to reach loss $\ell$.  If
optimizer $o$ achieves loss $\ell_o$ after normalized token count $t$, its
token multiplier relative to $b$ is
\[
  M_{T,o\mid b}(p,t)
  = \frac{t_b(\ell_o;p)}{t}.
\]
A value above one means that the baseline optimizer requires more training
tokens to reach optimizer $o$'s loss, while a value below one favors the
baseline.  For most optimizer comparisons in this paper, we use AdamW as the
baseline.

Under the shared-$m$ Skaling fit, the fitted token count required by a baseline
optimizer $b$ to reach loss $\ell$ is
\[
  \widehat t_b(\ell;p)
  = \left[
      \frac{B_b}{
        (\ell-E)^{1/k_b} - A_b p^{-m/k_b}
      }
    \right]^{1/\beta_b}.
\]
This inverse is finite when $\ell$ lies above the baseline's fitted
finite-model high-token limit,
\[
  \mathcal L_{b,\infty}(p)
  = E + A_b^{k_b}p^{-m}.
\]
As $\ell$ approaches this limit from above, the required token count diverges.
If $\ell$ falls below the fitted limit, the baseline cannot match it with any
finite token count under the fitted model.

For the smooth curves in the main text, we evaluate optimizer $o$'s fitted
loss and invert the fitted baseline curve; we refer to these as
\emph{fit-to-fit token multipliers}.  At each experimental
$(\text{optimizer},\text{model size},\text{OT factor})$ setting, we instead
compute a \emph{measured-to-fit token multiplier} by inverting the observed
validation loss through the fitted baseline curve.  A measured-to-fit
multiplier therefore depends on the baseline fit but not on a fitted curve for
optimizer $o$.

When the inferred baseline token count exceeds the largest token budget
measured for that baseline at the same model size, the multiplier depends on
extrapolating rather than interpolating the baseline fit.  We denote
extrapolated token multiplier points using translucent marker fills with
opaque outlines.  For fit-to-fit curves, the line becomes dotted starting at
the OT factor where the equivalent baseline token count exceeds the largest
token budget measured for that baseline at the same model size.

\section{Outscaling Across Training Horizons}
\label{app:outscaling-analysis}

As in the main text, we use outscaling to describe an improvement in an
optimizer's relative efficiency as OT increases.  Here we use the scaling fits
from Appendix~\ref{app:scaling-fits} to examine whether that improvement is
better described by an improved token-decay exponent, a lower finite-model
high-token loss limit, or both.

\subsection{Definition and floor--decay tradeoff}
\label{app:floor-decay-tradeoff}

Outscaling can arise through an improved token-decay exponent, a lower
high-token validation-loss limit, or both. Optimizer treatments may approach
different finite-model limits because they can select different solutions or
retain different residual optimization and noise floors. Under our
experiments, we typically cannot identify these two effects separately, so we
propose a profile-fit analysis that traces their tradeoff and determines
whether the observed outscaling requires an advantage in at least one.

At fixed model size $P$, let
$\mathcal L_{o,\infty}(P)$ denote optimizer $o$'s finite-model high-token
loss limit, and let $\beta_o$ denote its token-decay exponent under the
shared-$m$ Skaling fit. For optimizer $o$ and baseline $b$, define
\[
  \Delta_\infty(P)
  = \mathcal L_{b,\infty}(P)-\mathcal L_{o,\infty}(P),
  \qquad
  \Delta_\beta = \beta_o-\beta_b.
\]
Positive $\Delta_\infty$ means that optimizer $o$ approaches a lower
finite-model loss limit, while positive $\Delta_\beta$ means that its excess
loss decays faster with additional training tokens. Outscaling across the
overtraining axis may therefore arise from either advantage or from both
simultaneously.

To expose this tradeoff, we fix a candidate value of $\Delta_\infty$, refit
the remaining scaling parameters, and record the corresponding best-fitting
value of $\Delta_\beta$. Each profile therefore shows the token-decay exponent
difference implied by a particular assumption about the high-token-limit
difference.

We plot the portion of each profile curve that contains the unconstrained
optimum and satisfies a 95\% profile-likelihood threshold. Let $R(d)$ be the
minimum residual sum of squares after fixing
$\Delta_\infty=d$, and let $R_{\min}$ be the unconstrained minimum. We
estimate a common residual variance as
$\widehat\sigma^2=R_{\min}/(n-p)$, where $n$ is the number of fitted loss
values and $p$ is the number of free parameters in the unconstrained fit, and
display the contiguous segment containing the optimum for which
\[
  \frac{R(d)-R_{\min}}{\widehat\sigma^2}
  \leq \chi^2_{1,0.95} \simeq 3.84.
\]
The value $3.84$ is the $95$th percentile of a chi-squared distribution with
one degree of freedom, corresponding to the 95\% one-parameter
profile-likelihood threshold under independent Gaussian residuals with common
variance. Our loss values are selected experimental measurements rather than
independent noisy replicates, so we use this threshold only as a common display
criterion across functional forms, not as a calibrated confidence interval.

We repeat this procedure for four of the alternative functional forms: the
additive-OT and Skaling families each with either a model exponent shared
across optimizers or optimizer-specific model exponents. This separates
sensitivity to the model--token coupling from sensitivity to the
model-exponent constraint.

We interpret the profiles as sets of compatible floor--decay decompositions
rather than precise independent estimates of either quantity. We use them to
determine whether the observed outscaling requires the compared optimizer to
have an advantage in its token-decay exponent, its high-token loss limit, or
both.

\subsection{Pairwise comparisons under uniform weight decay}

We first compare all optimizers under uniform weight decay, since this
provides a fair comparison under the most standard weight decay setting.

\begin{figure}[tp]
  \centering
  \includegraphics[width=\textwidth]{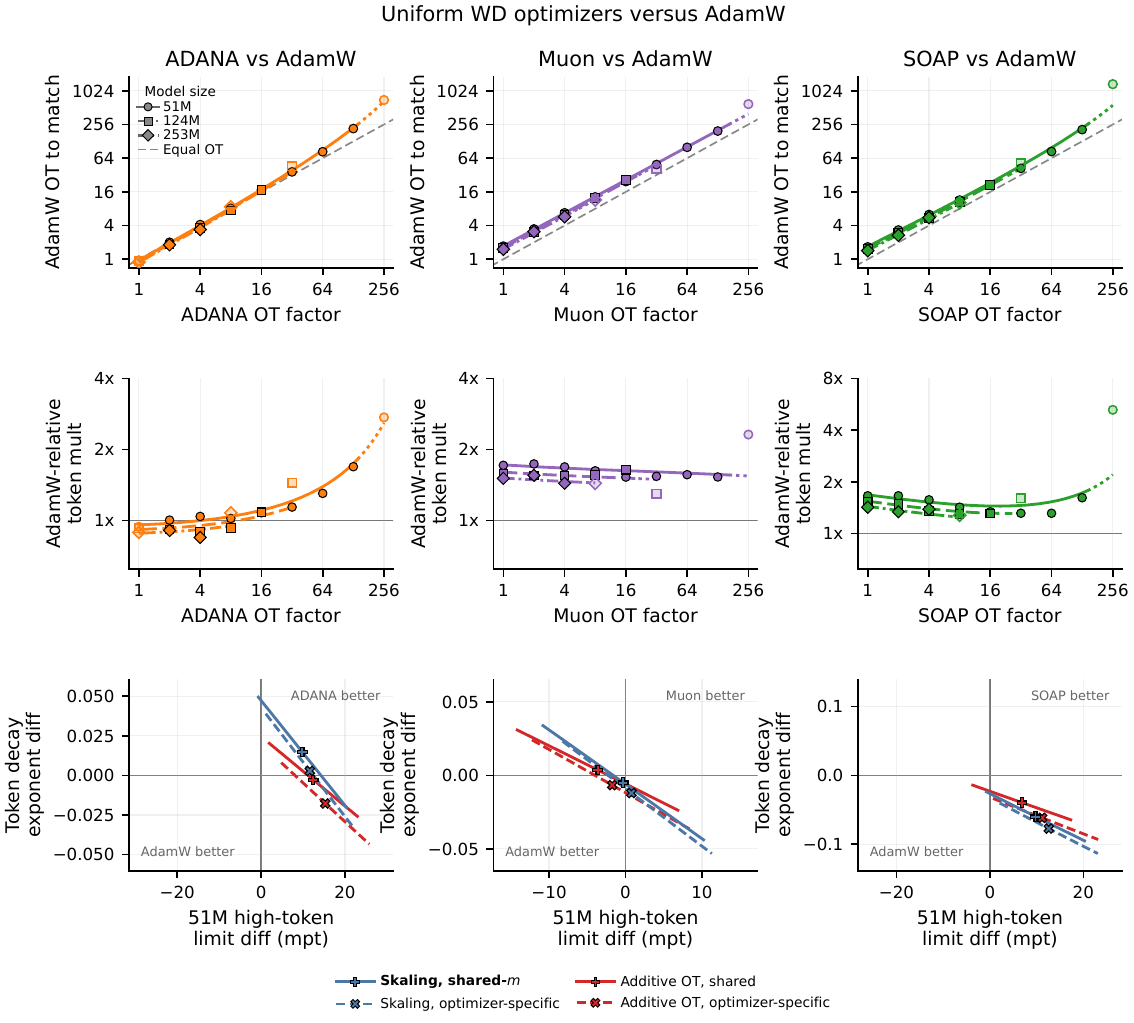}
  \caption{\textbf{Uniform-WD optimizers exhibit distinct outscaling behavior
  relative to AdamW.} Columns compare ADANA, Muon, and SOAP with AdamW. The
  top row plots the AdamW OT factor required to match each optimizer at the OT
  factor on the horizontal axis; values above the equal-OT diagonal favor the
  optimizer named first. The middle row reports the corresponding
  AdamW-relative token multiplier. Curves are fit-to-fit comparisons under the
  shared-$m$ Skaling fit, while markers invert measured loss values through the
  fitted AdamW curve. Dotted curve segments and translucent marker fills
  require extrapolating AdamW beyond its largest measured token budget at that
  model size. The bottom row profiles the tradeoff between the 51M high-token
  limit difference and token-decay exponent difference under four functional
  forms; symbols mark the unconstrained optima and the displayed curves
  satisfy the 95\% profile threshold. Positive values on either profile axis
  favor the optimizer named first. ADANA's token multiplier rises strongly
  with OT, Muon's is comparatively flat, and SOAP's turns upward at the
  longest horizons.}
  \label{fig:uniform-wd-vs-adamw-outscaling-grid}
\end{figure}

\paragraph{Relative to AdamW.}
The three non-Adam optimizers exhibit different horizon dependence
(Figure~\ref{fig:uniform-wd-vs-adamw-outscaling-grid}). ADANA begins slightly
behind AdamW, but its AdamW-relative token multiplier increases with
overtraining factor. Muon is strongest over most short and intermediate
horizons, although its advantage over AdamW remains comparatively flat.
SOAP's advantage decreases through intermediate OT factors and rises at the
longest horizons. The floor--decay profiles show that these trends need not
have the same asymptotic explanation: ADANA must improve upon AdamW in its
token-decay exponent, its high-token loss limit, or both, whereas SOAP's late
gain is more compatible with a lower fitted high-token limit despite a worse
token-decay exponent.

\begin{figure}[tp]
  \centering
  \includegraphics[width=\textwidth]{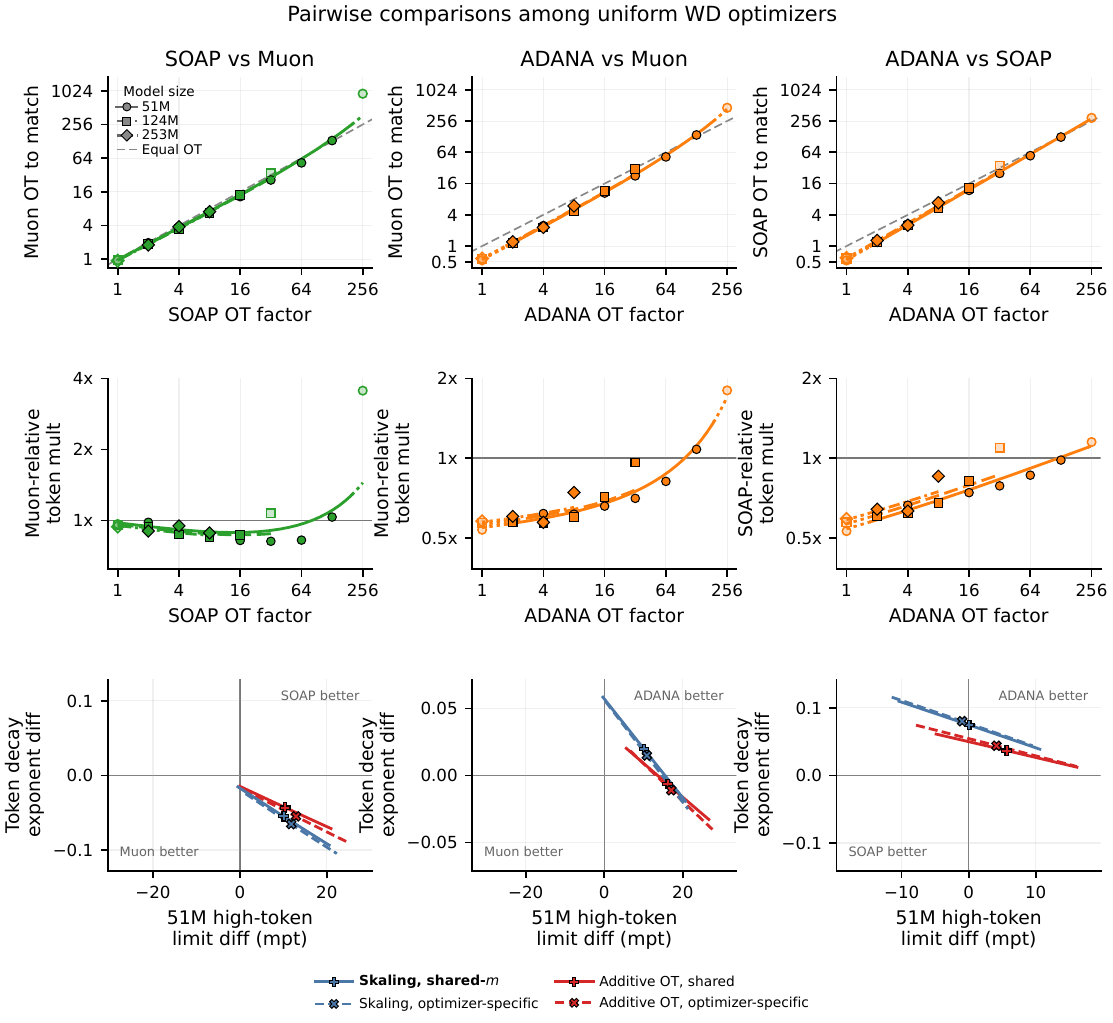}
  \caption{\textbf{Rankings among the non-Adam uniform-WD treatments also
  change with training horizon.} Columns compare SOAP with Muon, ADANA with
  Muon, and ADANA with SOAP. The top, middle, and bottom rows respectively
  show equivalent OT, baseline-relative token multipliers, and floor--decay
  profiles using the conventions of
  Figure~\ref{fig:uniform-wd-vs-adamw-outscaling-grid}. SOAP overtakes Muon
  only at the longest measured horizons, while ADANA gains on both
  matrix-preconditioned treatments and reaches or exceeds them at some of the
  highest measured model--horizon settings. The profiles show that these
  ranking changes need not have the same asymptotic explanation: SOAP's late
  advantage over Muon is compatible with a lower fitted high-token limit
  despite a worse token-decay exponent, whereas ADANA's comparisons retain a
  tradeoff between the two quantities.}
  \label{fig:uniform-wd-non-adam-outscaling-grid}
\end{figure}

\paragraph{Muon versus SOAP.}
We see that the two matrix-preconditioned optimizers have different scaling
behavior and in fact reverse their ordering as the OT factor increases. Muon
is stronger through most short and intermediate horizons, while SOAP
overtakes it at the longest measured horizons for the 51M and 124M models
(Figure~\ref{fig:uniform-wd-non-adam-outscaling-grid}). Across the profiled
functional forms, this crossover is consistently associated with a tradeoff:
SOAP approaches a lower fitted high-token limit, while Muon retains the better
token-decay exponent.

\paragraph{ADANA versus matrix-preconditioned optimizers.}
ADANA begins behind both Muon and SOAP but becomes increasingly competitive as
the overtraining factor grows. It is the strongest primary treatment for the
51M model at $128\times$ OT, while SOAP is strongest at $256\times$ OT and for
the 124M model at $32\times$ OT. Thus, scheduled memory catches the
matrix-preconditioned treatments over the measured overtraining axis without
uniformly dominating them. The profile comparisons leave open whether ADANA
would ultimately win through a better token-decay exponent, a lower high-token
limit, or both.

\subsection{Strongest treatment comparisons}

The uniform weight decay comparisons above provide a controlled comparison in
which all optimizers use the same weight decay treatment. We next ask how the
optimizers compare when each uses the strongest treatment among those we
studied with sufficient coverage across model sizes and training horizons.

For ADANA, we use log-time weight decay with momentum cooldown. For AdamW and
SOAP, we use log-time weight decay because it improves performance relative to
uniform weight decay. For Muon, we retain uniform weight decay because log-time
weight decay does not improve its performance at high overtraining factors.

Our horizon-tuned momentum experiments show that fixed-memory optimizers can
benefit from tuning momentum separately at each overtraining factor. However,
these experiments are available only for the 51M model, so we cannot perform
the full comparison using horizon-tuned momentum across all three model sizes.

\begin{figure}[tp]
  \centering
  \includegraphics[width=\textwidth]{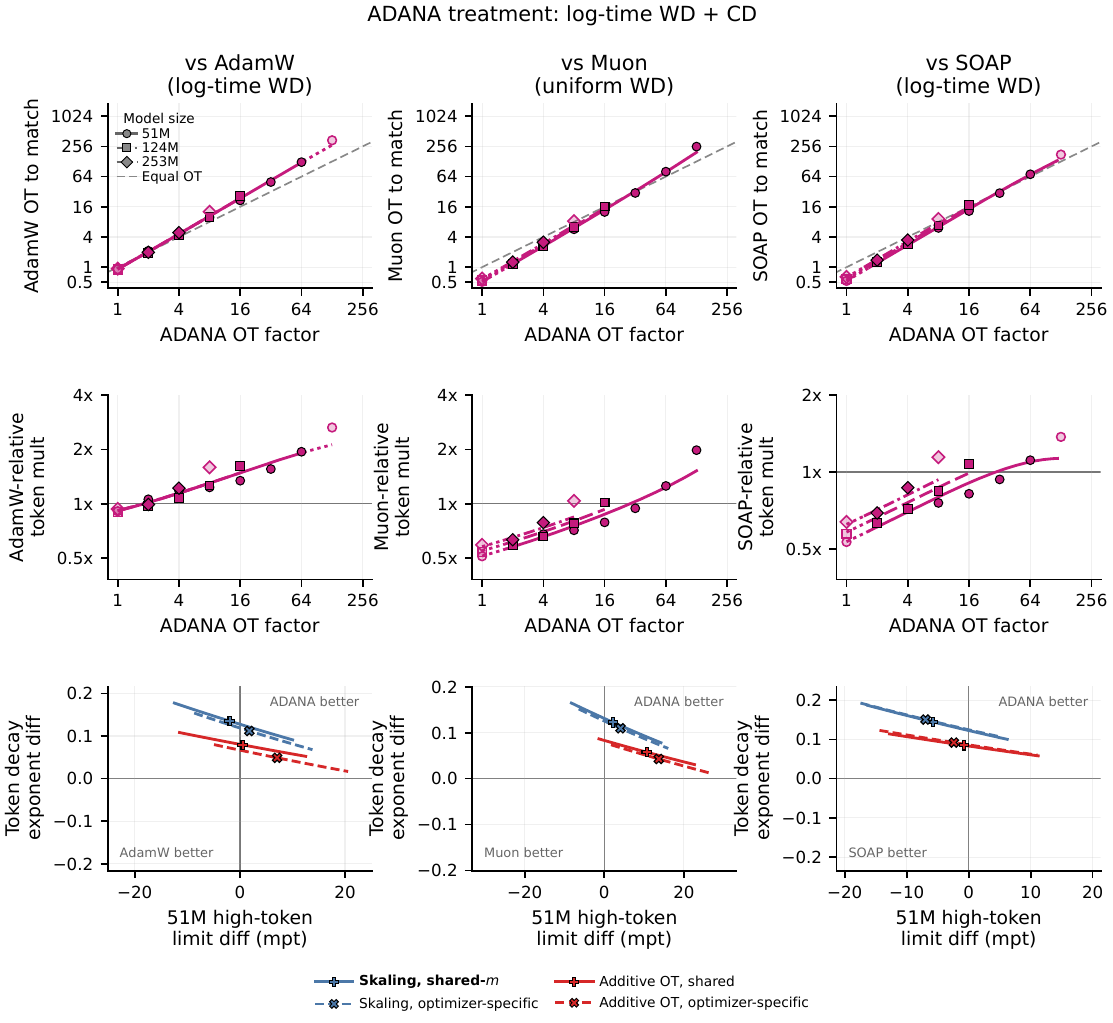}
  \caption{\textbf{Log-time WD and momentum cooldown make ADANA increasingly
  competitive with the strongest measured versions of the three other
  optimizers.} Columns compare ADANA using log-time WD and momentum cooldown
  with AdamW using log-time WD, Muon using uniform WD, and SOAP using log-time
  WD. The top, middle, and bottom rows respectively show equivalent OT,
  baseline-relative token multipliers, and floor--decay profiles using the
  same conventions as
  Figure~\ref{fig:uniform-wd-vs-adamw-outscaling-grid}. Strong ADANA's relative
  token efficiency increases with OT in all three comparisons. Across the
  profiled functional forms, the comparisons favor an improved ADANA
  token-decay exponent, while the corresponding high-token limit differences
  remain less well identified. Open circles indicate points whose learning
  rate sweeps were incomplete; the fits use only the filled points.}
  \label{fig:strong-adana-selected-references-outscaling-grid}
\end{figure}

With each optimizer using its strongest available treatment,
Figure~\ref{fig:strong-adana-selected-references-outscaling-grid} shows that
ADANA's relative token efficiency increases with overtraining factor against
each of the other three optimizers. ADANA begins behind at short horizons,
then catches or overtakes log-time-WD AdamW, uniform-WD Muon, and log-time-WD
SOAP as the training horizon increases.

Overall, these comparisons provide evidence that the strongest ADANA treatment
outscales AdamW and Muon and gains substantially on SOAP across the measured
overtraining axis. We state the SOAP comparison more cautiously because SOAP
performs particularly well for the 51M model at $256\times$ OT. Across all
four profiled functional forms, the data favor an improved ADANA token-decay
exponent, while the precise division between token-decay and high-token-limit
advantages remains less well identified.

\subsection{Equivalent OT scaling and the DANA PLRF prediction}

\paragraph{Effective optimization time.}
An effective optimization clock is a reparameterization of training time that
measures how much progress the optimizer has made through the problem's
spectrum, rather than simply counting updates.  As an example, consider
gradient descent on a quadratic with a time-dependent learning rate $\eta_t$.
Along an eigendirection with curvature $\lambda$, the residual evolves
approximately as
\[
  r_{t,\lambda}
  \approx
  \exp\!\left(-\lambda\sum_{s<t}\eta_s\right)r_{0,\lambda}.
\]
The accumulated learning rate $\sum_{s<t}\eta_s$, rather than the update count
alone, therefore acts as an effective optimization clock.  Two learning rate
schedules that reach the same accumulated value make approximately the same
progress along that eigendirection.

The PLRF analysis of DANA derives an analogous clock for its coupled gradient
and momentum dynamics~\citep{ferbach2025dana}.  In the notation of that
analysis,
\[
  \vartheta(t)
  =
  1+2\gamma_2Bt+
  \left(
    \int_0^t\sqrt{B\gamma_3(s)}\,\mathrm{d}s
  \right)^2,
\]
where $B$ is the batch size, $\gamma_2$ is the direct-gradient step size, and
$\gamma_3(t)$ is the time-dependent step size for the momentum channel.  Note
that $\gamma_3(t)$ in the DANA notation is distinct from the constant ADANA
coefficient $g_3$ used elsewhere in this paper.  The term linear in $t$ is the
direct-gradient contribution to the effective clock, while the
squared-integral term arises from the coupled momentum dynamics.  For the DANA
schedule
\[
  \gamma_3(t)\propto t^{-\kappa},
\]
the latter term grows as
\[
  \left(\int_0^t s^{-\kappa/2}\,\mathrm{d}s\right)^2
  \propto t^{2-\kappa}.
\]
ADANA retains this time exponent through its equivalent EMA formulation, in
which the momentum direction is multiplied by
$(1+t)^{1-\kappa}$~\citep{ferbach2026adana}.

\paragraph{From optimizer updates to equivalent OT.}
In our experiments, batch size and sequence length are fixed, so the number of
optimizer updates is proportional to the number of training tokens and hence
to the overtraining factor at fixed model size.  If the baseline follows a
clock linear in update count while ADANA follows the DANA clock above, the
corresponding equivalent-OT relation is
\[
  f_{\mathrm{baseline}}
  \propto
  f_{\mathrm{ADANA}}^{\,2-\kappa}.
\]
For our choice $\kappa=0.85$, this gives a predicted log--log slope
\[
  2-\kappa=1.15.
\]
Equivalently, DANA's effective clock grows relative to the linear baseline
with exponent $1-\kappa=0.15$.

This prediction has a specific theoretical scope.  In PLRF, DANA-decaying
outscales SGD above the high-dimensional line across the signal- and
noise-controlled phases before the model-capacity floor is reached.  The exact
loss exponent depends on the data and target spectra and on whether population
bias, embedding bias, or variance controls the loss.  Thus, $2-\kappa$ is the
exponent of DANA's effective optimization clock, not a universal loss exponent
shared by every PLRF phase.

\begin{figure}[tp]
  \centering
  \includegraphics[width=\textwidth]{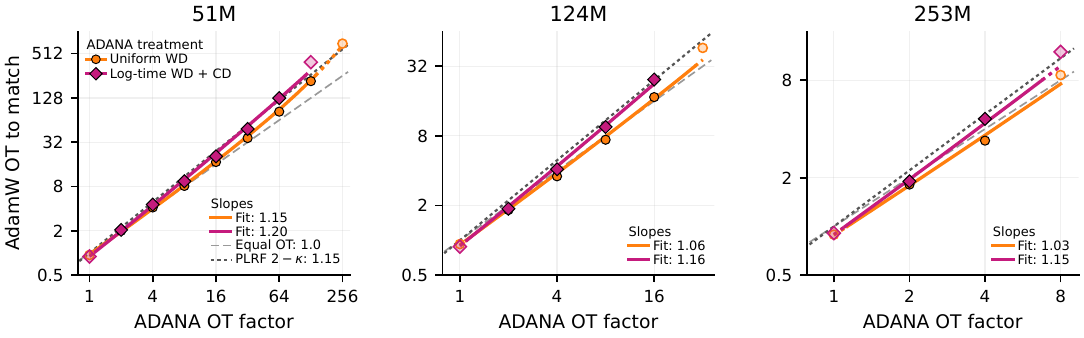}
  \caption{\textbf{With log-time WD and momentum cooldown, ADANA outscaling is close
  to the PLRF prediction $2-\kappa$ across model sizes.} We plot equivalent
  OT scaling across model sizes: for each ADANA OT factor, the vertical axis
  gives the AdamW OT factor required to attain the same fitted validation
  loss. Lines compare uniform-WD ADANA with ADANA using log-time WD and
  momentum cooldown; markers instead invert measured ADANA loss values
  through the fitted AdamW curve. Translucent markers and dotted curve
  segments require extrapolating AdamW beyond its largest measured token
  budget. The displayed fit values are log--log slopes over each curve's
  plotted range. The dashed reference has slope $1$, while the dotted
  reference has the PLRF-predicted slope $2-\kappa=1.15$; both pass through
  $(1,1)$. The combined-treatment slopes are $1.20$, $1.16$, and $1.15$ for
  the 51M, 124M, and 253M models, respectively, whereas the uniform-WD slopes
  are $1.15$, $1.06$, and $1.03$.}
  \label{fig:equivalent-ot-scaling-model-sizes}
\end{figure}

\paragraph{Comparison with the transformer experiments.}
Figure~\ref{fig:equivalent-ot-scaling-model-sizes} compares this prediction
with the equivalent AdamW OT required to match ADANA.  With log-time weight
decay and momentum cooldown, the fitted equivalent-OT slopes are $1.20$,
$1.16$, and $1.15$ for the 51M, 124M, and 253M models.  These values are close
to the PLRF prediction $2-\kappa=1.15$ across all three model sizes.  Under
uniform weight decay, the corresponding slopes are $1.15$, $1.06$, and $1.03$,
indicating that the agreement is strongest for the ADANA treatment combining
log-time weight decay and momentum cooldown.

We treat $2-\kappa$ as a theoretically motivated reference rather than a
direct prediction for these transformer experiments.  DANA theory compares
DANA with SGD on PLRF, whereas our experiments compare adaptively
preconditioned ADANA with AdamW; the strongest ADANA treatment also includes
log-time weight decay and momentum cooldown.  Moreover, the floor--decay
profiles above show that a fitted token-decay exponent remains coupled to the
assumed high-token loss limit.  Our conclusion is therefore that ADANA's
measured equivalent-OT scaling is quantitatively close to the effective-clock
prediction from DANA theory, not that these experiments identify a particular
PLRF phase or establish that the complete PLRF theory transfers unchanged to
transformers.

\section{Batch size and optimizer update count}
\label{app:batch-steps}

At fixed token count, changing the batch size also changes the number of
sequential optimizer updates. If $T$ is the token budget, $B$ is the batch
size in sequences, and $L$ is the sequence length, then the number of updates
is approximately
\[
  S=\frac{T}{B\times L}.
\]
Increasing $B$ therefore simultaneously reduces the number of optimizer
updates and the stochastic variation between updates. This creates a
fixed-token identifiability problem: an optimizer can appear to scale poorly
with batch either because it benefits less from larger batches or because it
benefits more from the additional updates available at smaller batches. The
fixed-token batch comparisons in \citet{ferbach2026adana} and
\citet{marek2025smallbatch} have the same limitation.

\begin{figure}[tp]
  \centering
  \includegraphics[width=\textwidth]{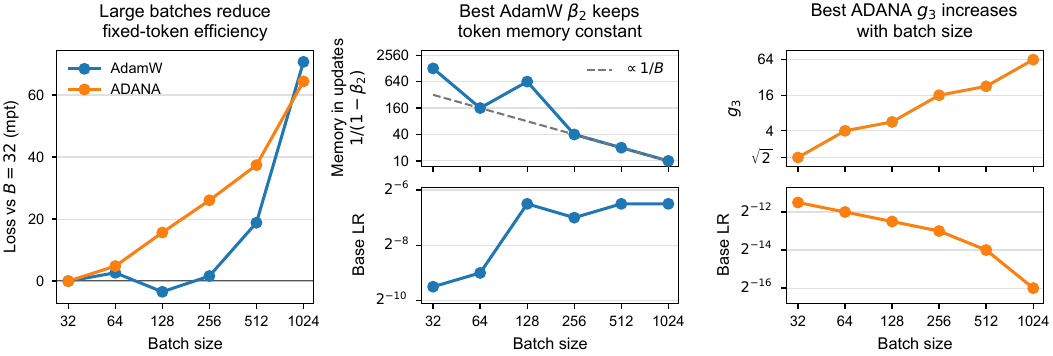}
  \caption{\textbf{At fixed token count, ADANA loses efficiency at smaller
  batch sizes than AdamW in the measured $8\times$-OT slice.} We train the 51M
  parameter model for $8\times$ OT while holding the total number of training
  tokens fixed and varying the global batch size from 32 to 1,024. At each
  batch size, we jointly sweep the base learning rate and either AdamW
  $\beta_2$ or ADANA $g_3$; AdamW $\beta_1=0.9$ and ADANA
  $\kappa=0.85,\delta=8$ remain fixed. Left: the best final validation loss
  relative to $B=32$. Middle: the selected AdamW second-moment memory
  $M_2=1/(1-\beta_2)$; the $M_2\propto 1/B$ guide corresponds to
  approximately constant token memory $BM_2$, consistent with
  \citet{marek2025smallbatch}. Right: the selected ADANA $g_3$. The lower
  panels show the base learning rate selected jointly with each optimizer
  hyperparameter. This fixed-token slice does not by itself establish a
  general batch scaling law for ADANA.}
  \label{fig:appendix-batch-scaling-wave1}
\end{figure}

Figure~\ref{fig:appendix-batch-scaling-wave1} shows our initial batch scaling
experiment at $8\times$ OT. ADANA achieves a lower loss at $B=32$, but its
loss begins to increase at $B=64$ and continues to rise with batch size.
AdamW remains close to its $B=32$ performance through approximately $B=256$
before its loss increases at larger batch sizes. Both optimizers lose
efficiency at the largest measured batch sizes.

The optimal memory hyperparameters for both optimizers follow clear batch
scaling trends. For AdamW, the optimal second-moment memory
$M_2=1/(1-\beta_2)$ decreases approximately as $1/B$, keeping $BM_2$
approximately constant and replicating the batch-transfer behavior observed
by \citet{marek2025smallbatch}. For ADANA, the optimal $g_3$ increases
approximately linearly with $B$, while its jointly optimized base learning
rate decreases.

We obtain these curves by jointly optimizing the base learning rate and
$\beta_2$ for AdamW, with $\beta_1=0.9$ fixed, and the base learning rate and
$g_3$ for ADANA, with $\kappa=0.85$ and $\delta=8$ fixed. For both optimizers,
we use uniform weight decay with $c(f)=8\sqrt{f}$. Because the per-update decay
is $c/S$ and the number of updates $S$ decreases inversely with batch size at
fixed token count, holding $c$ fixed implements the batch scaling prescription
of \citet{bergsma2025powerlines}.

We propose three possible explanations for the observed differences in how
the optimizers perform across batch sizes. First, ADANA's advantage may depend
on the number of sequential optimizer updates. In the PLRF derivation of DANA,
when $\gamma_3(t)\propto t^{-\kappa}$, the momentum contribution to the
effective optimization clock grows as $BS^{2-\kappa}$, compared with a
baseline contribution proportional to $BS$. Its relative contribution
therefore grows as $S^{1-\kappa}$. This suggests that ADANA's advantage should
become more visible over longer training horizons at a fixed batch size. At
fixed tokens, however, increasing $B$ reduces $S$ and may obscure this
advantage.

Second, ADANA may be particularly effective in the high-noise regime.
Analyses of accelerated-SGD-style methods suggest that some multi-timescale
methods derive their largest advantage from noisy, small-batch gradients
\citep{morwani2025connections}. Increasing the effective batch size could
then reduce ADANA's advantage independently of the accompanying reduction in
optimizer updates.

Third, we leave open whether broader joint hyperparameter optimization would
change the relative batch scaling. Although the optimal $g_3$ follows a clear
batch-dependent trend, $g_3$, $\kappa$, and $\delta$ jointly determine ADANA's
evolving first-moment timescale. Similarly, optimizing AdamW's $\beta_2$ at
fixed $\beta_1$ does not test whether its first- and second-moment timescales
should be transferred together. We have separately studied how ADANA's
optimal weight-decay coefficient scales with training horizon at a fixed
batch size, but not how it changes across batch sizes. Additional
batch-conditioned weight-decay sweeps could therefore test whether
weight-decay transfer contributes to the observed performance differences.

Finally, in language model training, each batch element is a sequence rather
than an independent token, making the appropriate notion of effective batch
size ambiguous. One natural measure is the total number of tokens processed
per update, $B\times L$. However, tokens within the same sequence share
context and are likely more correlated than tokens drawn from different batch
elements. Two configurations with the same $B\times L$ may therefore have
different gradient noise: increasing $B$ while shortening $L$ may provide
more independent information than decreasing $B$ while lengthening $L$.
Comparing different $(B,L)$ factorizations at matched tokens per update would
help identify which notion of batch size governs optimizer performance.

Disentangling these mechanisms requires complementary experimental slices:
varying the training horizon at fixed batch isolates dependence on sequential
updates; varying batch at fixed update count probes the effect of batch and
gradient noise; and varying batch at fixed tokens measures the operational
efficiency tradeoff. Comparing different $(B,L)$ factorizations at matched
tokens per update would further test whether the relevant quantity is tokens
per update or the effective number of independent sequences. The present
experiment should therefore be interpreted as an observation about this
fixed-token, fixed-sequence-length slice rather than as a general batch
scaling law for ADANA.

% Fail-closed data guardrail for subsequent batch figures in this manuscript:
% Do not admit S3380--S3385 or S3449--S3451 as ADANA. S3392--S3397 are also
% excluded. The retained Wave-1 figure above predates and does not depend on
% those families; any new multi-horizon batch figure must use the
% runtime-treatment-validated evidence package.

\section{Preconditioner Order in Fixed- and Scheduled-Memory Optimizers}
\label{sec:preconditioner-order}
\label{app:preconditioner-order-cooldown}

\begin{figure}[tp]
  \centering
  \includegraphics[width=\textwidth]{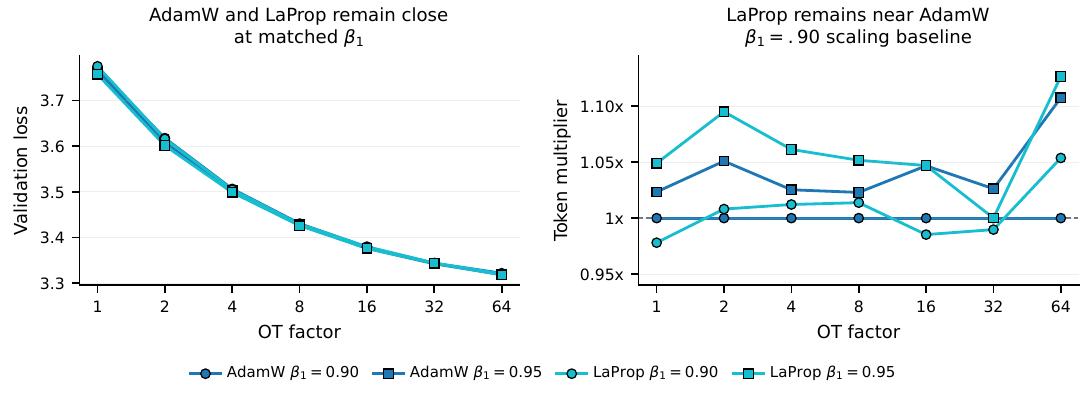}
  \caption{\textbf{AdamW and LaProp remain close at matched $\beta_1$.}
  We compare $\beta_1\in\{.90,.95\}$ at fixed $\beta_2=.98$, independently
  sweeping the base learning rate for every optimizer--$\beta_1$--OT
  coordinate. Left: best validation loss. Right: token multiplier relative
  to AdamW with $\beta_1=.90$ and $\beta_2=.98$.}
  \label{fig:appendix-adamw-laprop-preconditioner-order}
\end{figure}

In this section, we propose the LADANA optimizer, which first applies
adaptive preconditioning to each gradient and then applies the DANA log-time
momentum rule. LADANA can therefore be viewed as an RMSProp-style
preconditioner followed by DANA, reversing the order of preconditioning and
momentum accumulation used by ADANA.

The first scientific question is whether ADANA requires
unpreconditioned gradient-scale information to outscale across the
overtraining axis. ADANA accumulates raw gradients in a memory whose duration
grows throughout training and applies the current adaptive preconditioner only
after this accumulation. Changes in gradient magnitude across coordinates and
training time therefore remain represented in its long-memory state. LADANA
instead normalizes each gradient before it enters that state. If ADANA's
advantage comes primarily from the growing memory clock, LADANA should
preserve its improving token efficiency at longer horizons. If raw gradient
magnitudes contain information that the mechanism relies on, preconditioning
before accumulation may weaken the advantage. Comparing ADANA and LADANA
therefore tests whether log-time momentum can produce outscaling when applied
to preconditioned gradients.

The second scientific question is how preconditioner ordering affects
stability---in particular, whether LADANA has stability advantages over
ADANA. LaProp motivates normalization-first ordering by identifying a
potentially destabilizing interaction in Adam: Adam accumulates raw gradients
in its momentum buffer and then preconditions the entire buffer using the
current second-moment estimate. Past gradients are therefore rescaled using
statistics from a later point in training, which can be problematic when the
gradient distribution changes. LaProp instead normalizes each gradient using
its contemporaneous second-moment estimate before storing it in momentum,
decoupling the momentum and adaptivity time scales and remaining stable in
hyperparameter regimes where Adam can diverge \citep{ziyin2020laprop}. This
concern may be especially important for log-time momentum. Because DANA's
effective memory grows during training, a gradient spike or a mismatch between
past gradients and the current preconditioner can remain influential for much
longer than under conventional fixed-memory momentum. LADANA may therefore
offer a stability advantage by normalizing unusual gradients before they enter
the growing-memory state. Our present experiments compare final validation loss
but do not yet directly establish this stability advantage.

\subsection{AdamW and LaProp isolate preconditioner order}
\label{app:fixed-memory-preconditioner-order}

AdamW and LaProp provide a fixed-memory comparison of the two possible
preconditioner orderings. Let
\begin{equation}
  v_t=\beta_2v_{t-1}+(1-\beta_2)\widetilde g_t^{\odot 2},
  \qquad
  \widehat v_t=\frac{v_t}{1-\beta_2^t}.
\end{equation}
AdamW first accumulates the raw gradient,
\begin{equation}
  m_t=\beta_1m_{t-1}+(1-\beta_1)\widetilde g_t,
  \qquad
  d_t^{\mathrm{AdamW}}
  =\frac{\widehat m_t}{\sqrt{\widehat v_t}+\epsilon},
\end{equation}
and therefore applies the current preconditioner to the complete momentum
buffer.

LaProp reverses these operations \citep{ziyin2020laprop}. It first normalizes
the current gradient,
\begin{equation}
  z_t=\frac{\widetilde g_t}{\sqrt{\widehat v_t}+\epsilon},
\end{equation}
and then accumulates the normalized direction. We follow the official LaProp
implementation, which places the scheduled learning rate inside the momentum
state:
\begin{align}
  \mu_t&=\beta_1\mu_{t-1}+(1-\beta_1)\eta_tz_t,\\
  c_t&=\beta_1c_{t-1}+(1-\beta_1)\eta_t,
  &
  d_t^{\mathrm{LaProp}}&=\frac{\mu_t}{c_t}.
\end{align}
For both optimizers, the outer update is
\begin{equation}
  \theta_t=\theta_{t-1}-\eta_t\Gamma d_t-\omega_t\theta_{t-1},
\end{equation}
where $\Gamma$ denotes the parameter-group learning rate scale and $\omega_t$
is the matched decoupled weight-decay transition. Thus, AdamW averages
gradients and then normalizes, whereas LaProp normalizes each gradient using
the second-moment estimate available when it is observed and then averages the
resulting directions.

Figure~\ref{fig:appendix-adamw-laprop-preconditioner-order} compares the two
optimizers at matched $\beta_1\in\{0.90,0.95\}$, fixed $\beta_2=0.98$, and the
same uniform-WD prescription. We independently sweep the base learning rate
for every optimizer, momentum, and OT coordinate. AdamW and LaProp remain
close across the measured horizons. LaProp with $\beta_1=0.95$ has a modest
advantage at the shorter horizons, but reversing the preconditioner order does
not produce a separation that grows systematically with OT.

This comparison establishes the fixed-memory control for the experiments
below: changing preconditioner order alone has only a modest effect on final validation loss
performance under these treatments. It does not determine whether
preconditioned gradients can support outscaling from log-time momentum,
because both AdamW and LaProp retain a horizon-independent first-moment
window. The corresponding scheduled-memory comparison is therefore between
ADANA and LADANA.

\subsection{LADANA applies DANA after preconditioning}
\label{app:ladana-definition}

We now consider normalization-first ordering for log-time momentum.  For the
log-$v$ variant of LADANA, the second-moment state follows the same log-time
update as ADANA,
\begin{equation}
  v_t
  =
  (1-\Delta_t)v_{t-1}
  +\Delta_t\widetilde g_t^{\odot 2},
  \qquad
  \Delta_t=\frac{\delta}{\delta+t}.
\end{equation}
Let $P_t$ denote the elementwise linear operator
\begin{equation}
  P_t x
  =
  \frac{x}{\sqrt{v_t}+\epsilon},
\end{equation}
and define the gradient normalized at the time it is observed as
$z_t=P_t\widetilde g_t$.  LADANA applies the DANA memory update to these
preconditioned gradients,
\begin{align}
  n_t
  &=
  (1-\Delta_t)n_{t-1}+\Delta_t z_t, \\
  d_t^{\mathrm{LADANA}}
  &=
  g_2 z_t+g_3\chi_t n_t,
  \qquad
  \chi_t=(t+1)^{1-\kappa}+1,
\end{align}
with $n_0=v_0=0$.  The scalar learning rate remains outside this state
recursion.

This notation makes the reversal relative to ADANA explicit.  If $m_t$ is
ADANA's raw-gradient memory, its direction can be written as
\begin{equation}
  d_t^{\mathrm{ADANA}}
  =
  g_2 P_t\widetilde g_t+g_3\chi_t P_t m_t.
\end{equation}
The two optimizers therefore share the same direct-gradient channel.  Their
long-memory channels differ: ADANA stores each unpreconditioned gradient and
applies the current preconditioner $P_t$ to the complete memory, whereas
LADANA stores $P_s\widetilde g_s$ using the preconditioner available when the
gradient was observed.  More precisely, define
\begin{equation}
  w_{t,s}
  =
  \Delta_s\prod_{r=s+1}^{t}(1-\Delta_r).
\end{equation}
Then
\begin{equation}
  d_t^{\mathrm{LADANA}}-d_t^{\mathrm{ADANA}}
  =
  g_3\chi_t
  \sum_{s=1}^{t}w_{t,s}(P_s-P_t)\widetilde g_s.
  \label{eq:ladana-adana-difference}
\end{equation}
Thus the two rules coincide when the adaptive preconditioner is constant over
the effective memory window, and remain close when it changes little over
that window.  Equation~\ref{eq:ladana-adana-difference} isolates
preconditioner order rather than introducing a different direct-gradient
update.

We also consider a fixed-$v$ ablation.  In this variant,
\begin{align}
  r_t
  &=
  \beta_2 r_{t-1}+(1-\beta_2)\widetilde g_t^{\odot 2}, \\
  \widehat r_t
  &=
  \frac{r_t}{1-\beta_2^t},
  \qquad
  z_t
  =
  \frac{\widetilde g_t}{\sqrt{\widehat r_t}+\epsilon},
\end{align}
while $n_t$ retains the log-time update above.  This separates the effect of
preconditioner order from the effect of scheduling the denominator memory.
Unless otherwise stated, our LADANA experiments use log-time memory for both
$n_t$ and $v_t$, with $g_2=1$, $g_3=8$, $\kappa=0.85$, $\delta=8$, and
$\epsilon=10^{-15}$; fixed-$v$ LADANA instead uses $\beta_2=0.98$ for the
denominator memory.

\subsection{LADANA results and limitations}
\label{app:ladana-results}
\label{app:ladana-uniform-wd}
\label{app:ladana-logtime-wd}
\label{app:preconditioner-order-limitations}

Under uniform weight decay, LADANA with log-time second-moment memory performs
similarly to ADANA through $128\times$ OT on the 51M model and slightly
underperforms it at $256\times$.  The two optimizers also remain close across
the shorter horizons measured for the 124M and 253M models.  By contrast,
fixed-$v$ LADANA falls behind at long horizons on the 51M model.  These results
give a qualified positive answer to our first question: outscaling from
log-time momentum can occur on preconditioned gradients, although scheduling
the denominator memory appears important at long horizons.

\begin{figure}[tp]
  \centering
  \includegraphics[width=\textwidth]{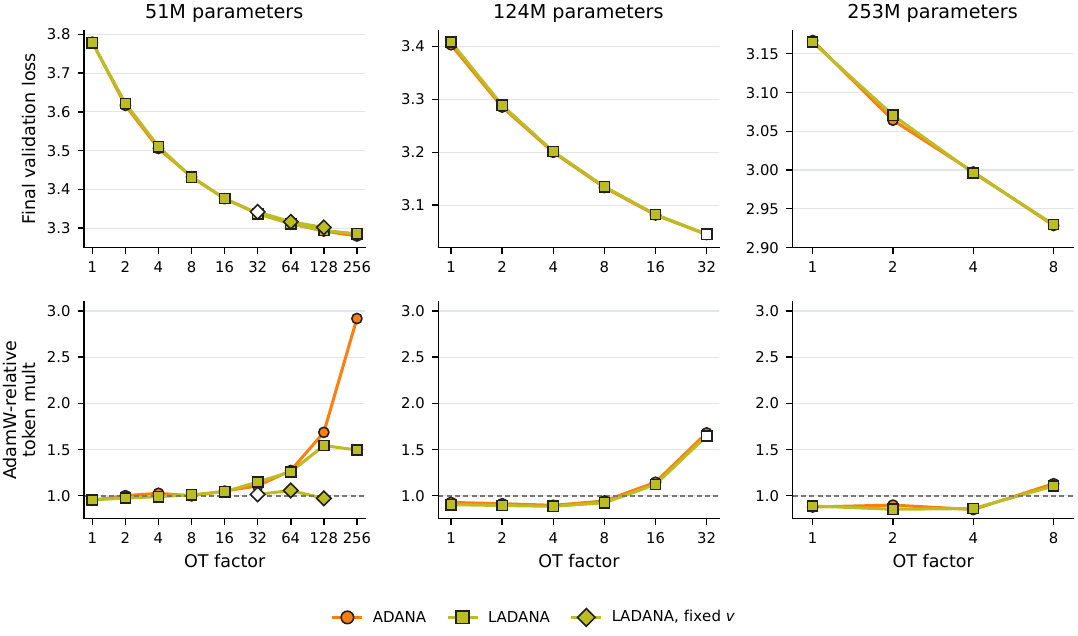}
  \caption{\textbf{Preconditioner order under uniform weight decay across
  model sizes.} The top row reports final validation loss for ADANA and LADANA
  with log-time denominator memory. The bottom row converts the same loss values
  to token multipliers relative to uniform-WD AdamW at each model size. The
  51M panels additionally show the fixed-$v$ LADANA ablation. Open circles
  indicate points whose learning rate sweeps were incomplete.}
  \label{fig:main-preconditioner-order}
\end{figure}

Under log-time weight decay, LADANA performs similarly to ADANA through
$64\times$ OT but deteriorates sharply at $128\times$.  We cannot yet
determine whether this reflects a mismatched weight-decay coefficient, an
interaction between preconditioner order and log-time weight decay, or genuine
instability of the combined treatment.  In particular, we used the log-time
weight-decay coefficient rule determined from experiments on the other
optimizers (Appendix~\ref{app:wd-coefficient-scaling}) without performing a
LADANA-specific weight-decay coefficient sweep.  We therefore treat the
$128\times$ result as an unresolved interaction rather than evidence that
LADANA is inherently unstable.

\begin{figure}[tp]
  \centering
  \includegraphics[width=\textwidth]{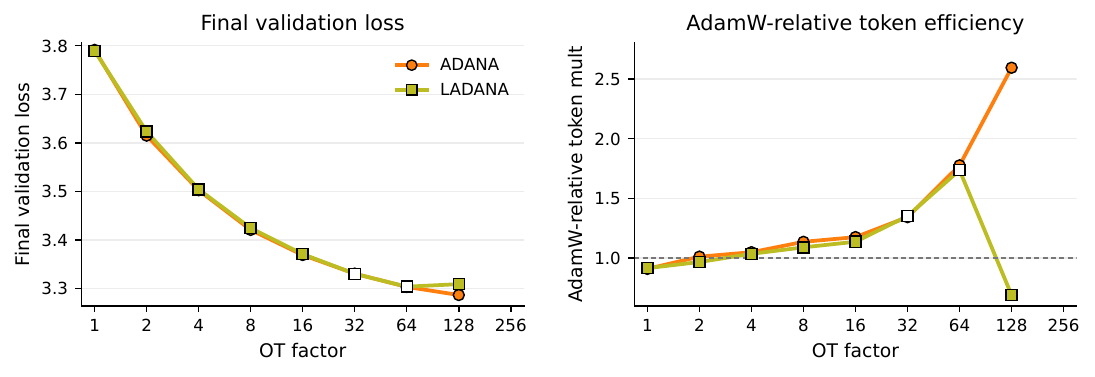}
  \caption{\textbf{Preconditioner order under log-time weight decay.} Final
  validation loss (left) and token multipliers relative to uniform-WD AdamW
  (right) for ADANA and LADANA with log-time denominator memory on the 51M
  model. Both treatments use the log-time weight-decay coefficient rule from
  Appendix~\ref{app:wd-coefficient-scaling}.}
  \label{fig:appendix-preconditioner-order-logtime-wd}
\end{figure}

We leave our second question about stability as future work, as final loss and
learning rate sensitivity cannot establish whether LADANA is more robust to
gradient spikes or changing gradient distributions.  For the current results,
our conclusion is narrower: preconditioning before accumulation preserves
most of ADANA's uniform-WD outscaling, while its stability properties and
compatibility with log-time weight decay remain open.

The complete learning rate sweep evidence for the retained
preconditioner-order comparisons appears in
Figures~\ref{fig:adamw-laprop-beta1-lr-atlas},
\ref{fig:uniform-wd-preconditioner-order-lr-atlas}, and
\ref{fig:logtime-wd-preconditioner-order-lr-atlas}.
Dashed curves and hollow markers are retained wherever the full
$\leq1$ millipoint set reaches a tested learning rate boundary.

\section{Momentum Cooldown for Log-Time Memory}
\label{app:momentum-cooldown}

\subsection{Motivation: long memory can become stale during terminal decay}

Long memory improves gradient averaging, but reduces responsiveness to
changes in the optimization trajectory. Under ADANA's log-time rule, the
momentum timescale grows approximately linearly with the number of optimizer
updates. While the training dynamics change slowly, the optimizer can average
over an increasingly long history without fixing one memory length for the
entire run.

The same persistence may become undesirable during terminal learning rate
decay. Although the learning rate is not stored in the momentum state, the
buffer retains gradients evaluated at earlier parameter values reached using
larger step sizes. When the learning rate changes faster than the momentum
state can respond, substantial kernel mass remains on directions from an
earlier optimization regime. We refer to this mismatch as stale memory: the
stored gradients are not intrinsically invalid, but are old relative to the
timescale on which the current update rule is changing.

This suggests a schedule-aware compromise: retain ADANA's growing memory while
the learning rate changes slowly, but shorten it once terminal decay becomes
faster than the momentum response. Momentum cooldown implements this principle
directly. We treat stale memory as its motivation rather than as an
empirically established explanation of its gains.

\subsection{Definition of momentum cooldown}

Let
\begin{equation}
  \tau_{\mathrm{mom},t}
  =
  \frac{1}{\Delta_t}
  =
  \frac{\delta+t}{\delta}
\end{equation}
denote the momentum timescale of ADANA's original log-time update. We define
the local learning rate decay timescale as
\begin{equation}
  \tau_{\eta,t}
  =
  \frac{\eta_t}{\eta_t-\eta_{t+1}}
\end{equation}
whenever $\eta_{t+1}<\eta_t$, and set $\tau_{\eta,t}=\infty$ during constant
or increasing portions of the schedule. Momentum cooldown replaces the
original timescale with
\begin{equation}
  \tau_{\mathrm{CD},t}
  =
  \max\!\left(
    1,
    \min\!\left\{
      \tau_{\mathrm{mom},t},
      \tau_{\eta,t}
    \right\}
  \right),
  \qquad
  \Delta_t^{\mathrm{CD}}
  =
  \frac{1}{\tau_{\mathrm{CD},t}}.
\end{equation}
The cooled memory state therefore follows
\begin{equation}
  m_t^{\mathrm{CD}}
  =
  \left(1-\Delta_t^{\mathrm{CD}}\right)m_{t-1}^{\mathrm{CD}}
  +
  \Delta_t^{\mathrm{CD}}\widetilde g_t.
\end{equation}

Before the two timescales cross,
$\tau_{\eta,t}\geq\tau_{\mathrm{mom},t}$, so cooldown leaves the original
log-time memory rule unchanged. During sufficiently rapid learning rate
decay, it increases the weight placed on the current gradient and
concentrates the momentum kernel on more recent updates. The construction
depends only on relative changes in the learning rate and is therefore
invariant to its overall scalar scale.

\begin{figure}[tp]
  \centering
  \includegraphics[width=\textwidth]{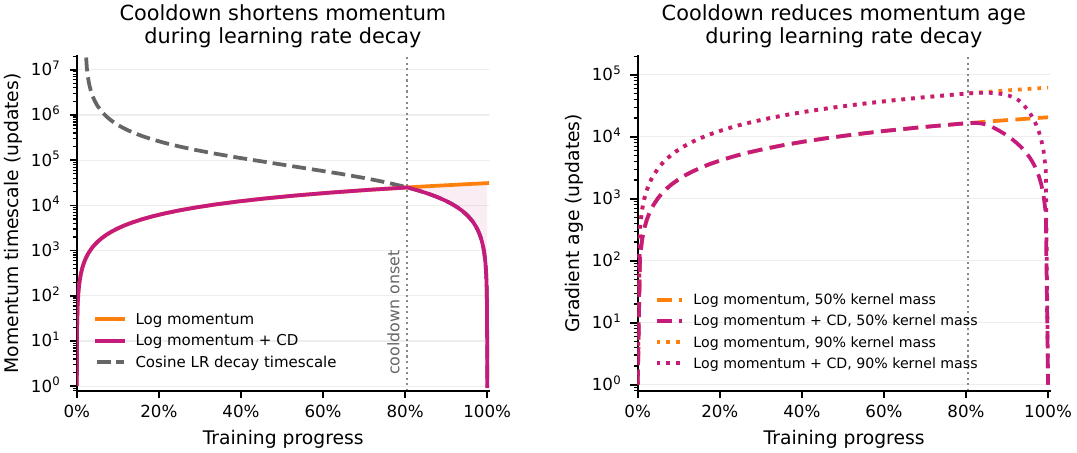}
  \caption{\textbf{Mechanics of momentum cooldown.} We illustrate the exact
  responsive-memory rule used in our experiments for a representative $h8$
  ADANA run at $128\times$ OT with cosine learning rate decay. Left: log-time
  momentum normally uses an increasingly long timescale. We define the
  learning rate decay timescale as
  $\eta_t / |\eta_{t+1}-\eta_t|$ and set the cooled momentum timescale to the
  minimum of this quantity and the original log-time momentum timescale.
  Right: the resulting median and 90th-percentile ages of the momentum kernel,
  defined as the shortest recent history containing 50\% and 90\% of its mass.
  Cooldown leaves the kernel unchanged earlier in training but makes it
  increasingly concentrated on recent gradients during terminal decay.}
  \label{fig:appendix-memory-cooldown-mechanics}
\end{figure}

Because the standard ADANA implementation uses the same log-time coefficient
for its first- and second-moment states, our principal ADANA treatment applies
$\Delta_t^{\mathrm{CD}}$ to both states. We separately examine which states
should be cooled in the treatment-selection experiments below.
Figure~\ref{fig:appendix-memory-cooldown-mechanics} shows how this rule changes
the momentum timescale and the age distribution of the resulting kernel.

\subsection{Momentum Cooldown Treatment Selection}

We distinguish two interventions on ADANA's long-memory channel.
\emph{Memory-window cooldown} shortens the kernel used to update the
long-memory state.  \emph{Long-memory contribution cooldown} leaves that
kernel unchanged and instead multiplies the long-memory contribution
$g_3\chi_t m_t$ by $\min(1,\tau_\eta/\tau_{\mathrm{mom}})$ during learning
rate decay.  The combined treatment applies both changes.  For AdamW, we
analogously apply the memory-window rule to the first moment, the second
moment, or both moments.

\begin{table}[h]
  \centering
  \small
  \setlength{\tabcolsep}{4.5pt}
  \begin{tabular}{@{}llccc@{}}
    \toprule
    Optimizer & Cooldown target & Best base LR & Validation loss &
    \shortstack{Gain vs. standard\\(mpt)} \\
    \midrule
    AdamW & None                     & $2^{-7.0}$  & 3.430405 & 0.00 \\
    AdamW & First moment             & $2^{-7.0}$  & 3.428106 & $+2.30$ \\
    AdamW & Second moment            & --          & \textbf{Unstable} & -- \\
    AdamW & First and second moments & $2^{-7.0}$  & 3.428112 & $+2.29$ \\
    \midrule
    ADANA & None                     & $2^{-12.5}$ & 3.432476 & 0.00 \\
    ADANA & Memory window            & $2^{-12.5}$ & 3.421004 & $+11.47$ \\
    ADANA & Long-memory contribution & $2^{-12.5}$ & 3.431558 & $+0.92$ \\
    ADANA & Window and contribution  & $2^{-12.5}$ & 3.426349 & $+6.13$ \\
    \bottomrule
  \end{tabular}
  \caption{\textbf{Momentum cooldown treatment selection at $8\times$ OT.} Each
  reported minimum comes from an independent base learning rate sweep, and
  gains are measured relative to the same optimizer without cooldown.}
  \label{tab:cooldown-factorial-ot8}
\end{table}

Cooling AdamW's first moment improves validation loss by 2.30 millipoints at
$8\times$ OT.  Cooling both moments gives essentially the same improvement.
By contrast, cooling only the second moment is unstable at every tested
learning rate, producing validation losses from 15.2 to 42.9.  This behavior
is consistent with a response-time mismatch: the second-moment denominator
becomes rapidly responsive during learning rate decay while the first-moment
numerator continues to retain older gradients.  Cooling both moments removes
this mismatch and restores stable training, but provides no additional gain
beyond cooling the first moment alone.  We treat this explanation as a
mechanistic interpretation rather than a demonstrated causal result.

For ADANA, memory-window cooldown improves validation loss by 11.47
millipoints.  Tapering only the long-memory contribution improves loss by
0.92 millipoints, while combining that taper with memory-window cooldown
improves loss by 6.13 millipoints. All three selected minima come from learning
rate sweeps satisfying the interiority criterion. Thus shortening the memory window is the most
effective tested intervention; tapering the long-memory contribution alone
helps only modestly, and combining the two interventions does not improve on
memory-window cooldown alone.

\subsection{Where momentum cooldown helps and what remains open}

We show in Figure~\ref{fig:main-momentum-cooldown} that momentum cooldown
provides a small benefit to AdamW but a substantial improvement to ADANA,
consistent with our motivation that growing log-time memory can become stale
during terminal learning rate decay. The ADANA benefit is consistent across
all three model sizes.

\begin{figure}[h]
  \centering
  \includegraphics[width=\textwidth]{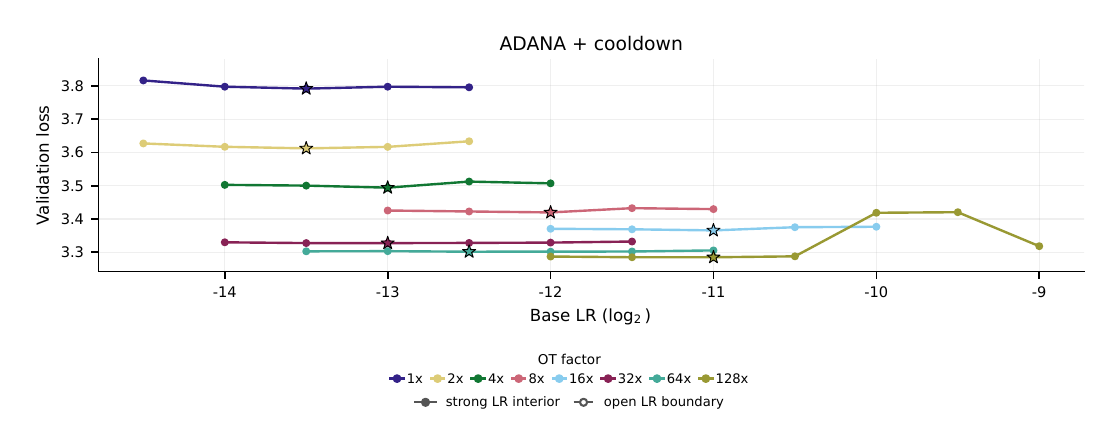}
  \caption{\textbf{Complete log-time-WD learning rate sweeps for the momentum
  cooldown intervention.} The panel shows ADANA with momentum cooldown across
  OT factors $1$ through $128$; the corresponding standard
  ADANA sweeps appear in the log-time-WD preconditioner sweep grid and are not
  repeated here. All eight displayed sweeps satisfy the one-millipoint
  interiority criterion.}
  \label{fig:logtime-wd-cooldown-lr-atlas}
\end{figure}

Moreover, in our experiments on the 51M parameter model, momentum cooldown
and log-time weight decay provide complementary gains. Each intervention
improves long-horizon ADANA, and combining them produces the lowest validation
losses at the longest measured horizons.
Figure~\ref{fig:main-momentum-cooldown} reports these final-loss comparisons;
Figures~\ref{fig:uniform-wd-cooldown-lr-atlas} and
\ref{fig:logtime-wd-cooldown-lr-atlas} provide the underlying learning rate
sweeps.

Several questions remain open. We have not tested analogous cooldown rules
for Muon or SOAP, and the present rule has only been evaluated with cosine
decay to zero. We set the crossover at the unit timescale ratio, but have not
established that this threshold is optimal. Final validation losses also cannot
determine whether downweighting older gradient history is the causal mechanism
behind the improvement. Our present conclusion is therefore narrower:
shortening ADANA's memory during terminal decay produces a substantial
empirical gain, while the optimal rule and its mechanism remain
unresolved.

\clearpage

\section{Complete Hyperparameter Sweeps}
\label{app:complete-sweeps}

This section collects the complete learning rate and nested-hyperparameter
sweeps that support figures elsewhere in the paper. Solid curves with filled markers indicate sweeps
that satisfy the one-millipoint interiority criterion in
Section~\ref{app:sweep-interiority}; dashed curves with hollow markers indicate
sweeps whose within-one-millipoint set reaches a tested boundary.

\subsection{Learning rate schedule selection}

\begin{table}[h]
  \centering
  \small
  \setlength{\tabcolsep}{5pt}
  \renewcommand{\arraystretch}{1.08}
  \resizebox{\textwidth}{!}{%
  \begin{tabular}{@{}llccc@{}}
    \toprule
    Optimizer & Fixed momentum treatment & $c$ at $1\times$ &
      $c$ at $8\times$ & $c$ at $32\times$ \\
    \midrule
    AdamW & $\beta_1=0.9,\ \beta_2=0.95$ &
      $\{2,4,8,16\}$ & $\{4,8,16,32\}$ & $\{16,32,64,128\}$ \\
    ADANA & $g_3=8,\ \kappa=0.85$ &
      $\{4,8,16\}$ & $\{8,16,32,64\}$ & $\{16,32,64,128\}$ \\
    Muon & $\beta=0.95$ &
      $\{2,4,8,16\}$ & $\{8,16,32,64\}$ & $\{16,32,64,128\}$ \\
    SOAP & $\beta_1=0.95,\ \beta_2=0.99,\ \beta_{\mathrm{Sh}}=0.95$ &
      $\{2,4,8,16\}$ & $\{4,8,16,32,64\}$ & $\{16,32,64,128\}$ \\
    \bottomrule
  \end{tabular}}
  \caption{Fixed momentum treatments and coefficients of the uniform WD schedule for
  the linear-to-zero and cosine-to-zero comparison on the 51M parameter
  model.  At every listed coefficient, each schedule receives a separate
  base learning rate sweep.}
  \label{tab:zero-ending-schedule-grid}
\end{table}

\begin{figure}[h]
  \centering
  \includegraphics[width=\textwidth]{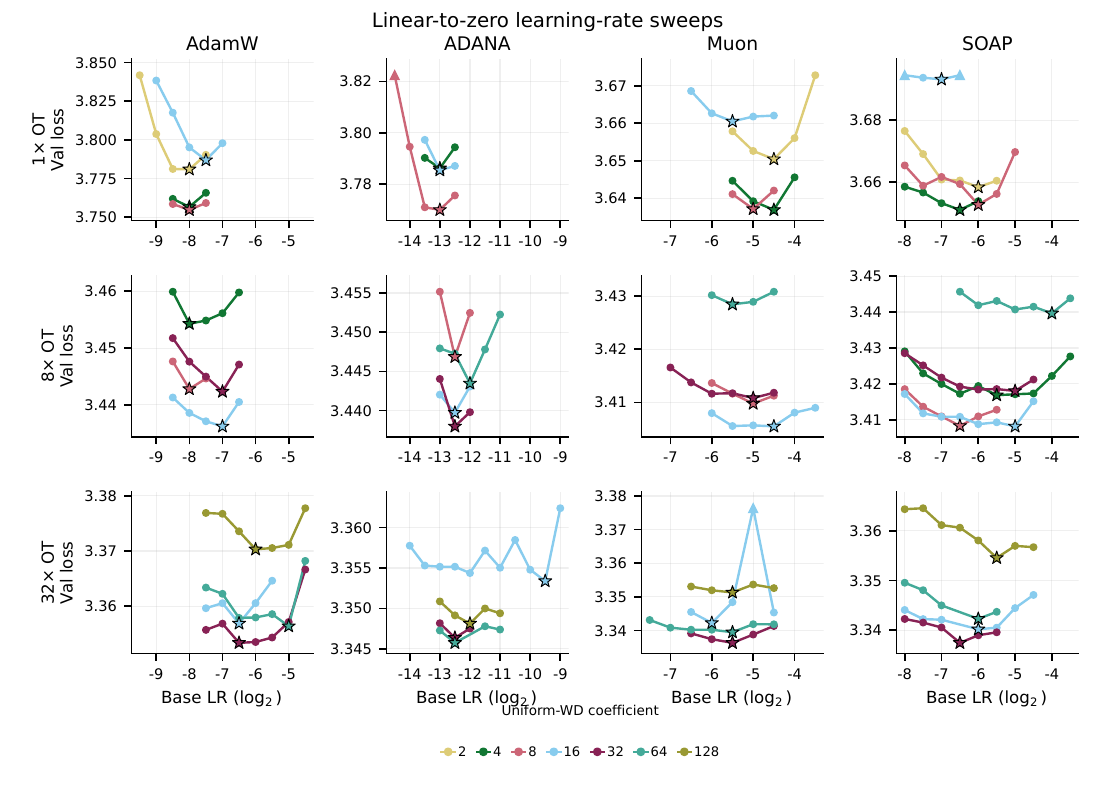}
  \caption{\textbf{Base learning rate sweeps at each uniform WD coefficient
  on the 51M parameter model with linear decay to zero.} Each colored curve
  holds the WD coefficient fixed while varying the base learning rate; stars
  mark the minimum final validation loss within each sweep. Momentum
  hyperparameters are AdamW $(\beta_1=0.9,\beta_2=0.95)$, ADANA
  $(g_3=8,\kappa=0.85)$, Muon $(\beta=0.95)$, and SOAP
  $(\beta_1=0.95,\beta_2=0.99,\beta_{\mathrm{Sh}}=0.95)$. All 48
  displayed sweeps satisfy the one-millipoint interiority criterion.}
  \label{fig:lr-schedule-sweeps-linear}
\end{figure}

\begin{figure}[h]
  \centering
  \includegraphics[width=\textwidth]{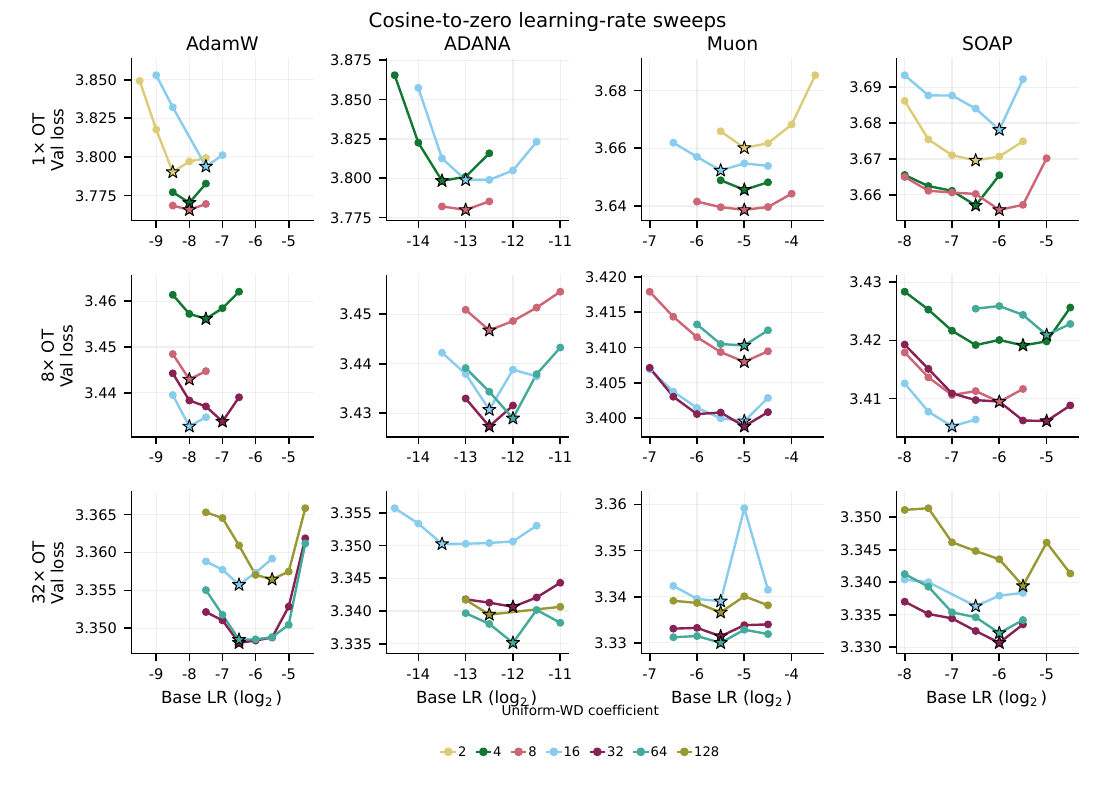}
  \caption{\textbf{Base learning rate sweeps at each uniform WD coefficient
  on the 51M parameter model with cosine decay to zero.} Each colored curve
  holds the WD coefficient fixed while varying the base learning rate; stars
  mark the minimum final validation loss within each sweep. Solid lines with
  filled markers denote sweeps that satisfy the one-millipoint interiority criterion. Momentum
  hyperparameters are AdamW $(\beta_1=0.9,\beta_2=0.95)$, ADANA
  $(g_3=8,\kappa=0.85)$, Muon $(\beta=0.95)$, and SOAP
  $(\beta_1=0.95,\beta_2=0.99,\beta_{\mathrm{Sh}}=0.95)$. These sweeps
  provide the uniform WD grid used in the learning rate schedule comparison.
  All 48 displayed sweeps satisfy the one-millipoint interiority criterion.}
  \label{fig:lr-schedule-sweeps-cosine}
\end{figure}

\clearpage

\subsection{Weight decay sweeps}

The WD schedule comparison and coefficient scaling analysis in
Appendix~\ref{app:weight-decay} use nested sweeps.  At every WD coefficient,
we first select the minimum of an independently extended base learning rate
sweep; only after each contributing learning rate sweep satisfies the
interiority criterion in Section~\ref{app:sweep-interiority} do we compare
coefficients.  The summary figures in Appendix~\ref{app:weight-decay} collapse
this pointwise evidence by minimizing over the learning rate.  Here we report
the complete underlying grids.

Table~\ref{tab:wd-coefficient-grids} gives the fixed momentum treatment and
tested coefficients for every optimizer, WD schedule, and OT factor.

\begin{table}[h]
  \centering
  \small
  \setlength{\tabcolsep}{3pt}
  \renewcommand{\arraystretch}{1.12}
  \resizebox{\textwidth}{!}{%
  \begin{tabular}{@{}llcccc@{}}
    \toprule
    & & \multicolumn{4}{c}{WD coefficient $c$} \\
    \cmidrule(lr){3-6}
    Optimizer & Fixed momentum treatment & $1\times$ OT & $8\times$ OT &
      $32\times$ OT & $64\times$ OT \\
    \midrule
    \multicolumn{6}{@{}l}{\textit{Uniform WD}} \\
    AdamW & $\beta_1=0.9,\ \beta_2=0.98$ &
      $4,8,16$ & $8,16,32$ & $16,32,64,128$ & $16,32,64,128$ \\
    ADANA & $g_3=8,\ \kappa=0.85$ &
      $4,8,16$ & $8,16,32,64$ & $16,32,64,128$ & $16,64,128,256$ \\
    Muon & $\beta=0.98$ &
      $4,8,16$ & $8,16,32,64$ & $16,32,64,128$ & $16,32,64,128,256$ \\
    SOAP & $\beta_1=0.95,\ \beta_2=0.98,\ \beta_{\mathrm{Sh}}=0.95$ &
      $4,8,16$ & $8,16,16\sqrt{2},32$ & $16,32,32\sqrt{2},64$ & $8,32,64,128$ \\
    \addlinespace
    \multicolumn{6}{@{}l}{\textit{Log-time WD}} \\
    AdamW & $\beta_1=0.9,\ \beta_2=0.98$ &
      $1,2,4,8$ & $1,2,4,4\sqrt{2},8$ & $1,2,4,8,16$ & $2,8,16,32,64$ \\
    ADANA & $g_3=8,\ \kappa=0.85$ &
      $1,2,4,8$ & $1,2,4,8,16$ & $1,2,4,8,16,32$ & $2,8,16,32,64$ \\
    Muon & $\beta=0.98$ &
      $1,2,4,8$ & $1,2,4,8$ & $1,2,4,8,16$ & $2,4,8,16,32$ \\
    SOAP & $\beta_1=0.95,\ \beta_2=0.98,\ \beta_{\mathrm{Sh}}=0.95$ &
      $1,2,4,8$ & $2,4,4\sqrt{2},8$ & $1,2,4,8,16,32$ & $1,2,4,8,16,32$ \\
    \bottomrule
  \end{tabular}}
  \caption{WD coefficient grids for the uniform and log-time WD sweep figures.
  Each OT cell lists the coefficients $c$ included in the corresponding
  base learning rate sweeps.}
  \label{tab:wd-coefficient-grids}
\end{table}

\begin{figure}[h]
  \centering
  \includegraphics[width=\textwidth]{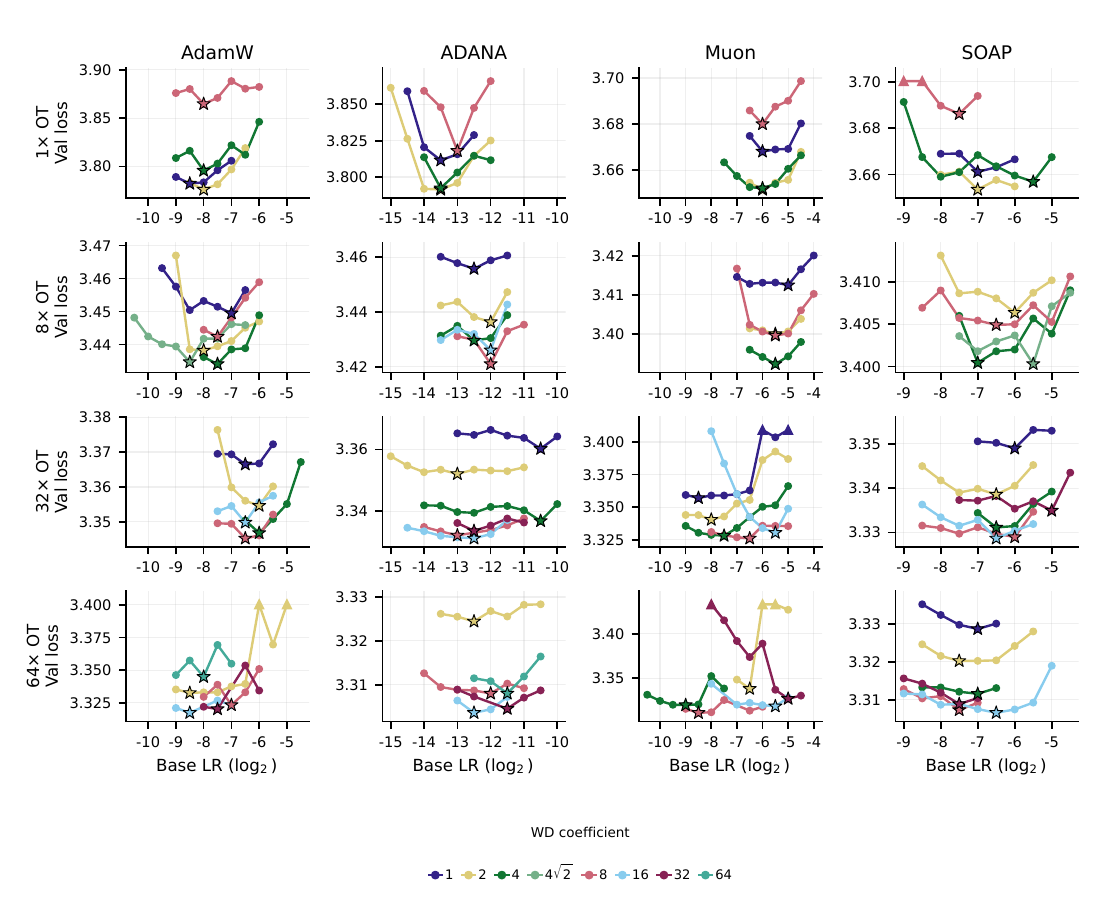}
  \caption{\textbf{Base learning rate sweeps at each log-time WD coefficient
  on the 51M parameter model with cosine decay to zero.} Columns are
  optimizers and rows are OT factors. Each colored curve holds $c$ fixed;
  circles are measured learning rates and black-edged stars mark the minimum
  within each sweep. We use the $1\times$-anchored offset
  $\tau=0.1S_{1\times}$. All 77 sweeps satisfy the interiority
  criterion in Section~\ref{app:sweep-interiority}. Facets use independent vertical
  ranges, and upward triangles retain observed losses above a facet's
  displayed range.}
  \label{fig:wd-log-clock-learning-rate-sweeps}
\end{figure}

\paragraph{Log-time WD coefficient sweeps.}
Figure~\ref{fig:wd-log-clock-learning-rate-sweeps} reports the corresponding
$1\times$-anchored log-time WD coefficient sweeps.

\clearpage
\subsection{Uniform-WD and Log-Time-WD Scaling Sweeps}

\begin{figure}[h]
  \centering
  \includegraphics[width=\textwidth]{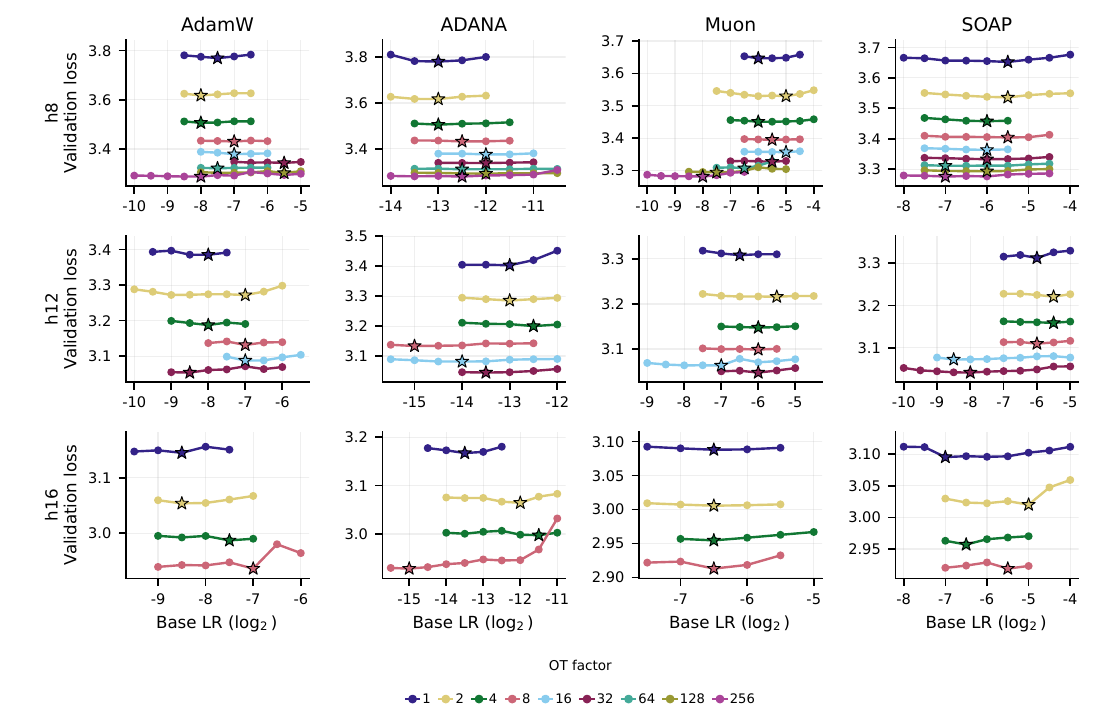}
  \caption{\textbf{Complete uniform-WD learning rate sweeps across model
  sizes and OT factors.}
  Columns are optimizers and rows are model sizes; colors identify OT factor.
  The treatment uses cosine decay to zero and uniform WD with
  $c(f)=8\sqrt{f}$. Circles are measured learning rates and black-edged stars
  mark the selected minimum within each sweep. All 76 displayed sweeps
  satisfy the one-millipoint interiority criterion.}
  \label{fig:primary-gamma-one-lr-sweeps}
\end{figure}

\begin{figure}[h]
  \centering
  \includegraphics[width=\textwidth]{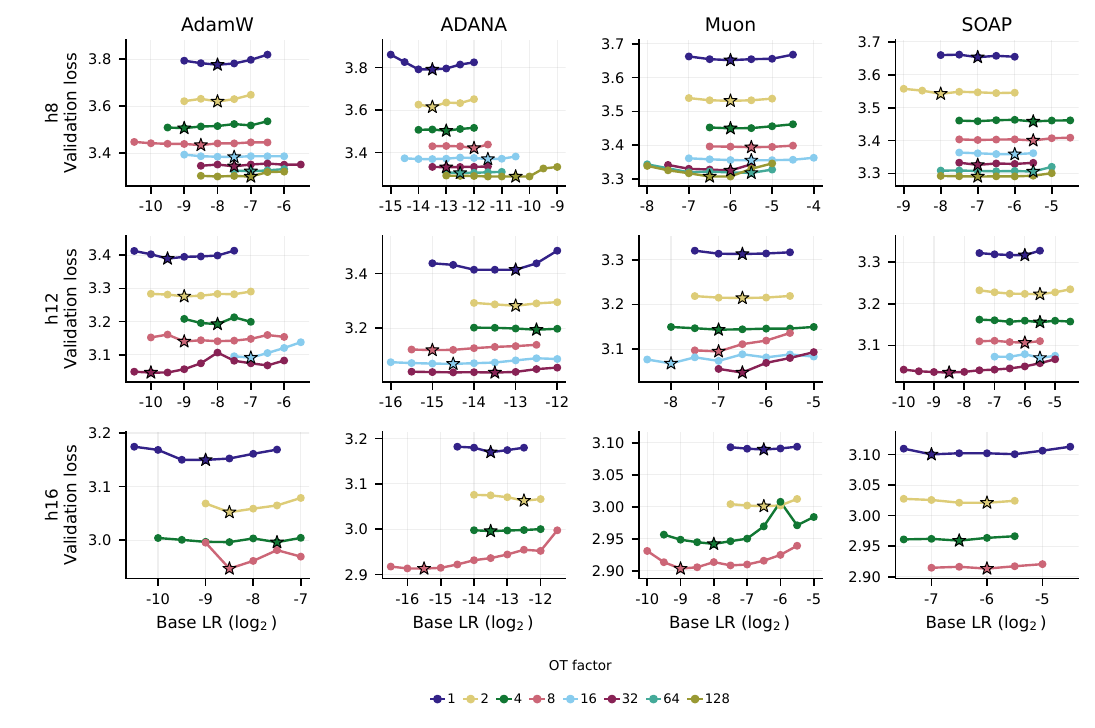}
  \caption{\textbf{Complete log-time-WD learning rate sweeps across model
  sizes and OT factors.} Columns are optimizers and rows are model sizes; colors
  identify OT factor. The treatment uses cosine decay to zero and the
  $1\times$-anchored log-time WD coefficient $a(f)=2\sqrt{f}$. Circles are
  measured learning rates and black-edged stars mark the selected minimum
  within each sweep. All 72 displayed sweeps satisfy the one-millipoint
  interiority criterion.}
  \label{fig:secondary-gamma-one-lr-sweeps}
\end{figure}

Figures~\ref{fig:primary-gamma-one-lr-sweeps} and
\ref{fig:secondary-gamma-one-lr-sweeps} show every independently tuned
learning rate sweep used in the uniform-WD and log-time-WD scaling
comparisons.

\clearpage
\subsection{Fixed-memory sweeps}

\begin{figure}[h]
  \centering
  \includegraphics[page=1,width=\textwidth]{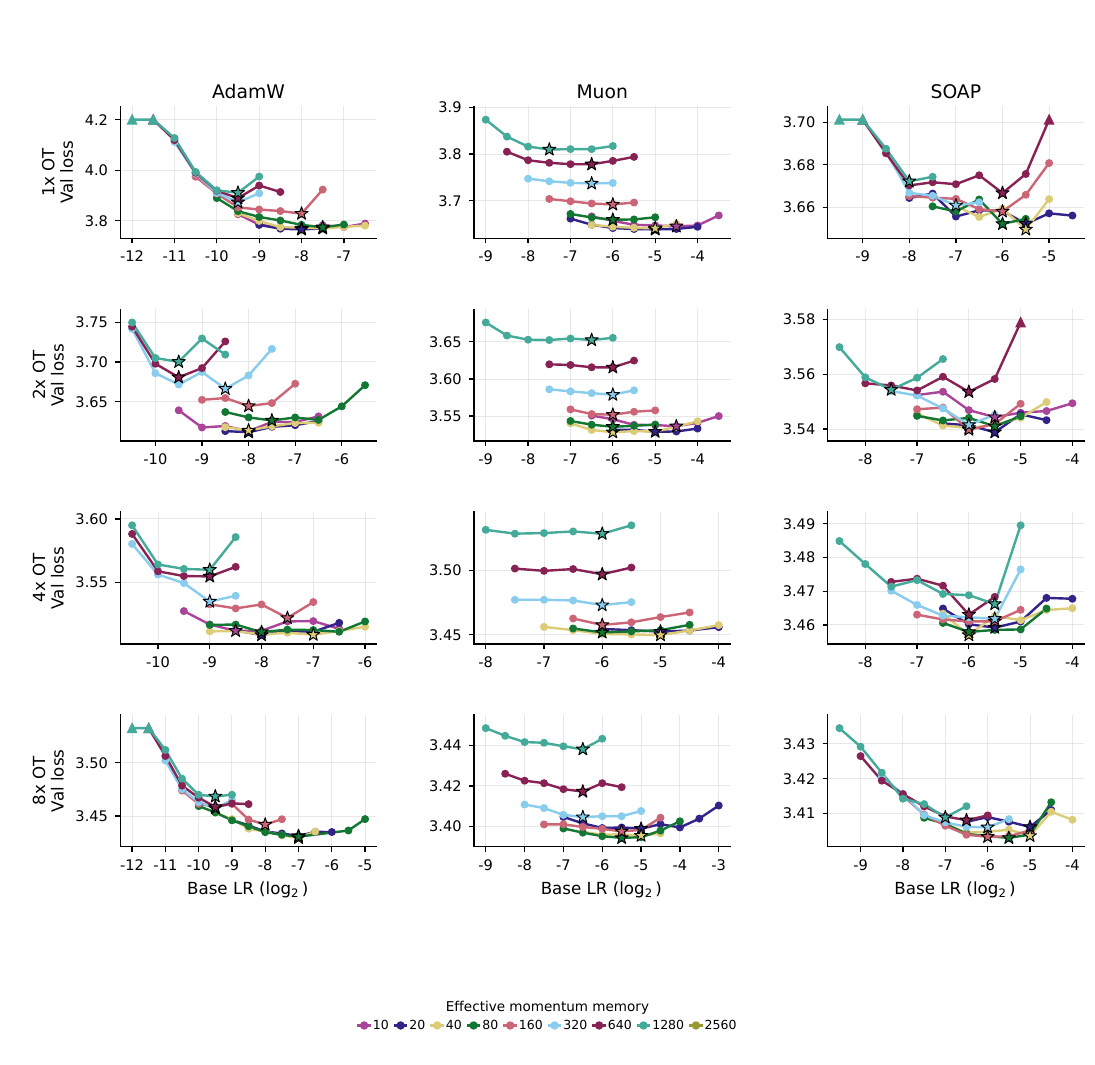}
  \caption{\textbf{Complete fixed-memory learning rate sweeps from
  $1\times$ through $8\times$ OT.} Columns are optimizers, rows are OT
  factors, and colors identify effective memory. Black-edged stars mark the
  minimum of each learning rate sweep. Solid curves with filled markers denote
  sweeps satisfying the one-millipoint interiority criterion; dashed curves
  with hollow markers denote sweeps whose within-one-millipoint set reaches a
  tested boundary.}
  \label{fig:fixed-momentum-lr-sweeps-low-ot}
\end{figure}

\begin{figure}[h]
  \centering
  \includegraphics[page=2,width=\textwidth]{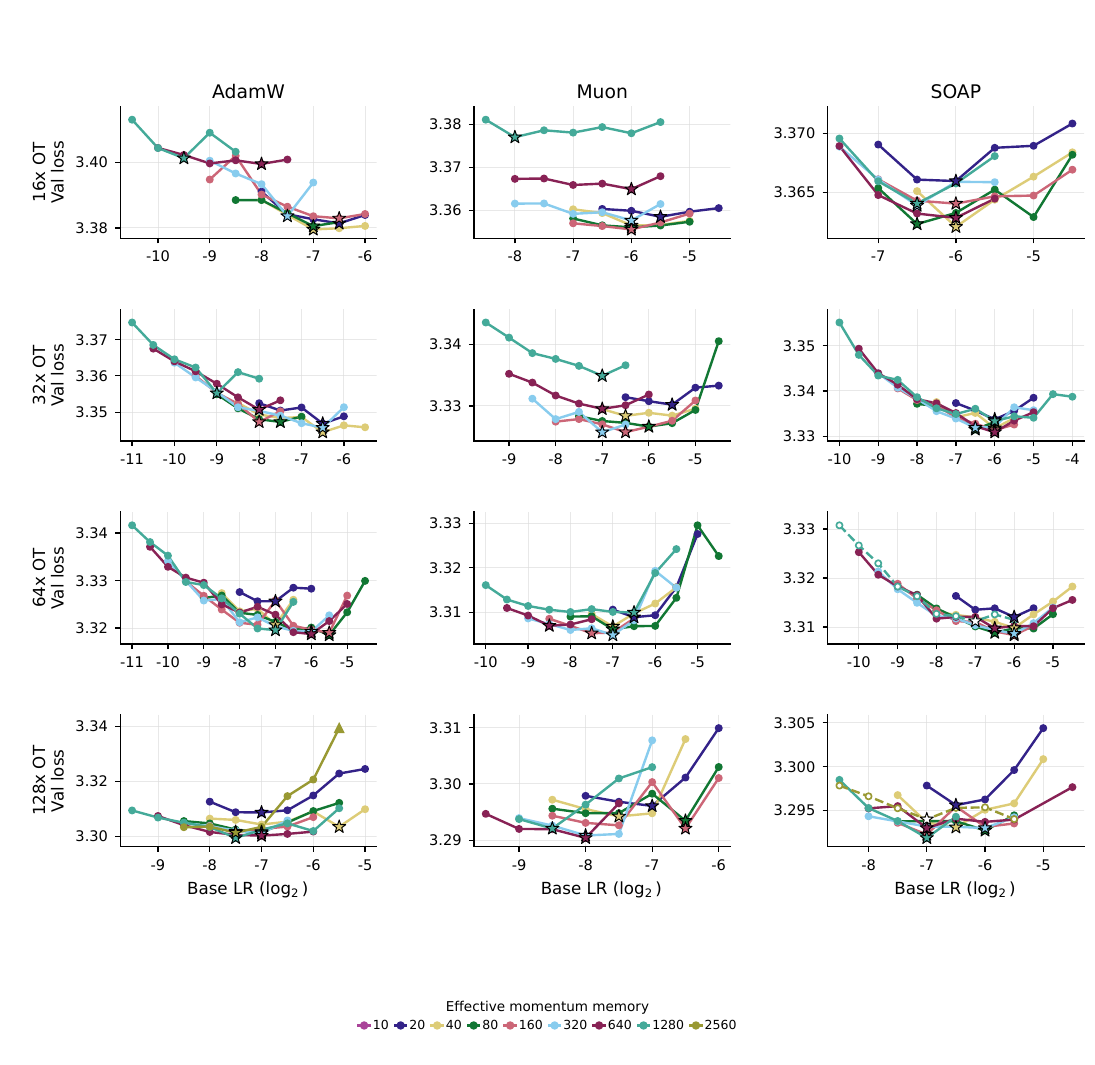}
  \caption{\textbf{Complete fixed-memory learning rate sweeps from
  $16\times$ through $128\times$ OT.} The layout and visual encoding match
  Figure~\ref{fig:fixed-momentum-lr-sweeps-low-ot}. Open circles indicate
  points whose learning-rate sweeps were incomplete. The sweep for SOAP at
  $128\times$ OT with $M=2560$ is incomplete.}
  \label{fig:fixed-momentum-lr-sweeps-high-ot}
\end{figure}

Figures~\ref{fig:fixed-momentum-lr-sweeps-low-ot} and
\ref{fig:fixed-momentum-lr-sweeps-high-ot} report the complete learning rate
sweeps underlying the memory-tuning results in
Appendix~\ref{app:fixed-memory-tuning} and
Figure~\ref{fig:main-memory-frontier}.

\clearpage
\subsection{Preconditioner-order sweeps}

\begin{figure}[h]
  \centering
  \includegraphics[width=\textwidth]{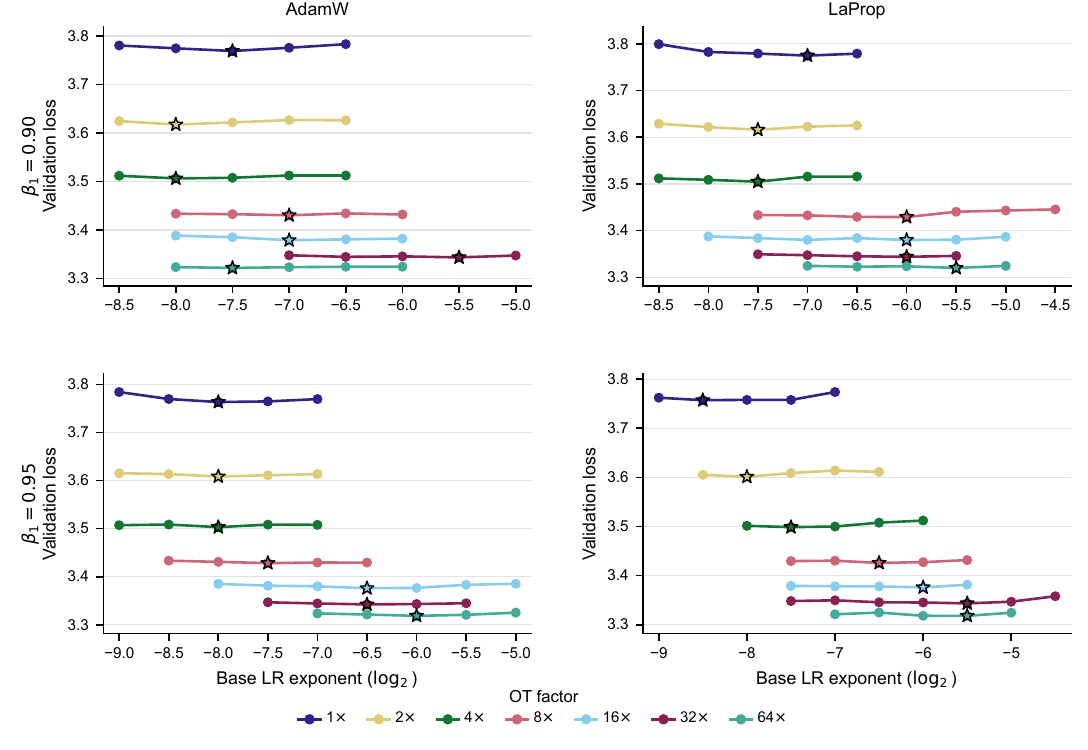}
  \caption{\textbf{Complete learning rate sweeps for the fixed-memory
  preconditioner-order comparison.} Panels compare AdamW and LaProp at
  $\beta_1\in\{.90,.95\}$, fixed $\beta_2=.98$, and OT factors $1$ through
  $64$. Each curve is an independent base learning rate sweep. All 28
  displayed sweeps satisfy the one-millipoint interiority criterion.}
  \label{fig:adamw-laprop-beta1-lr-atlas}
\end{figure}

\begin{figure}[tp]
  \centering
  \includegraphics[width=\textwidth]{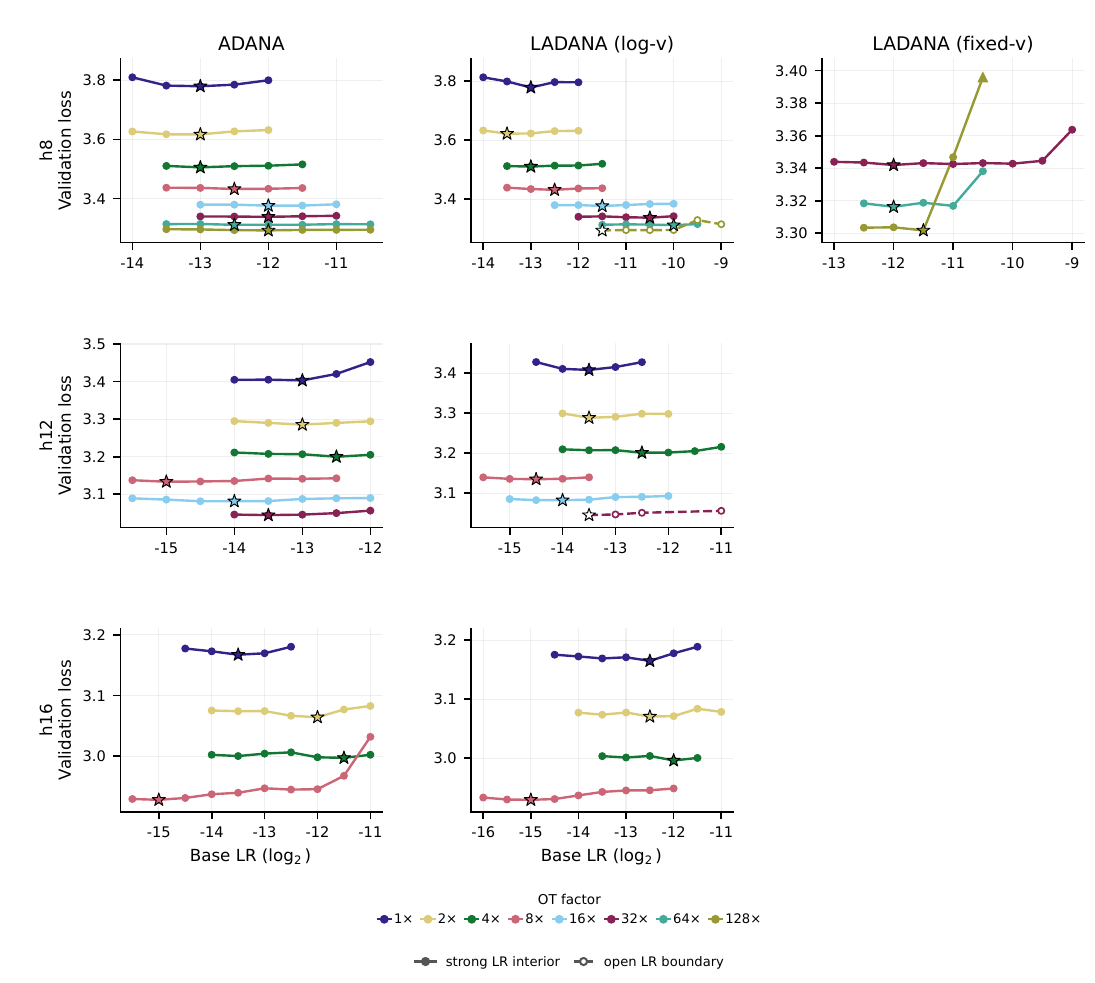}
  \caption{\textbf{Uniform-WD learning rate sweeps for preconditioner order
  across model sizes.} Rows are the 51M, 124M, and 253M models; columns are
  ADANA, LADANA with log-time denominator memory, and the fixed-$v$ LADANA
  ablation (available only for the 51M model). Thirty-seven of 39 displayed
  sweeps satisfy the one-millipoint interiority criterion. Open circles indicate
  points whose learning-rate sweeps were incomplete. The incomplete sweeps are
  log-$v$ LADANA at 51M/$128\times$ and 124M/$32\times$.
  ADANA is retained once as the visual reference
  for the two preconditioner-order treatments.}
  \label{fig:uniform-wd-preconditioner-order-lr-atlas}
\end{figure}

\begin{figure}[tp]
  \centering
  \includegraphics[width=\textwidth]{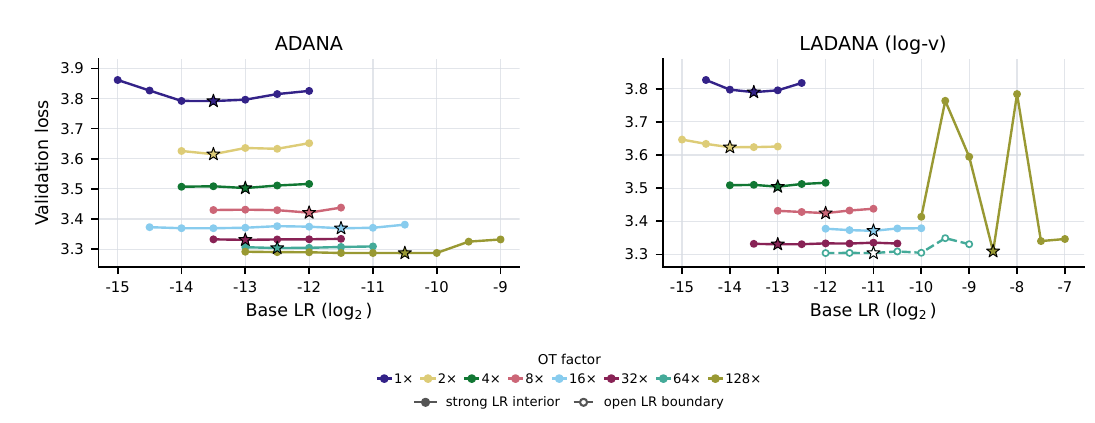}
  \caption{\textbf{Log-time-WD learning rate sweeps for h8
  preconditioner order.} Panels compare ADANA and LADANA with log-time
  denominator memory across OT factors $1$ through $128$. Open circles indicate
  points whose learning rate sweeps were incomplete.}
  \label{fig:logtime-wd-preconditioner-order-lr-atlas}
\end{figure}

Figures~\ref{fig:adamw-laprop-beta1-lr-atlas}--
\ref{fig:logtime-wd-preconditioner-order-lr-atlas} report the learning rate
sweep evidence underlying the retained preconditioner-order comparisons in
Appendix~\ref{app:preconditioner-order-cooldown}. Momentum cooldown treatments are
kept in the separate sweeps below.

\clearpage

\subsection{Momentum Cooldown Sweeps}

\begin{figure}[h]
  \centering
  \includegraphics[width=\textwidth]{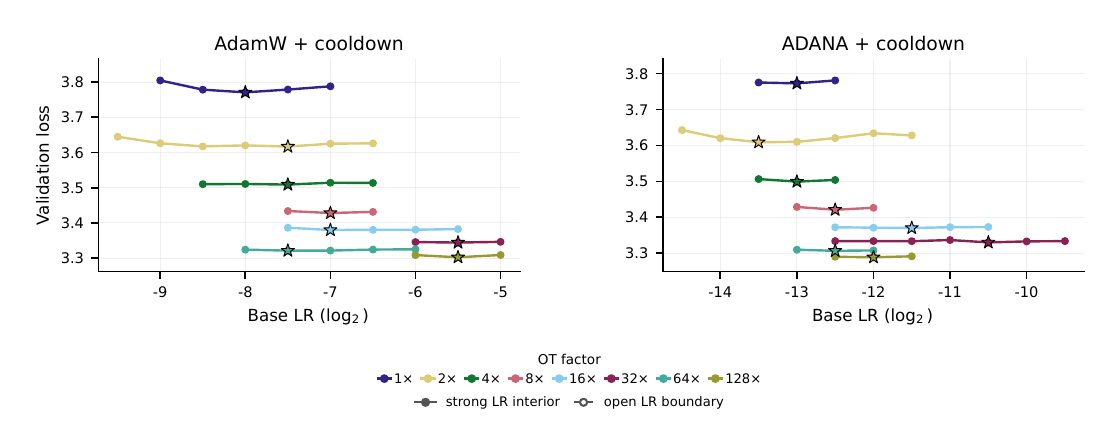}
  \caption{\textbf{Complete uniform-WD learning rate sweeps for the momentum
  cooldown interventions.} Panels show the momentum-cooldown variants of AdamW and
  ADANA across OT factors $1$ through $128$. Corresponding standard-treatment
  sweeps appear in the primary-treatment and preconditioner-order sweeps and
  are not repeated here. All 16 displayed sweeps satisfy the one-millipoint
  interiority criterion.}
  \label{fig:uniform-wd-cooldown-lr-atlas}
\end{figure}

\begin{figure}[h]
  \centering
  \includegraphics[width=\textwidth]{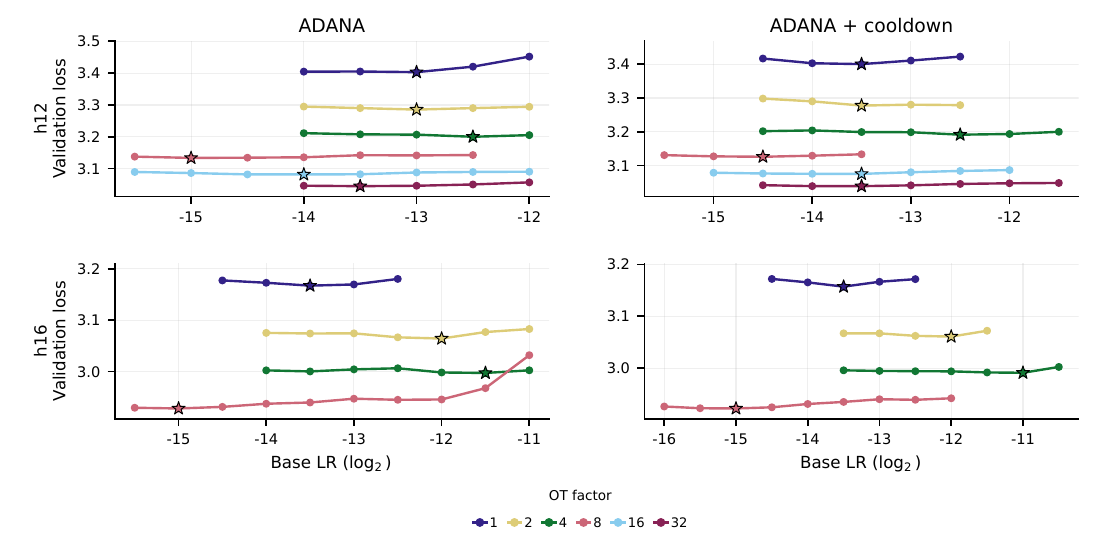}
  \caption{\textbf{Large-model uniform-WD learning rate sweeps for ADANA
  with and without momentum cooldown.} Rows are the 124M and 253M parameter
  models, columns are the standard and momentum-cooldown treatments, and
  colors identify OT factor.}
  \label{fig:large-model-uniform-wd-cooldown-lr-sweeps}
\end{figure}

Figures~\ref{fig:uniform-wd-cooldown-lr-atlas}--
\ref{fig:large-model-uniform-wd-cooldown-lr-sweeps} report the learning rate
sweep evidence underlying the retained momentum-cooldown comparison in
Figure~\ref{fig:main-momentum-cooldown}.

\begin{figure}[h]
  \centering
  \includegraphics[width=\textwidth]{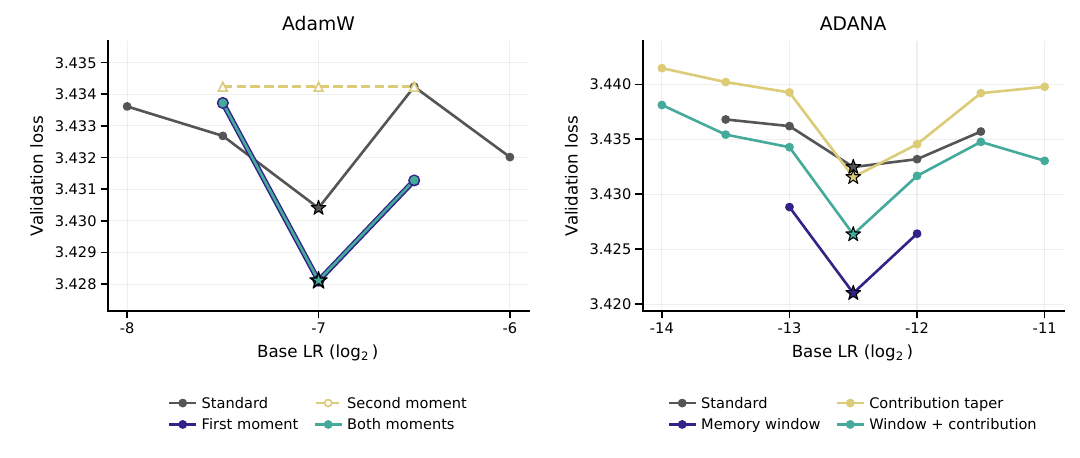}
  \caption{\textbf{Learning rate sweeps for cooldown-component selection at
  $8\times$ OT.} Each panel overlays the independently swept component
  treatments for one optimizer.  AdamW first-moment and both-moment cooldown
  are nearly coincident.  The three second-moment-only AdamW observations have
  losses from 15.2 to 42.9 and are shown as capped upward triangles; they are
  instability outcomes rather than ordinary open minima.  All other displayed
  selected minima satisfy the one-millipoint learning rate interiority rule.}
  \label{fig:cooldown-factorial-ot8-lr-sweeps}
\end{figure}

\paragraph{Momentum cooldown treatment-selection sweeps.}
Figure~\ref{fig:cooldown-factorial-ot8-lr-sweeps} exposes the component-wise
learning rate sweeps underlying Table~\ref{tab:cooldown-factorial-ot8}.

\end{document}